\documentclass[a4paper,12pt,texgyre,numbered,print,authoryear,index]{PhDThesisPSnPDF}

\input{Preamble/preamble}

\title{Multitask Multimodal Self-Supervised Learning for Medical Images}

\subtitle{Towards a framework for foundational models in the Medical Image domain}

\author{Cristian Simionescu}

\dept{Department of Computer Science}

\university{"Alexandru Ioan Cuza" University}
\crest{\includegraphics[width=0.2\textwidth]{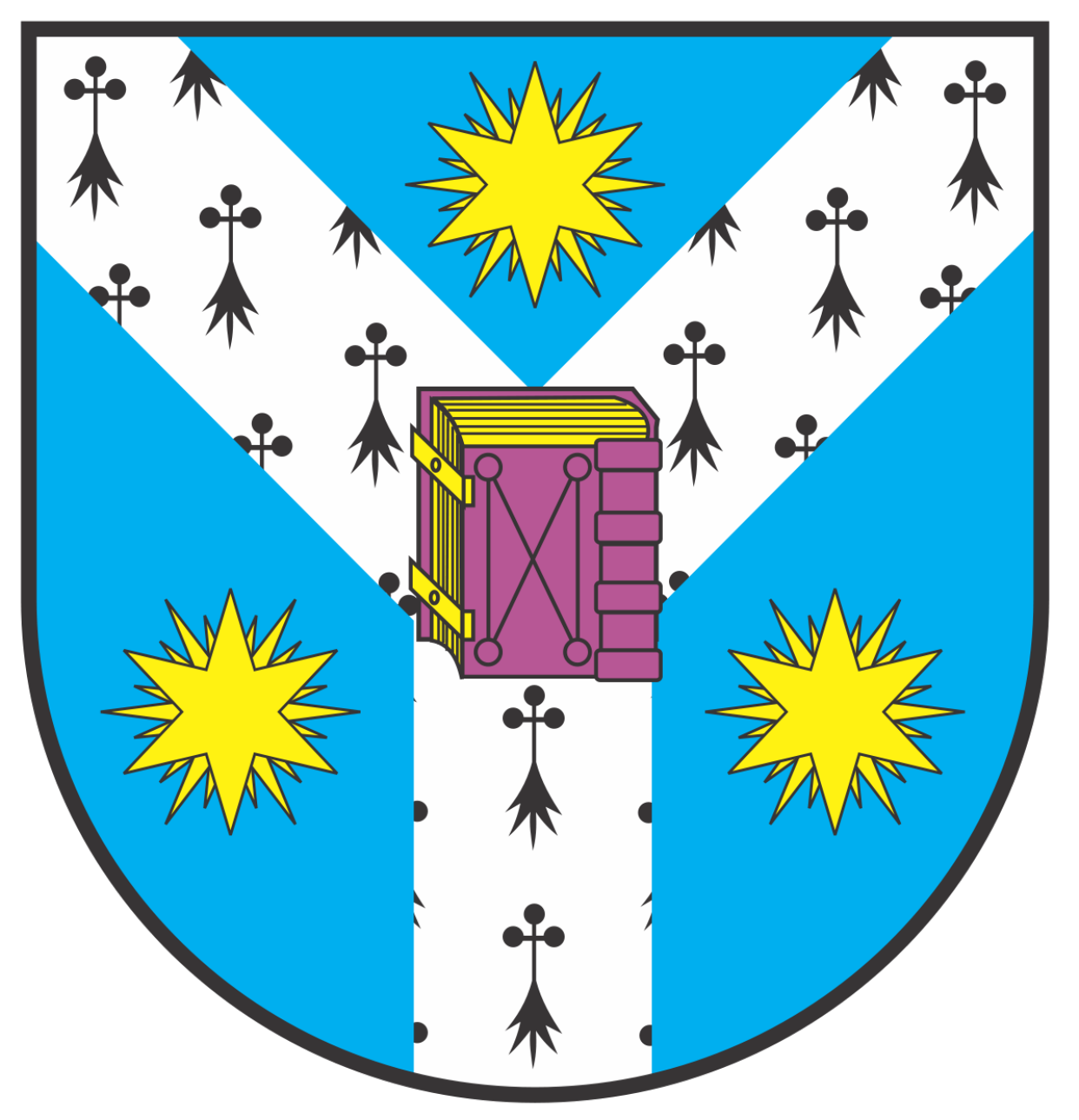}}

\collegeshield{\includegraphics[width=0.2\textwidth]{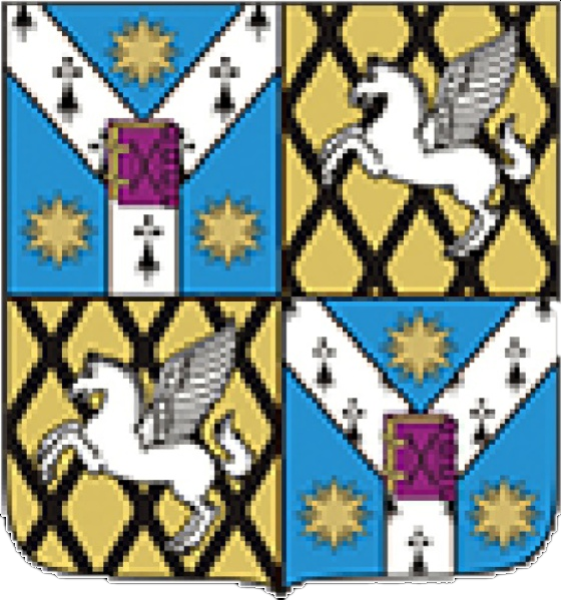}}

\supervisor{Prof. PhD Adrian Iftene}

\supervisorlinewidth{0.4\textwidth}

\advisor{
PhD Anca Ignat\newline
PhD Mihaela Breaban\newline
PhD Razvan Benchea 
}
     
\advisorrole{Advisors: }

\advisorlinewidth{0.4\textwidth}

\degreetitle{Doctor of Philosophy}

\degreedate{March 2025} 

\subject{Machine Learning} \keywords{{Deep Learning} {Self-Supervised Learning} {Medical Image Analysis} {Foundational Model} {Representation Learning} {PhD Thesis} {Computer Science} {Faculty of Computer Science} {"Alexandru Ioan Cuza" University}}

\ifdefineAbstract
\fi

\begin{document}

\frontmatter
\renewcommand{\baselinestretch}{1.2}
\maketitle


\begin{dedication} 

I would like to dedicate this thesis to my loving wife and parents \dots

\end{dedication}


\begin{acknowledgements}

First and foremost, I would like to express my deepest gratitude to my supervisor, 
\textbf{Prof.\ Dr.\ Adrian Iftene}, for his unwavering support, guidance, and 
encouragement throughout my entire doctoral journey. His expertise, patience, and 
faith in my abilities have been instrumental in shaping both my research and my 
personal growth.

I would also like to extend my sincere appreciation to my PhD advisor commission: 
\textbf{Dr.\ Anca Ignat}, \textbf{Dr.\ Mihaela Breaban}, and \textbf{Dr.\ Razvan Benchea}. 
Your constructive feedback, diverse perspectives, and steadfast support were invaluable 
in refining my ideas and challenging me to push the boundaries of my work. While we 
sometimes held differing viewpoints during the review process, it was always productive.

I owe special thanks to my collaborators on various research projects who have contributed 
immensely to my academic and professional development. \textbf{Robert Herscovici}, \textbf{George Stoica}, and \textbf{Cosmin Pascaru}---it has been a privilege to work 
alongside you, learning from your insights and sharing our collective passion for research. 
I am also grateful to the \textbf{IMAGO MOL} team, with whom I had the opportunity to 
collaborate on the REVERT project. More recently, 
\textbf{Cezar Tudor} joined as a valued collaborator, and I am thankful for his 
enthusiasm and expertise, which have broadened the scope and impact of my research.

I also wish to mention some of my fellow PhD colleagues---\textbf{Ingrid Stoleru}, 
\textbf{Gheorghe Balan}, and again \textbf{Cosmin Irimia}---who have been a constant 
source of motivation and friendship. We navigated the challenges of doctoral life together, 
supporting each other throughout.

Finally, I wish to extend my profound gratitude to my family and friends for their 
unwavering love, patience, and encouragement. Their belief in my goals and their consistent 
support, both in moments of triumph and difficulty, have been the bedrock of my resilience.

\end{acknowledgements}

\newpage

\centerline{\textbf{{Research Activity}}}
Publications directly relevant to the thesis:
\begin{enumerate}
    \item \textbf{Simionescu, C.}, Insuratelu, M., \& Herscovici, R. (2020). Prehospital cerebrovascular accident detection using artificial intelligence powered mobile devices. Procedia Computer Science, 176, 2773-2782 - Rank B Conference - 4 points
    \item Hanganu, A., \textbf{Simionescu, C.}, Coca, L. G., \& Iftene, A. (2021). UAIC2021: Lung Analysis for Tuberculosis Classification. In CLEF (Working Notes) (pp. 1253-1263) - Rank D Conference - 0.5 point
    \item \textbf{Simionescu, C.}, \& Iftene, A. (2022). Deep Learning Research Directions in Medical Imaging. Mathematics, 10(23), 4472 - Q1 Journal - 8 points
    \item \textbf{Simionescu, C.} (2023). Privacy-Aware Self-Supervised Deep Learning Approaches for Medical Data. Logic \& Artificial Intelligence, 33 - Rank D Conference - 1 point
    \item Herscovici, R., \& \textbf{Simionescu, C.} (2023, May). BrainFuse: Self-Supervised Data Fusion Augmentation for Brain MRI’s via Frame Interpolation. In International KES Conference on Innovation in Medicine and Healthcare (pp. 164-174). Singapore: Springer Nature Singapore - Rank D Conference - 1 point
    \item Stoica, G., \& \textbf{Simionescu, C.} (2023, June). Backforward propagation (student abstract). In Proceedings of the AAAI Conference on Artificial Intelligence (Vol. 37, No. 13, pp. 16338-16339) - Rank A* Shortpaper - 8 points
    \item \textbf{Simionescu, C.}, \& Stoica, G. (2023, September). Efficient dynamic batch adaptation (student abstract). In Proceedings of the AAAI Conference on Artificial Intelligence (Vol. 37, No. 13, pp. 16328-16329) - Rank A* Shortpaper - 8 points
    \item \textbf{Simionescu, C.}, Herscovici, R., \& Pascaru, C. (2024, June). Cascading Sum Augmentation: Leveraging Populated Feature Spaces. In International KES Conference on Intelligent Decision Technologies (pp. 13-24). Singapore: Springer Nature Singapore - Rank C Conference - 2 points
\end{enumerate}

Other publications:
\begin{enumerate}
    \item \textbf{Simionescu, C.}, Stoleru, I., Lucaci, D., Balan, G., Bute, I., \& Iftene, A. (2019, June). UAIC at SemEval-2019 Task 3: Extracting much from little. In Proceedings of the 13th International Workshop on Semantic Evaluation (pp. 355-359) - Rank A Conference Workshop - 1 point
    \item Minuţ, M. D., Crainic, D. I., Sumănaru, C., Daniş, C., Sava, I., \textbf{Simionescu, C.}, \& Iftene, A. (2022, August). Social Media Post Impact Prediction using Computer Vision and Natural Language Processing. In 2022 International Conference on INnovations in Intelligent SysTems and Applications (INISTA) (pp. 1-6). IEEE - Rank C Conference - 0.4 point
    \item Tudorache, D., \& \textbf{Simionescu, C.} (2022, August). ElectroWay: smart routing for electric car charging. In 2022 International Conference on INnovations in Intelligent SysTems and Applications (INISTA) (pp. 1-5). IEEE - Rank C Conference - 2 points
    \item Vasiliţa, I., Bucnaru, R. I., Barbu, A., Pavel, A., Hoamea, T., \textbf{Simionescu, C.}, \& Iftene, A. (2022, August). Renewable Energy Investment Calculator. In 2022 International Conference on INnovations in Intelligent SysTems and Applications (INISTA) (pp. 1-6). IEEE - Rank C Conference - 0.4 point
    \item Cretu, B. A., Cojocariu, A., Vranceanu, A., Bicu, A., Datco, M., \textbf{Simionescu, C.}, \& Iftene, A. (2022, August). Music Generation using Neural Nets. In 2022 International Conference on INnovations in Intelligent SysTems and Applications (INISTA) (pp. 1-6). IEEE - Rank C Conference - 0.4 point
    \item Minuț, M. D., \textbf{Simionescu, C.}, \& Iftene, A. (2023). Classic Approaches to Bird Song Classification - Rank D Conference - 1 point
    \item Vararu, C., \textbf{Simionescu, C.}, \& Iftene, A. (2023). Integrated Approach for Clothing Detection and Comparison using Structural Shape Detection and Texture Analysis. Procedia Computer Science, 225, 2546-2555 - Rank B Conference - 4 points
    \item Submitted: \textbf{Simionescu C.}, Tudor C.  (2025, September). Zonal Estimation Method for Offline Travel Time Prediction in Sparse Data Environments - Rank B Conference - 4 points
    \item Submitted: \textbf{Simionescu C.}, Tudor C., Archirei S.  (2025, September). ContaGPT: A Domain-Adapted LLM for Romanian Financial and Accounting Applications - Rank B Conference - 4 points
    \item Submitted: \textbf{Simionescu C.} (2025, September). Medformer: A Multitask Multimodal Foundational Model for Medical Imaging - Rank B Conference - 4 points
\end{enumerate}
Total Points (D Rank Capped at 2pct): 40.2\\
Total Including Submitted: 52.2

\begin{abstract}
This thesis works to address a pivotal challenge in medical image analysis: the reliance on extensive labeled datasets, which are often limited due to the need for expert annotation and constrained by privacy and legal issues. By focusing on the development of self-supervised learning techniques and domain adaptation methods, this research aims to circumvent these limitations, presenting a novel approach to enhance the utility and efficacy of deep learning in medical imaging.

Central to this thesis is the development of the Medformer, an innovative neural network architecture designed for multitask learning and deep domain adaptation. This model is adept at pre-training on diverse medical image datasets, handling varying sizes and modalities, and is equipped with a dynamic input-output adaptation mechanism. This enables efficient processing and integration of a wide range of medical image types, from 2D X-rays to complex 3D MRIs, thus mitigating the dependency on large labeled datasets.

Further, the thesis explores the current state of self-supervised learning in medical imaging. It introduces novel pretext tasks that are capable of extracting meaningful information from unlabeled data, significantly advancing the model's interpretative abilities. This approach is validated through rigorous experimentation, including the use of the MedMNIST dataset, demonstrating the model's proficiency in learning generalized features applicable to various downstream tasks.

In summary, this thesis contributes to the advancement of medical image analysis by offering a scalable, adaptable framework that reduces reliance on labeled data. It paves the way for more accurate, efficient diagnostic tools in healthcare, signifying a major step forward in the application of deep learning in medical imaging.
\end{abstract}


\tableofcontents

\listoffigures

\listoftables


\printnomenclature[3em]

\mainmatter

\renewcommand{\baselinestretch}{1.5}

\chapter{Introduction}

\ifpdf
    \graphicspath{{Chapter1/Figs/Raster/}{Chapter1/Figs/PDF/}{Chapter1/Figs/}}
\else
    \graphicspath{{Chapter1/Figs/Vector/}{Chapter1/Figs/}}
\fi

As modern societies continue to integrate digital technologies into every facet of daily life, computational systems are steadily assuming roles once reserved for human judgment and expertise. From personal wearable devices that monitor individual health parameters and recommend lifestyle adjustments to large-scale automated trading systems that guide global financial markets, the influence of computational decision making is pervasive and often subtly embedded in our routines. Even if this transformative shift remains partially hidden from the general public’s awareness, there is little doubt that machine-driven decisions are reshaping the contours of our existence at multiple levels. These systems promise improved efficiency, improved safety, economic gains, and widespread improvements in quality of life; yet, as with any powerful tool, their misapplication or manipulation can produce detrimental results, raising urgent ethical and regulatory concerns \cite{chesney2019deep}.

Against this backdrop, \acro{DL}[Deep Learning] has emerged as a driving force to advance the capabilities of automated decision making. Over the past decade, \acro{DL} approaches have achieved remarkable performance across numerous application domains, including \acro{NLP}[Natural Language Processing], \acro{CV}[Computer Vision], and \textit{Speech Recognition}. These advances are not merely incremental: In some instances, \acro{DL} models have reached or even exceeded human performance on tasks once considered prohibitively challenging for computational systems \cite{wang2018glue, liu2019multi, berner2019dota, arulkumaran2019alphastar}. The relentless pursuit of \acro{SOTA}[State Of The Art] results has fueled a trend toward building larger and more complex neural network architectures, culminating in models such as GPT-3, which contains 175 billion parameters and required millions of dollars of computational resources to train \cite{brown2020language, lambdaGPT3, tweet12MGPT3}, the more recent models presumably costing even more and being larger, but at the same time less transparent about the actual numbers. Although larger models often yield performance gains, this brute-force scaling approach is not without concerns. The economic, environmental, and practical implications of developing such massive networks prompt serious reflection on the sustainability and universality of this trajectory, especially when the internal workings of these so-called “black-box” models remain difficult to interpret \cite{8400040}.

Nowhere are the stakes more apparent than in the medical domain. Healthcare is increasingly reliant on advanced imaging modalities such as \acro{MRI}[Magnetic Resonance Imaging], \acro{CT}[Computed Tomography], and \acro{PET}[Positron Emission Tomography] to support critical decisions spanning diagnosis, prognosis, and therapy planning. Deep learning has shown substantial promise in this area: neural networks can detect subtle pathological changes in imaging data, predict patient outcomes, and assist clinicians in therapeutic decision-making. The potential benefits are immense. As the global population ages, health systems face significant challenges: a shortage of medical staff, increasing costs, and the risk of human error due to overwork and subjective assessments \cite{clements2008overcrowding, schwab2012understaffing, metcalf2018hospital, popescu2014economic}. By harnessing the power of \acro{DL}, it may be possible to alleviate these burdens, streamline diagnostic workflows, and ultimately enhance patient care.

However, the translation from research breakthroughs to real-world clinical practice remains limited. Several challenges hinder the widespread adoption of deep learning in medicine. First, conventional supervised learning methods require large-scale annotated datasets for training. In the medical context, obtaining sufficiently large and reliably labeled data is expensive, time-consuming, and encumbered by strict regulatory and ethical constraints designed to preserve patient privacy. Second, \acro{DL} models often operate as “black-boxes,” providing little insight into the features and patterns that drive their predictions. Clinicians and regulatory bodies understandably hesitate to rely on opaque decision-makers for high-stakes medical tasks, demanding greater explainability, transparency, and accountability \cite{arrieta2020explainable, wachter2017transparent, speicher2018unified, singh2020explainable}. Third, both medicine and deep learning are dynamic fields. Rapid advances in imaging technology, evolving clinical standards, and the ongoing refinement of machine learning methodologies continuously shift the research frontier. These factors complicate the effective alignment of \acro{DL} with clinical workflows and best practices.

In response to these challenges, recent trends in machine learning research are shifting toward strategies that reduce reliance on large labeled datasets and improve model robustness, generalization, and interpretability. Among the promising avenues are \acro{SSL}[Self-Supervised Learning] and related paradigms such as unsupervised or weakly supervised learning. Instead of depending solely on manually annotated data, \acro{SSL} leverages the vast reservoirs of unlabeled data. It does so by devising pretext tasks—auxiliary objectives that encourage a model to learn meaningful, domain-relevant representations from unlabeled inputs. Once trained on these pretext tasks, the model’s learned representations can be adapted or fine-tuned for specific downstream tasks, often yielding performance improvements even when the number of labeled samples is limited.

Such approaches are particularly compelling in the medical domain, where the scarcity of annotated data can be a severe bottleneck. By effectively unlocking the informative structure embedded in large-scale unlabeled medical images, \acro{SSL} can pave the way for more data-efficient models. These models may generalize better to new tasks, demonstrate improved robustness to shifts in data distribution, and potentially facilitate more interpretable decision-making. Given these attributes, it is increasingly clear that \acro{SSL} and related methods will play a pivotal role in the future of medical image analysis.

The aim of this PhD thesis is to explore advanced deep learning methods that harness self-supervised and related learning paradigms for medical imaging applications. The core objective is to propose techniques that reduce the heavy dependence on large annotated datasets and maintain or improve predictive accuracy. By synthesizing insights from the latest research in both general \acro{DL} methodologies and medical imaging analytics, this work seeks to contribute to a more sustainable, trustworthy, and clinically impactful integration of AI in healthcare.

In the subsequent chapters, we will first present the theoretical and methodological foundations of deep learning, self-supervised learning, and their intersection with medical image analysis. We will then provide a comprehensive review of the existing literature, identifying critical gaps and opportunities that inform the contributions of this thesis. Following this, we will detail the proposed methods, empirical evaluations, and results. Finally, we will engage in a broader discussion of the implications, limitations, and future directions of this research. By addressing the challenges and opportunities at the nexus of deep learning and medical imaging, this thesis aspires to advance the field and encourage the ethical, effective, and patient-centric deployment of AI-driven healthcare solutions.

\section{Motivation}

In the realm of medical imaging, the quest for advancements in analysis techniques has been driven by the urgent need for more accurate, efficient, and automated diagnosis methods. This doctoral research, located at the confluence of deep learning, medical imaging, and self-supervised learning, seeks to address several critical challenges that currently impede progress in this field.

Firstly, the scarcity of labeled medical image data poses a formidable challenge. The annotation of medical images requires extensive expert knowledge and is often hampered by privacy and legal constraints. This makes the acquisition of large labeled datasets not only costly but sometimes unfeasible. This issue sits at the core of our motivation to pursue the application of Self-Supervised Learning methods. Since the amount of available data is already significantly smaller than natural images, having to constrain ourselves even more to only annotated data is greatly limiting. 
The impact of this issue is further exemplified when we consider that even for the same data modality differences in labeling methodology further split up our datasets, for example, if we have two datasets of lung X-rays for patients who suffer from tuberculosis, one where the samples are classified with numbers from 0 to 5, signifying the severity or progression of the disease, and the other dataset having labels for the type of tuberculosis present in the image, under a supervised learning paradigm we would have no real way to make use of both datasets to train a model, hence we are limited in using smaller, fragmented datasets. We investigated and reviewed the state of the literature field in prior works, such as: \cite{simionescu2022deep} and \cite{simionescu2023privacy}.
Our motivation, therefore for working on self-supervised learning techniques for the medical domain comes from this need to unlock the use of more data to mitigate the dependency on labeled data, enabling better-performing deep learning models. We have also attempted to improve sample efficiency in other works (\cite{simionescu2023efficient}), on general supervised computer vision tasks, that would help in any data-scarce scenario.

\begin{figure}[htbp]
    \centering
    \begin{subfigure}[b]{0.32\textwidth}
        \centering
        \includegraphics[height=4cm]{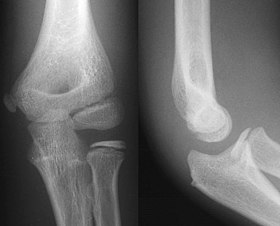}
        \caption{X-ray of an arm\protect\footnotemark}
        \label{fig:xray_arm}
    \end{subfigure}
    \hfill
    \begin{subfigure}[b]{0.32\textwidth}
        \centering
        \includegraphics[height=4cm]{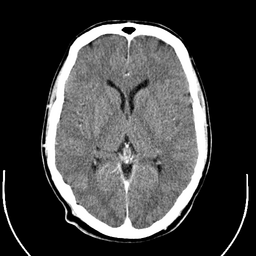}
        \caption{CT scan of the brain\protect\footnotemark}
        \label{fig:ct_brain_1}
    \end{subfigure}
    \hfill
    \begin{subfigure}[b]{0.32\textwidth}
        \centering
        \includegraphics[height=4cm]{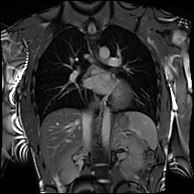}
        \caption{MRI of the chest\protect\footnotemark}
        \label{fig:mri_chest}
    \end{subfigure}
    \caption{Examples of heterogeneous medical imaging modalities illustrating the diversity in data (X-ray, CT, and MRI).}
    \label{fig:medical_modalities_heterogeneity}
\end{figure}

\footnotetext{https://en.wikipedia.org/wiki/Projectional\_radiography}
\footnotetext{https://en.wikipedia.org/wiki/File:Computed\_tomography\_of\_human\_brain\_-\_large.png}
\footnotetext{https://en.wikipedia.org/wiki/Cardiac\_magnetic\_resonance\_imaging}

Secondly, the heterogeneity of medical data, in terms of imaging modalities and tasks, adds another layer of complexity. Different modalities, such as X-rays, CT scans, and MRIs, each offer unique insights but also present distinct challenges, this is further underscored by the multitude of medical specializations in certain organs or body parts, see figure \ref{fig:medical_modalities_heterogeneity}. These two factors cause the overall field to generate very distinct data, a brain CT image will have very little in common with an X-Ray of an arm. We believe this is one of the crucial reasons why self-supervised methods haven't been as prevalent in the medical domain even if, in the natural imaging setting, these are key in obtaining state-of-the-art results. 
This characteristic also applies in the more general sense to the prospects of transfer learning and representation learning. For natural images, the use of models pre-trained on ImageNet as starting points to learning a new task is common and works very well due to the homogeneity of these images, all contain similar patterns and shapes. These properties aren't present for medical images making the task, of learning good general representations that can be reused for any type of medical imaging modality, much harder.
This motivated us to also look into developing ways to train a foundational model that can seamlessly adapt to this diversity and extract pertinent features across various modalities that can be reused for finetuning on downstream tasks. 
Our research introduces the Medformer architecture, a neural network structure adept at multitask learning and deep domain adaptation, specifically tailored for medical image analysis.

\begin{figure}[htbp]
    \centering
    \begin{subfigure}[b]{0.32\textwidth}
        \centering
        \includegraphics[height=5cm]{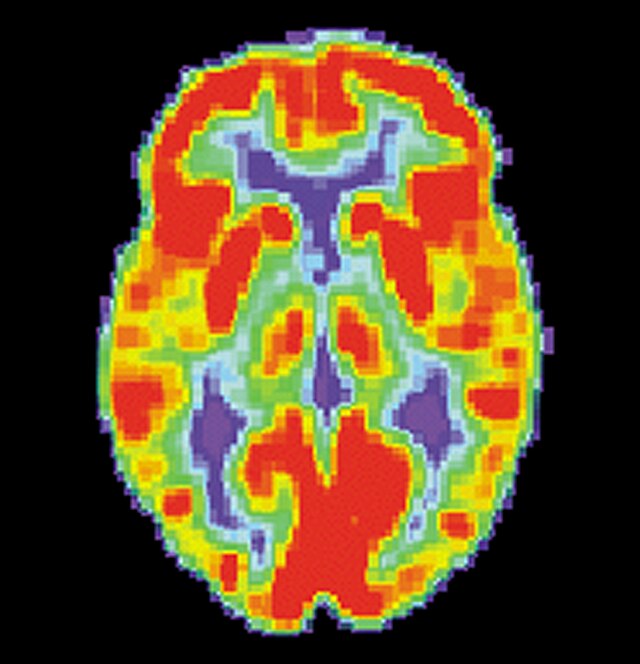}
        \caption{PET scan of the brain\protect\footnotemark}
        \label{fig:pet_brain_2}
    \end{subfigure}
    \hfill
    \begin{subfigure}[b]{0.32\textwidth}
        \centering
        \includegraphics[height=5cm]{Chapter1/Figs/Raster/Computed_tomography_of_human_brain_15.png}
        \caption{CT scan of the brain\protect\footnotemark}
        \label{fig:ct_brain}
    \end{subfigure}
    \hfill
    \begin{subfigure}[b]{0.32\textwidth}
        \centering
        \includegraphics[height=5cm]{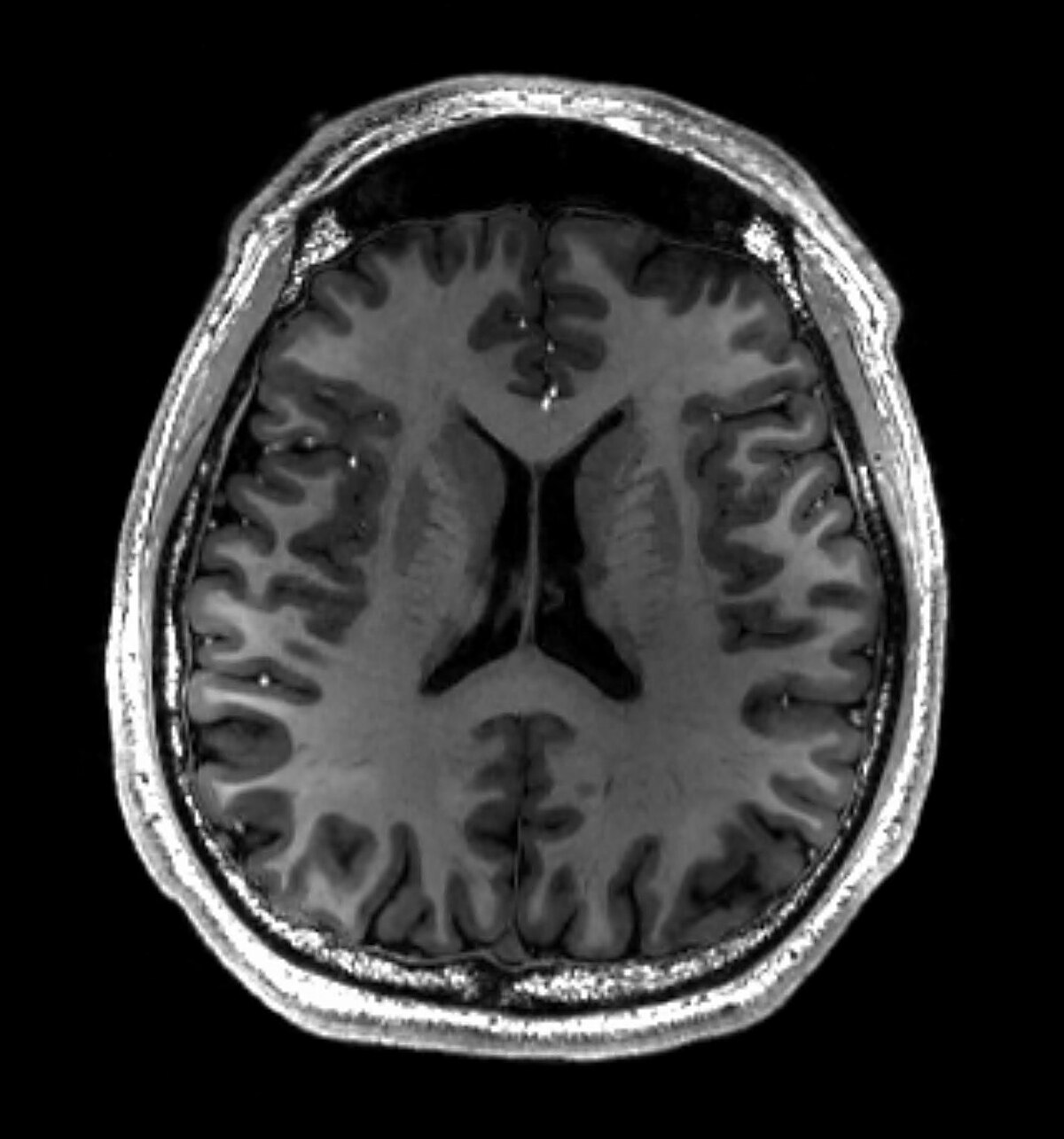}
        \caption{MRI of the brain\protect\footnotemark}
        \label{fig:mri_brain}
    \end{subfigure}
    \caption{Examples of homogeneity in medical imaging modalities illustrating the similarity of the underlying subject (PET, CT, and MRI).}
    \label{fig:medical_modalities_homogeneity}
\end{figure}

\footnotetext{https://en.wikipedia.org/wiki/Brain\_positron\_emission\_tomography}
\footnotetext{https://en.wikipedia.org/wiki/File:Computed\_tomography\_of\_human\_brain\_-\_large.png}
\footnotetext{https://en.wikipedia.org/wiki/Magnetic\_resonance\_imaging\_of\_the\_brain}

Thirdly, while seeming counter to the second point, while in a broad sense medical images are heterogeneous due to the very different imaging techniques and separation based on medical specializations (head scans vs foot scans), within a given data type, there is very high visual similarity.
For example, all chest X-rays will look very similar due to standardized acquisition methods and tools, but also in large part, due to the high homogeneity of the human biology, see figure \ref{fig:medical_modalities_homogeneity}. For the chest scan example, most human bones and organs will have almost identical structures, similar sizes, and composition. Furthermore, the way these scans are collected is standardized, so the patients will all be positioned at the same angle and distance from the imaging machines. This is true even for data samples that have completely different labels, as the characteristics that indicate one diagnosis or another are often identified by very small, granular differences, which are visually very small in absolute terms.
This poses a challenge to researchers in the sense that they can't directly adapt many of the more recent and best-performing classes of Self-Supervised methods, such as contrastive learning. This has allowed us to develop methods that are tailored specifically for medical images.

The overarching motivation for this research is to enhance the utility and efficiency of deep learning in medical imaging. By developing a system that requires less reliance on labeled data, we aim to make deep learning models more accessible and effective for healthcare diagnostics. This approach holds the promise of facilitating the creation of more precise and efficient diagnostic tools, while also respecting the privacy concerns inherent in medical data.

In summary, this thesis is motivated by the desire to make a contribution to the field of medical image analysis. By integrating self-supervised learning strategies with advanced neural network architectures, the goal is to overcome existing barriers and pave the way for breakthroughs in automated medical diagnosis, ultimately enhancing patient care and health outcomes.

\section{Contributions}

The doctoral research encapsulated in this thesis makes several contributions to the field of deep learning, particularly in the application of self-supervised learning for medical image analysis. These contributions are delineated as follows:

\begin{enumerate}
  \item \textbf{Development of the Medformer Architecture}: At the heart of this thesis is the conceptualization and implementation of Medformer, a novel multitask multimodal foundational model architecture. This architecture is designed to process and learn from a broad range of medical image datasets, including those of varied sizes and modalities. A key feature of Medformer is the introduction of Adaptformers, which are instrumental in dynamically adapting the input and output of the neural network to suit different tasks and datasets. This approach allows for a single model to be pre-trained on multiple datasets, extracting general features that enhance performance on specific downstream tasks, thereby addressing the challenges posed by the heterogeneity of medical imaging data. 

  \item \textbf{Advancement in Self-Supervised Learning Methods}: This research extends the frontiers of self-supervised learning within the domain of medical imaging. It demonstrates the feasibility and effectiveness of using self-supervised learning paradigms in training deep neural networks with limited labeled data. This advancement is particularly crucial in addressing the challenge of data scarcity and the dependency on extensive labeled datasets in medical image analysis.
  
  \item \textbf{BrainFuse - Data Fusion Augmentation for Brain MRI}: An advanced technique for augmenting brain MRI datasets by fusing data from different sources. This method enhances the diversity and quality of data available for training deep learning models in medical imaging, particularly in brain age prediction tasks.
  
\end{enumerate}

\section*{Other Contributions}

In addition to the primary focus of this thesis on self-supervised learning for medical images, several contributions have also been made in the general Deep Learning field, many having an indirect impact on the medical field by trying to improve aspects of training these models in a general sense, which would also benefit us when applying them to a medical image context. These contributions, though not directly related to the main theme of the thesis, work to improve the general field of Deep Learning. They include:

\begin{itemize}
  \item \textbf{Cascading Sum Augmentation}: A novel data augmentation procedure that builds upon the success of recent works that propose the combinations of multiple samples in order to generate new data. Building on this idea, \acro{CSA}[Cascading Sum Augmentation] extends the linear combination paradigm to larger groups of examples and deploys a tiered (or “cascading”) training schedule to enhance accuracy on image classification tasks.

  \item \textbf{Backforward Propagation}: A novel approach to neural network training, which involves a unique mechanism of propagating gradients. This method opens new possibilities in the optimization of deep learning algorithms and provides insights into alternative training methodologies.

  \item \textit{Prehospital Cerebrovascular Accident Detection Using AI-Powered Mobile Devices}: A breakthrough in the application of artificial intelligence for early detection of cerebrovascular accidents, contributing to the advancement of mobile healthcare technologies.
  
  \item \textit{Urban Development Prediction Using Satellite Imagery}: An innovative use of deep learning for analyzing satellite imagery to predict urban development patterns, demonstrating the versatility of AI in various domains.
  
  \item \textit{Bird Song Classification with Deep Learning}: A unique application of deep learning in the field of bioacoustics, showcasing the ability to classify and analyze bird songs using advanced neural network models.
  
  \item \textit{Social Media Post Impact Prediction using Computer Vision and Natural Language Processing}: A cutting-edge approach to predicting the impact of social media posts by integrating computer vision and \acro{NLP}, further exemplifying the interdisciplinary nature of AI research.

\end{itemize}

These contributions, while diverse in their application areas, collectively underscore the wide-ranging impact of artificial intelligence and deep learning across various fields.

\section{Thesis Outline}

This section contains the outline of our thesis.

\begin{description}
    \item[Chapter 2: Prerequisites] 
    This chapter lays the groundwork for understanding the concepts and methodologies utilized in this thesis. It covers topics such as Deep Learning, Transformer Architecture, Medical Image Analysis, and Self-Supervised Learning in both general Computer Vision and Medical Image Analysis.

    \item[Chapter 3: Advanced Self-Supervised Learning Techniques in Medical Imaging] This chapter delves into current self-supervised learning methods, proposing novel pretext tasks for medical imaging, and integrating these techniques with medical image analysis for enhanced model performance.

    \item[Chapter 4: Medformer: A Multitask Multimodal Foundational Model for Medical Imaging] Here, we introduce the Medformer architecture, its design, Adaptformers for dynamic input-output adaptation, self-supervised training methodologies, and experimental studies with the MedMNIST dataset.

    \item[Chapter 5: Other Contributions] 
    This chapter encompasses a range of off-topic but significant contributions, including Cascading Sum Augmentation, Backforward Propagation, BrainFuse data augmentation for brain MRI, and various domain-specific contributions such as AI applications in prehospital care, urban development prediction, and more.

    \item[Chapter 6: Conclusion and Future Work] 
    The final chapter summarizes the contributions of the thesis, discusses the implications for medical image analysis, and outlines future research directions.
\end{description}

\chapter{Prerequisites}
\label{chap:prereqs}
\ifpdf
    \graphicspath{{Chapter2/Figs/Raster/}{Chapter2/Figs/PDF/}{Chapter2/Figs/}}
\else
    \graphicspath{{Chapter2/Figs/Vector/}{Chapter2/Figs/}}
\fi

Over the past decade, advancements in data acquisition, storage, and high-performance computation have fundamentally transformed both scientific inquiry and industrial practice. Innovations in sensor hardware and imaging techniques—ranging from high-resolution medical scanners to ubiquitous mobile devices—have rapidly accumulated heterogeneous datasets. These developments have not only spurred progress in traditional computer vision but have also catalyzed new paradigms in deep learning, where the focus increasingly shifts toward learning rich representations from vast amounts of unlabeled data.

In the context of medical imaging, the demand for accurate and efficient image interpretation is paramount. Clinical decision-making often hinges on the precise detection of subtle pathological features such as small lesions or early signs of tissue atrophy. However, the creation of large-scale, expert-annotated datasets is frequently constrained by high costs, limited availability of expert time, and strict privacy requirements. To address these challenges, modern deep learning research has increasingly embraced \acro{SSL} techniques that construct synthetic or automatically derived pretext tasks. These tasks enable models to learn underlying feature representations directly from unlabeled data, thereby reducing reliance on manual annotation and facilitating more robust performance in downstream clinical applications.

This thesis is built upon a body of work that spans both general deep learning and its specialized application in medical image analysis. Our prior contributions—including Efficient Dynamic Batch Adaptation, Backforward Propagation, and novel data fusion augmentation methods such as BrainFuse and Cascading Sum Augmentation—exemplify strategies designed to overcome the inherent challenges in training deep neural networks on limited medical datasets. For instance, Efficient Dynamic Batch Adaptation addresses optimization challenges by dynamically adjusting the composition and size of training batches, thereby mitigating issues such as internal covariate shift and gradient instability. Similarly, our work on Backforward Propagation has provided new insights into refining the weight update process to improve convergence in the presence of complex, high-dimensional medical data.

Medical images, unlike many natural images, often exhibit intricate anatomical structures across multiple scales. Modalities such as \acro{MRI}, \acro{CT}, \acro{PET}, and X-ray each capture different facets of human anatomy or physiology. In addition, critical diagnostic features can be extremely subtle, necessitating models that are not only powerful but also sensitive to fine-grained variations. These factors contribute to the complexity and heterogeneity of medical datasets, further compounded by privacy concerns that limit data sharing and labeling. The confluence of these issues underscores the need for robust learning algorithms capable of leveraging the wealth of unlabeled data available, a need that self-supervised learning techniques are uniquely positioned to address.

The evolution of self-supervised learning in recent years has given rise to several families of algorithms. Techniques such as predictive pretext tasks—where models learn to reconstruct missing image regions or predict geometric transformations—enable the extraction of features that capture both local texture and global structure. Contrastive methods, information maximization approaches, and multi-task frameworks have all been explored as means to learn discriminative representations in data-scarce scenarios. In medical imaging, where the availability of labeled data is limited and the structures of interest are both complex and subtle, these approaches provide a valuable alternative to conventional supervised learning.

Furthermore, our research has emphasized the importance of efficient training methodologies. For example, our work on Efficient Dynamic Batch Adaptation has shown that dynamically modifying the training batch composition can lead to significant improvements in convergence speed and generalization performance. Similarly, Backforward Propagation offers a novel perspective on mitigating internal covariate shift by recalculating gradient updates in a manner that better accounts for inter-layer distributional changes. In parallel, our data augmentation methods—such as the cascading sum augmentation used for satellite imagery and the BrainFuse technique for MRI data fusion—demonstrate how carefully designed augmentation strategies can expand the effective training dataset and improve model robustness.

In summary, this chapter presents the foundational concepts and technological advances that underpin the research presented in this thesis. We review the evolution of deep learning from early neural network models to modern self-supervised frameworks and discuss the specific challenges posed by medical image analysis, including issues of scale, subtlety, heterogeneity, and privacy. By integrating ideas from our previous contributions and the broader literature, we establish a comprehensive set of prerequisites that set the stage for the advanced techniques and experimental studies described in subsequent chapters. The discussion that follows not only motivates the use of large-scale unlabeled data but also provides the theoretical and practical basis for the innovative approaches developed in this work.

\section{Deep Learning}
Deep learning has redefined how machine learning systems learn representations from large-scale datasets \cite{lecun1998gradient}. Unlike traditional machine learning, which relies heavily on handcrafted features, deep learning methods attempt to learn multi-level, hierarchical features directly from data. This approach has led to substantial improvements in tasks such as image classification \cite{he2016deep}, object detection, speech recognition \cite{graves2013speech, amodei2016deepspeech}, and natural language understanding \cite{ruder2016sgdoverview, brown2020language}.

At its core, machine learning can be broadly partitioned into supervised and unsupervised paradigms. In the supervised framework, models are trained on well-curated input-output pairs; the algorithm is explicitly taught to map inputs to corresponding outputs. This approach has yielded significant successes in applications such as image classification, semantic segmentation, and object detection, particularly when vast repositories of annotated data are available. However, the acquisition of high-quality labeled data in medical contexts is frequently hindered by the expense and time required for expert annotation, as well as by stringent privacy constraints.

To circumvent these limitations, unsupervised and, more specifically, self-supervised learning techniques have emerged as powerful alternatives. Self-supervised learning capitalizes on the latent structure inherent in the data by formulating pretext tasks that generate supervisory signals without the need for manual labeling. For instance, a model may be challenged to reconstruct missing regions of an image, predict the correct sequence of shuffled patches, or infer the geometric transformations applied to its input. Such tasks compel the network to internalize the underlying semantics and spatial relationships within the data, thereby yielding robust feature representations that can be subsequently fine-tuned with minimal annotated examples.

Our recent research, as reflected in projects such as BrainFuse and our work on efficient dynamic batch adaptation and backforward propagation, exemplifies the profound impact of self-supervised approaches in the realm of medical image analysis. In these endeavors, models are designed not only to extract domain-specific features from modalities such as magnetic resonance imaging (MRI) but also to leverage auxiliary tasks—such as frame interpolation for generating smooth transitions between disparate brain regions—to enhance data diversity and representation quality. This methodology enables a more nuanced understanding of anatomical structures and pathological variations, even when faced with limited labeled data.

Moreover, advancements in deep learning architectures—ranging from residual networks to transformer-based models—have been integrated with these self-supervised strategies to further boost performance. Techniques such as adaptive input-output modules (Adaptformers) and novel data augmentation protocols like Cascading Sum Augmentation illustrate how architectural innovation, when coupled with self-supervision, can address challenges of data scarcity and heterogeneity. These approaches facilitate multitask and multimodal learning, ensuring that models are not only accurate but also generalize well across varying clinical conditions.

In summary, deep learning provides a versatile framework for the automatic discovery of latent representations, and its evolution—from early neural networks to contemporary self-supervised and dynamic training methods—has significantly influenced the field of medical imaging. By synthesizing these diverse strategies, our work demonstrates that it is possible to develop intelligent systems capable of achieving high diagnostic performance while mitigating the limitations imposed by scarce and heterogeneous data. This integrated perspective lays a robust foundation for subsequent chapters that detail domain adaptation, multimodal fusion, and privacy-preserving strategies in medical image analysis.

\subsection{Artificial Neural Networks}

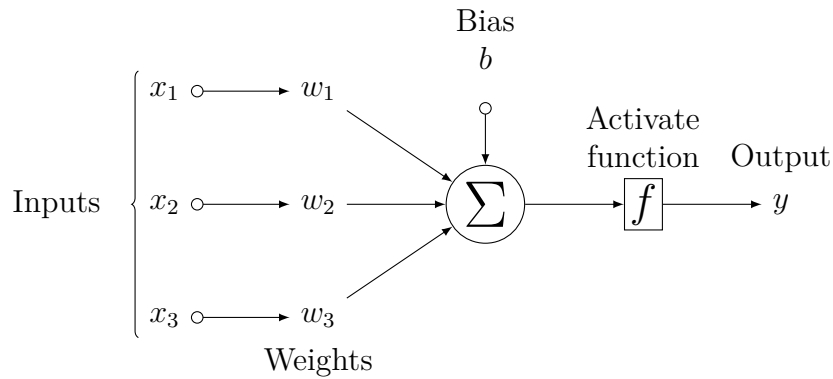
\begin{figure}[ht]
    \centering
    \begin{tikzpicture}[
        init/.style={
          draw,
          circle,
          inner sep=2pt,
          font=\Huge,
          join = by -latex
        },
        squa/.style={
          draw,
          inner sep=2pt,
          font=\Large,
          join = by -latex
        },
        start chain=2,node distance=13mm,align=center]
        \node[on chain=2] 
          (x2) {$x_2$};
        \node[on chain=2,join=by o-latex] 
          {$w_2$};
        \node[on chain=2,init] (sigma) 
          {$\displaystyle\Sigma$};
        \node[on chain=2,squa,label=above:{\parbox{2cm}{\centering Activate \\ function}}]   
          {$f$};
        \node[on chain=2,label=above:Output,join=by -latex] 
          {$y$};
        \begin{scope}[start chain=1]
        \node[on chain=1] at (0,1.5cm) 
          (x1) {$x_1$};
        \node[on chain=1,join=by o-latex] 
          (w1) {$w_1$};
        \end{scope}
        \begin{scope}[start chain=3]
        \node[on chain=3] at (0,-1.5cm) 
          (x3) {$x_3$};
        \node[on chain=3,label=below:Weights,join=by o-latex] 
          (w3) {$w_3$};
        \end{scope}
        \node[label=above:\parbox{2cm}{\centering Bias \\ $b$}] at (sigma|-w1) (b) {};
        
        \draw[-latex] (w1) -- (sigma);
        \draw[-latex] (w3) -- (sigma);
        \draw[o-latex] (b) -- (sigma);
        
        \draw[decorate,decoration={brace,mirror}] (x1.north west) -- node[left=10pt] {Inputs} (x3.south west);
    \end{tikzpicture}
    \caption{Neuron structure \label{fig:neuron}}
\end{figure}

\begin{figure}[ht]
    \centering
    \begin{tikzpicture}[
        plain/.style={
          draw=none,
          fill=none,
          },
        net/.style={
          matrix of nodes,
          nodes={
            draw,
            circle,
            inner sep=10pt
            },
          nodes in empty cells,
          column sep=2cm,
          row sep=-9pt
          },
        >=latex
        ,align=center]
        \matrix[net] (mat)
        {
        |[plain]| \parbox{1.3cm}{\centering Input\\layer} & |[plain]| \parbox{1.3cm}{\centering Hidden\\layer} & |[plain]| \parbox{1.3cm}{\centering Output\\layer} \\
        & |[plain]| \\
        |[plain]| & \\
        & |[plain]| \\
          |[plain]| & |[plain]| \\
        & & \\
          |[plain]| & |[plain]| \\
        & |[plain]| \\
          |[plain]| & \\
        & |[plain]| \\    };
        \foreach \ai [count=\mi ]in {2,4,...,10}
          \draw[<-] (mat-\ai-1) -- node[above] {Input \mi} +(-2cm,0);
        \foreach \ai in {2,4,...,10}
        {\foreach \aii in {3,6,9}
          \draw[->] (mat-\ai-1) -- (mat-\aii-2);
        }
        \foreach \ai in {3,6,9}
          \draw[->] (mat-\ai-2) -- (mat-6-3);
        \draw[->] (mat-6-3) -- node[above] {Ouput} +(2cm,0);
    \end{tikzpicture}
    \caption{Multi-layered perceptron \label{fig:MLP}}
\end{figure}
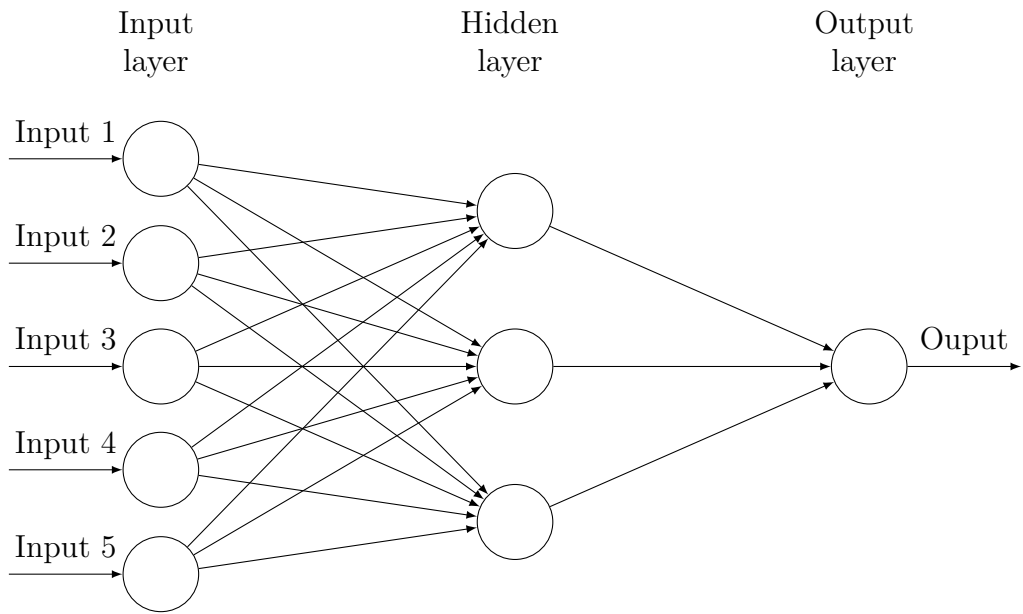

\acro{ANN}[Artificial neural networks] represent one of the most pivotal paradigms in modern machine learning, drawing inspiration from the complex information processing mechanisms observed in biological neural systems. Fundamentally, ANNs are composed of layers of interconnected processing units—commonly referred to as neurons—which collectively enable the modeling of intricate, non-linear relationships within high-dimensional data. In its simplest form, each neuron performs a linear transformation of an input vector \(x\) by computing a weighted sum augmented by a bias term, and then applies a non-linear activation function \(\sigma\) to produce an output, as formalized by
\[
y = \sigma(w^T x + b),
\]
where \(w\) is the vector of learnable weights, \(b\) is the bias term, and \(\sigma\) may be instantiated by functions such as the sigmoid, hyperbolic tangent, or \acro{ReLU}[Rectified Linear Unit]. This basic computational unit is illustrated in Figure~\ref{fig:neuron}. When neurons are arranged in sequential layers, the multilayered architecture—exemplified by the \acro{MLP}[Multi-Layered Perceptron] shown in Figure~\ref{fig:MLP}—facilitates the progressive abstraction of features, thereby enabling the network to learn hierarchical representations that are particularly advantageous in applications ranging from medical imaging to natural language processing.

Training these networks involves the systematic adjustment of the network’s parameters through gradient-based optimization methods. Techniques such as stochastic gradient descent (SGD) and its numerous adaptive variants are employed to minimize a suitably defined loss function, which quantifies the divergence between the network’s predictions and the ground-truth labels. The backpropagation algorithm, which leverages the chain rule of calculus, plays an indispensable role in efficiently propagating error gradients from the output layer back through the network. Over iterative training cycles, this process enables the network to refine its internal representations and to progressively converge towards a mapping that captures the underlying statistical structure of the data.

Early neural networks were shallow, making them susceptible to underfitting and limited representational capacity. Additionally, these networks often encountered difficulties such as vanishing or exploding gradients, hindering deeper architectures \cite{hochreiter1997long}. Modern practices like residual connections \cite{he2016identity} and careful weight initialization \cite{sutskever2013importance} have helped mitigate these issues, enabling the successful training of deep networks.

Moreover, recent advancements in self-supervised and multitask learning have further enriched the conceptual framework of ANNs. By integrating unsupervised pretext tasks and leveraging large-scale unlabeled data, contemporary training protocols not only mitigate the challenges associated with limited annotated datasets but also promote the extraction of robust, transferable features. These developments underscore the versatility of ANNs, which continue to evolve as the foundational substrates for a broad spectrum of deep learning models, including those tailored for domain adaptation and multimodal analysis.

Artificial neural networks serve as the theoretical and practical foundation for modern deep learning systems. Their ability to convert raw inputs into meaningful, hierarchical representations is central to the success of numerous applications, and ongoing innovations in training methodologies and architectural design continue to expand their capabilities. The rich interplay between classical network design and recent advances in adaptive and self-supervised learning highlights the enduring significance of ANNs in advancing both the theory and practice of artificial intelligence.

\subsection{Recurrent Neural Networks}

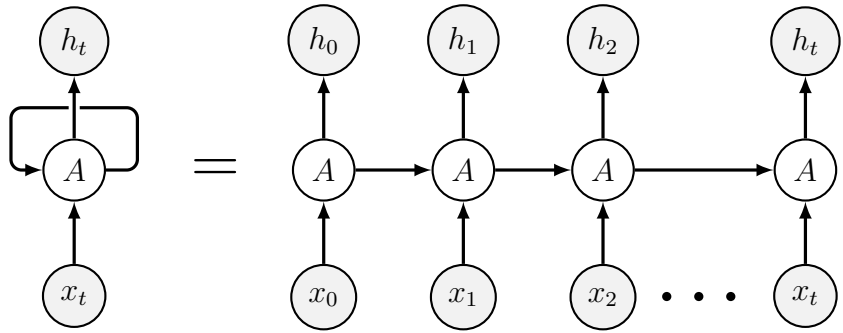
\begin{figure}[ht]
    \centering
    \begin{tikzpicture}[item/.style={circle,draw,thick,fill=none},
        itemc/.style={item,on chain,join}]
         \begin{scope}[start chain=going right,nodes=itemc,every
         join/.style={-latex,very thick},local bounding box=chain]
         \path node (A0) {$A$} node (A1) {$A$} node (A2) {$A$} node[xshift=2em] (At)
         {$A$};
         \end{scope}
         \node[left=1em of chain,scale=2] (eq) {$=$};
         \node[left=2em of eq,item] (AL) {$A$};
         \path (AL.west) ++ (-1em,2em) coordinate (aux);
         \draw[very thick,-latex,rounded corners] (AL.east) -| ++ (1em,2em) -- (aux) 
         |- (AL.west);
         \foreach \X in {0,1,2,t} 
         {\draw[very thick,-latex] (A\X.north) -- ++ (0,2em)
         node[above,item,fill=gray!10] (h\X) {$h_\X$};
         \draw[very thick,latex-] (A\X.south) -- ++ (0,-2em)
         node[below,item,fill=gray!10] (x\X) {$x_\X$};}
         \draw[white,line width=0.8ex] (AL.north) -- ++ (0,1.9em);
         \draw[very thick,-latex] (AL.north) -- ++ (0,2em)
         node[above,item,fill=gray!10] {$h_t$};
         \draw[very thick,latex-] (AL.south) -- ++ (0,-2em)
         node[below,item,fill=gray!10] {$x_t$};
         \path (x2) -- (xt) node[midway,scale=2,font=\bfseries] {\dots};
    \end{tikzpicture}
    \caption{Recurrent neural structure \label{fig:RNN}}
\end{figure}

Recurrent neural networks (RNNs) constitute a class of neural architectures that are intrinsically designed to process sequential data by maintaining an internal state, which acts as a dynamic memory encoding information from prior inputs. This sequential processing capability renders RNNs particularly suitable for tasks in which temporal dependencies are critical, such as language modeling, time-series forecasting, and the analysis of sequential medical data. Figure~\ref{fig:RNN} presents an unrolled depiction of a simple RNN, wherein the network processes a sequence of inputs in a time‐step‐wise manner while continuously updating its hidden state to reflect accumulated contextual information.

Over the years, recurrent architectures have evolved significantly. Early RNN models were hampered by issues such as vanishing and exploding gradients, phenomena that made training on long sequences challenging. In response to these difficulties, advanced variants—most notably \acro{LSTM}[Long Short-Term Memory] networks~\cite{hochreiter1997long} and Gated Recurrent Units (GRUs)—were introduced. These architectures incorporate specialized gating mechanisms that regulate the flow of information, thereby mitigating gradient decay and enabling the network to capture long-range dependencies. In our own work, concepts analogous to these gating mechanisms have been extended further through techniques such as Backforward Propagation, which recalculates weight updates to counteract internal covariate shifts, and Efficient Dynamic Batch Adaptation, which dynamically adjusts training batches to preserve the integrity of gradient information during sequential processing.

Notably, while the iterative nature of RNNs enables a natural treatment of sequential data, it also imposes limitations in terms of computational scalability when handling very large datasets. This intrinsic sequentiality contrasts with more recently developed architectures, such as Transformers, which offer parallel processing of sequence elements and have gained prominence in various domains. Nevertheless, the foundational principle underlying RNNs—the ability to accumulate and integrate information over time—remains indispensable, particularly in medical applications where temporal or sequential patterns (for instance, in patient monitoring or in longitudinal imaging studies) carry critical diagnostic and prognostic significance.

In recent research, including our prior investigations into self-supervised learning for medical imaging, we have demonstrated that integrating self-supervised pretext tasks within recurrent frameworks can yield representations that capture both local and global temporal structures. For example, by applying pretext tasks that involve predicting future frames or reconstructing missing segments in sequential medical scans, it becomes possible to refine the hidden state dynamics of RNNs. These modifications enhance the network's sensitivity to subtle variations in sequential data, which is paramount when addressing challenges such as gradual anatomical changes or the temporal progression of disease. Furthermore, multi-task and multimodal self-supervised strategies—as exemplified in our work on Medformer—have underscored the utility of combining recurrent models with dynamic adaptation mechanisms. Such approaches not only improve convergence and stability but also allow the model to learn transferable representations that generalize across different medical imaging modalities.

While contemporary trends in deep learning have increasingly favored parallelizable architectures for many large-scale tasks, recurrent neural networks continue to offer unique advantages in modeling temporal dynamics. Their inherent design—enhanced by modern training techniques and adaptive mechanisms—enables the extraction of nuanced sequential features, thereby contributing significantly to the analysis of time-dependent and volumetric medical data. The continued refinement of RNN training methods, including the incorporation of gradient correction strategies and dynamic batch adaptation, affirms their enduring relevance in both theoretical and applied research contexts.

\begin{figure}[ht]
    \centering
    \begin{tikzpicture}[
    font=\sf \scriptsize,
    >=LaTeX,
    cell/.style={
        rectangle, 
        rounded corners=5mm, 
        draw,
        very thick,
        },
    operator/.style={
        circle,
        draw,
        inner sep=-0.5pt,
        minimum height =.2cm,
        },
    function/.style={
        ellipse,
        draw,
        inner sep=1pt
        },
    ct/.style={
        circle,
        draw,
        line width = .75pt,
        minimum width=1cm,
        inner sep=1pt,
        },
    gt/.style={
        rectangle,
        draw,
        minimum width=4mm,
        minimum height=3mm,
        inner sep=1pt
        },
    mylabel/.style={
        font=\scriptsize\sffamily
        },
    ArrowC1/.style={
        rounded corners=.25cm,
        thick,
        },
    ArrowC2/.style={
        rounded corners=.5cm,
        thick,
        },
    ]

    \node [cell, minimum height =4cm, minimum width=6cm] at (0,0){} ;

    \node [gt] (ibox1) at (-2,-0.75) {$\sigma$};
    \node [gt] (ibox2) at (-1.5,-0.75) {$\sigma$};
    \node [gt, minimum width=1cm] (ibox3) at (-0.5,-0.75) {Tanh};
    \node [gt] (ibox4) at (0.5,-0.75) {$\sigma$};

    \node [operator] (mux1) at (-2,1.5) {$\times$};
    \node [operator] (add1) at (-0.5,1.5) {+};
    \node [operator] (mux2) at (-0.5,0) {$\times$};
    \node [operator] (mux3) at (1.5,0) {$\times$};
    \node [function] (func1) at (1.5,0.75) {Tanh};

    \node[ct, label={Cell}] (c) at (-4,1.5) {\empty{c}{t-1}};
    \node[ct, label={Hidden}] (h) at (-4,-1.5) {\empty{h}{t-1}};
    \node[ct, label={left:Input}] (x) at (-2.5,-3) {\empty{x}{t}};

    \node[ct, label={Cell}] (c2) at (4,1.5) {\empty{c}{t}};
    \node[ct, label={Hidden}] (h2) at (4,-1.5) {\empty{h}{t}};
    \node[ct, label={left:Output}] (x2) at (2.5,3) {\empty{h}{t}};

    \draw [ArrowC1] (c) -- (mux1) -- (add1) -- (c2);

    \draw [ArrowC2] (h) -| (ibox4);
    \draw [ArrowC1] (h -| ibox1)++(-0.5,0) -| (ibox1); 
    \draw [ArrowC1] (h -| ibox2)++(-0.5,0) -| (ibox2);
    \draw [ArrowC1] (h -| ibox3)++(-0.5,0) -| (ibox3);
    \draw [ArrowC1] (x) -- (x |- h)-| (ibox3);

    \draw [->, ArrowC2] (ibox1) -- (mux1);
    \draw [->, ArrowC2] (ibox2) |- (mux2);
    \draw [->, ArrowC2] (ibox3) -- (mux2);
    \draw [->, ArrowC2] (ibox4) |- (mux3);
    \draw [->, ArrowC2] (mux2) -- (add1);
    \draw [->, ArrowC1] (add1 -| func1)++(-0.5,0) -| (func1);
    \draw [->, ArrowC2] (func1) -- (mux3);

    \draw [-, ArrowC2] (mux3) |- (h2);
    \draw (c2 -| x2) ++(0,-0.1) coordinate (i1);
    \draw [-, ArrowC2] (h2 -| x2)++(-0.5,0) -| (i1);
    \draw [-, ArrowC2] (i1)++(0,0.2) -- (x2);

\end{tikzpicture}
    \caption{LSTM cell \label{fig:LSTM}}
\end{figure}
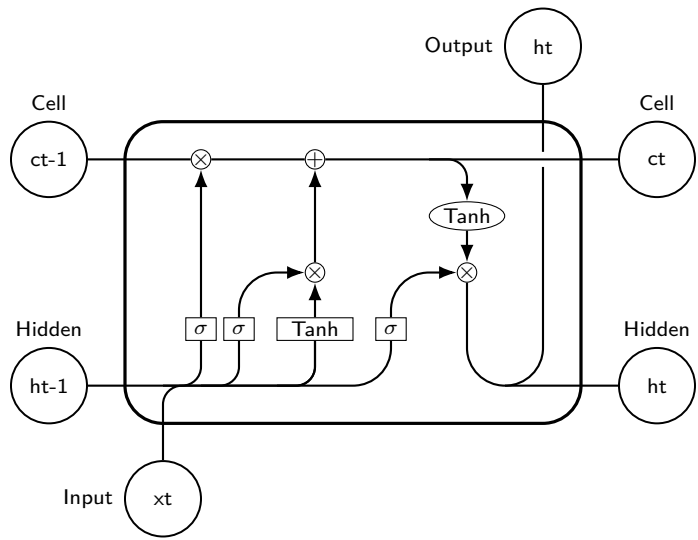

\subsection{Convolutional Neural Networks}

\acro{CNN}[Convolutional neural networks] have established themselves as a foundational framework for image analysis, primarily due to their intrinsic adept at processing grid-like topologies of data, such as images \cite{krizhevsky2017imagenet, he2016deep}. By employing convolutional filters that are shared across different regions of an input image, CNNs are able to capture local patterns effectively without the burden of learning distinct weights for each spatial location (see Figure~\ref{fig:convolutional_layer}). This mechanism of weight sharing not only confers robust translation invariance but also significantly reduces the overall number of parameters, thereby mitigating overfitting—a crucial advantage when operating under conditions of limited annotated data, as frequently encountered in medical imaging.

\begin{figure}[ht]
    \centering
    \begin{tikzpicture}[
        font=\sf\scriptsize,
        >=LaTeX,
        cell/.style={rectangle, rounded corners=5mm, draw, very thick, minimum width=2cm, minimum height=1cm},
        operator/.style={circle, draw, inner sep=-0.5pt, minimum size=1cm},
        ArrowC1/.style={rounded corners=.25cm, thick}
    ]
        \node[cell] (input) at (0,0) {Input};
        \node[cell] (conv1) at (3,0) {Conv1};
        \node[cell] (conv2) at (6,0) {Conv2};
        \node[operator] (add) at (9,0) {+};
        \node[cell] (output) at (12,0) {Output};
        
        \draw[->, ArrowC1] (input) -- (conv1);
        \draw[->, ArrowC1] (conv1) -- (conv2);
        \draw[->, ArrowC1] (conv2) -- (add);
        \draw[->, ArrowC1] (add) -- (output);
        
        \draw[->, ArrowC1, dashed] (input.north) to[bend left=45] (add.north);
    \end{tikzpicture}
    \caption{Residual block with skip connection \label{fig:skip_conn}}
\end{figure}
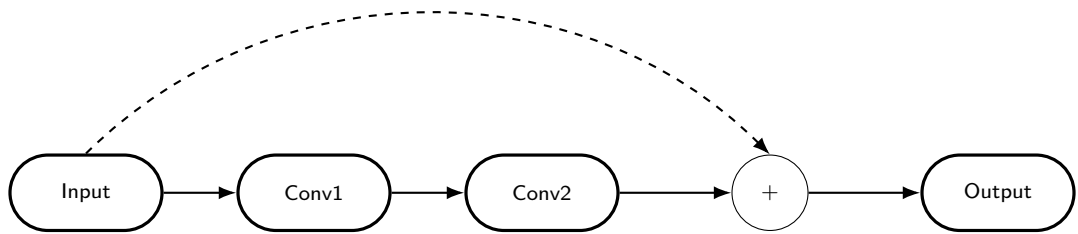

Over the years, as CNN architectures have evolved toward increased depth and complexity, innovative solutions such as residual connections~\cite{he2016deep} have been introduced to counteract optimization challenges such as the vanishing gradient problem. Residual blocks enable networks to learn modifications to an identity mapping by reintroducing the input signal after a series of transformations, as depicted in Figure~\ref{fig:skip_conn}. This design not only facilitates smoother gradient flow during training but also stabilizes the convergence of very deep networks, a quality that has proven essential in both conventional and specialized applications.

In the realm of medical image analysis, the hierarchical feature extraction capabilities of CNNs assume even greater significance. These networks progressively abstract complex anatomical structures from raw pixel data, thereby enabling the detection of subtle anomalies, accurate organ segmentation, and reliable diagnostic predictions. Recent research has extended the traditional CNN framework by integrating it with self-supervised learning paradigms and advanced data augmentation techniques—such as cascading sum augmentation and efficient dynamic batch adaptation—to address the challenges posed by data scarcity and inter-domain variability. Such integrations allow CNNs to serve as the backbone for multitask and multimodal models that not only capture clinically relevant features but also adapt to the diverse characteristics of medical imaging data.

Moreover, the flexibility of CNNs in learning hierarchical representations makes them well suited for domain adaptation tasks, where models pre-trained on large natural image datasets are fine-tuned for specific medical applications. This capability has been further enhanced through the incorporation of residual learning and normalization strategies, ensuring that the learned features remain robust despite the intrinsic heterogeneity and limited volume of medical datasets.

\begin{figure}[ht]
    \centering
    \begin{tikzpicture}
		\node at (1.5,4){\begin{tabular}{c}input image\\or input feature map\end{tabular}};
	
		\draw (0,0) -- (3,0) -- (3,3) -- (0,3) -- (0,0);
		
		\draw (2,2) -- (2.5,2) -- (2.5,2.5) -- (2,2.5) -- (2,2);
		\draw (2,0.5) -- (2.5,0.5) -- (2.5,1) -- (2,1) -- (2,0.5);
		\draw (1,1) -- (1.5,1) -- (1.5,1.5) -- (1,1.5) -- (1,1);
		
		\draw (2.5,2) -- (7,3.25);
		\draw (2.5,2.5) -- (7,3.25);
 
		\draw (2.5,1) -- (5.75,0.25);
		\draw (2.5,0.5) -- (5.75,0.25);
		
		\draw (1.5,1.5) -- (5.5,1.25);
		\draw (1.5,1) -- (5.5,1.25);
		
		\node at (5.75,4){\begin{tabular}{c}output feature maps\end{tabular}};
		
		\draw[fill=black,opacity=0.2,draw=black] (5.5,1.5) -- (7.5,1.5) -- (7.5,3.5) -- (5.5,3.5) -- (5.5,1.5);
		\draw[fill=black,opacity=0.2,draw=black] (5,1) -- (7,1) -- (7,3) -- (5,3) -- (5,1);
		\draw[fill=black,opacity=0.2,draw=black] (4.5,0.5) -- (6.5,0.5) -- (6.5,2.5) -- (4.5,2.5) -- (4.5,0.5);
		\draw[fill=black,opacity=0.2,draw=black] (4,0) -- (6,0) -- (6,2) -- (4,2) -- (4,0);
	\end{tikzpicture}
	\caption{Illustration of a convolutional layer}
	\label{fig:convolutional_layer}
\end{figure}
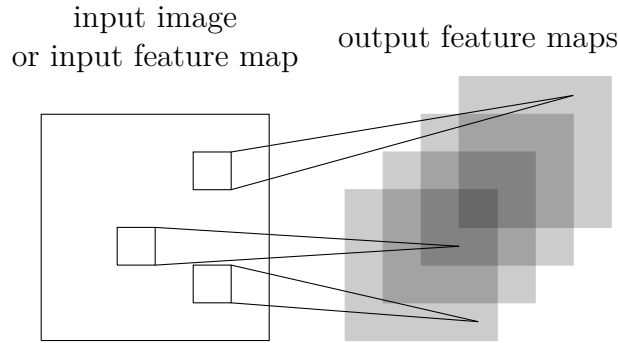

In summary, convolutional neural networks have transformed the field of image analysis by providing a scalable and efficient means of extracting complex visual features. Their ability to abstract hierarchical representations, combined with architectural innovations such as residual connections and integration with self-supervised methodologies, has proven indispensable for advancing medical imaging. These advances not only improve diagnostic accuracy and model generalization but also pave the way for future research into more adaptive and robust systems tailored to the unique challenges of clinical applications.

\subsection{Transformer Architecture}

Transformers constitute a class of models initially devised for natural language processing but increasingly adapted to computer vision and other domains. They rely on a self-attention mechanism, which was proposed by Vaswani et al.~\cite{vaswani2017attention}, to capture dependencies among tokens or elements in a sequence. Unlike recurrent networks, which process data sequentially and may suffer from difficulties when modeling long-range relationships, Transformers can attend to all positions in a sequence in parallel, thus handling more global interactions with relative efficiency. This structural design has significantly advanced the state of the art in tasks involving structured data, whether textual or visual.

Key to the Transformer model is the concept of attention, wherein each element in a sequence is transformed into query, key, and value representations. By comparing queries and keys, the model assigns weights to each position in the sequence and subsequently reweights the corresponding values based on these weights. This process is carried out by multiple attention heads, each of which captures a potentially different aspect of the interrelationships among sequence elements. A crucial advantage of this architecture lies in its flexibility: it does not impose strict assumptions about the length of the input, and it easily scales to large sequences provided sufficient memory resources.

\begin{figure}[ht]
    \centering
    \includegraphics[width=1.0\linewidth]{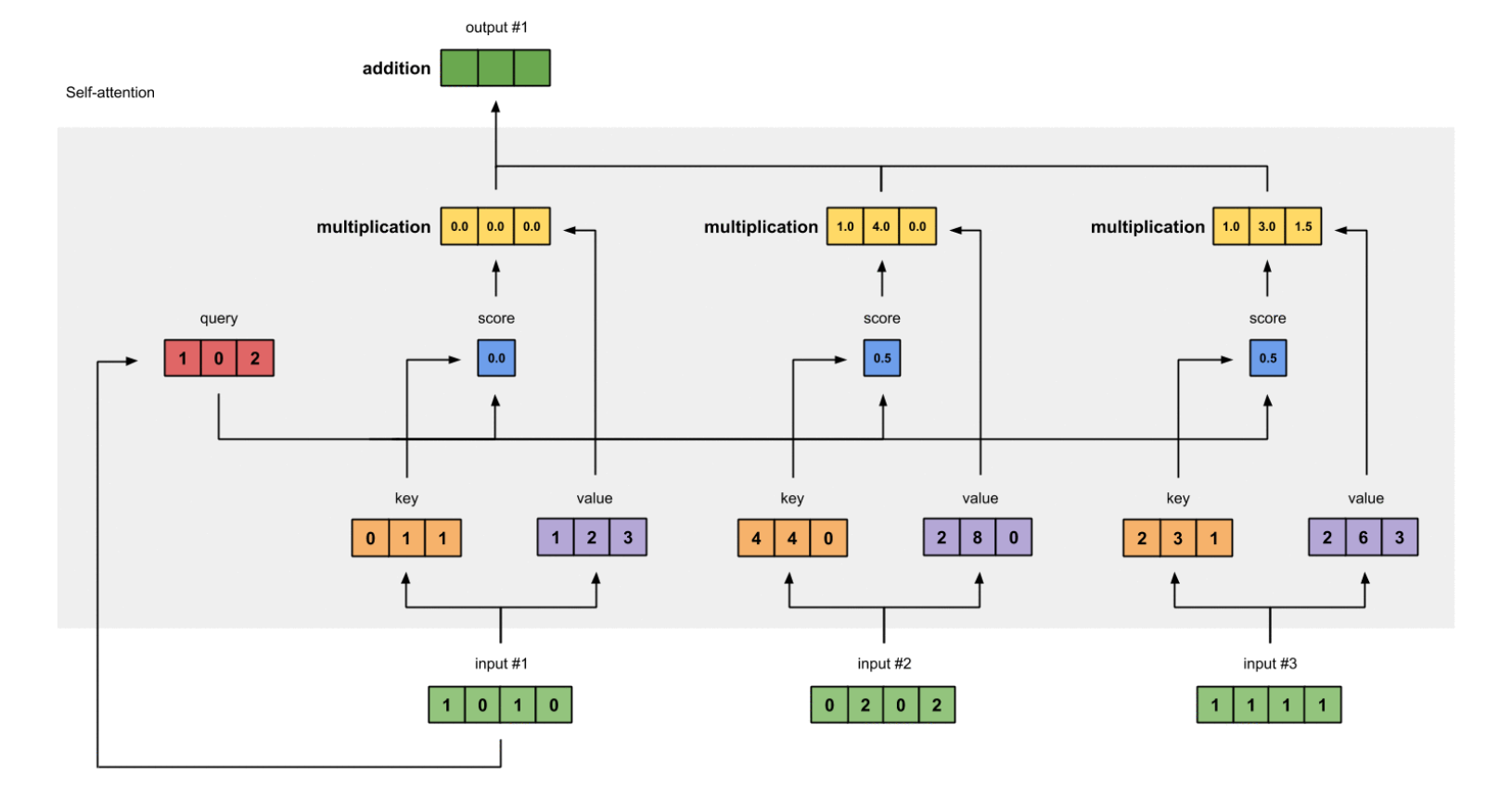}
    \caption{Self-attention mechanism\protect\footnotemark\label{fig:self_attention}}
\end{figure}

However, this increase in context understanding comes at a high computational cost. The fundamental operation in a self-attention module is typically computed as:
\[
    \text{Attention}(Q, K, V) = \text{softmax}
   \begin{pmatrix}
    \frac{QK^T}{\sqrt{d_k}}
    \end{pmatrix}
    V,
\]
where the query, key, and value vectors (depicted in Figure~\ref{fig:self_attention}) are obtained by passing the input through separate linear layers, usually denoted by $W^Q, W^K, W^V$. The term $d_k$ is the dimension of these vectors and is used to scale the dot product so that the resulting score maintains a mean of 0 and a variance of 1. Although this attention mechanism excels at capturing long-range dependencies, it scales poorly for large numbers of queries and keys. Consequently, memory usage and compute requirements can become prohibitive for high-resolution inputs or very long sequences. To address this, multiple approaches that approximate or optimize the attention mechanism have been proposed~\cite{fedus2021switch,choromanski2020rethinking}.

Attention-based models, while originally popularized in NLP, have begun to permeate imaging tasks due to their powerful contextual reasoning. Although convolutional neural networks remain the dominant architecture for image analysis, attention mechanisms enable models to dynamically consider the relationships between different regions of an image. As described earlier, the self-attention mechanism allows each part of the input to weight the importance of other parts, capturing context that is essential in understanding complex structures. This process is especially beneficial for sequenced data and can be visualized in the self-attention diagram in Figure~\ref{fig:self_attention}.

When applied to images, Transformers can interpret patches of an image as tokens, thus extending their applicability beyond language. The Vision Transformer (ViT)~\cite{dosovitskiy2020image} showcased how standard Transformers can operate directly on non-overlapping image patches, flattening them and adding positional encodings to preserve spatial relationships. Despite this approach initially seeming less intuitive than convolutions, experiments revealed competitive results with a sufficiently large dataset and careful hyperparameter tuning. Further refinements, such as DeiT~\cite{touvron2021training}, achieved high performance using fewer training samples, indicating that Transformers can excel in image tasks once certain training strategies are employed.

\footnotetext{Source: https://miro.medium.com/max/3000/1*\_92bnsMJy8Bl539G4v93yg.gif}

In medical imaging, Transformers have been adapted for segmentation, classification, and other tasks that require modeling complex, context-dependent structures. Hybrid architectures often combine convolutional layers for initial feature extraction with Transformer-based modules for enhancing global contextual understanding. This strategy balances the local feature extraction strengths of convolutional neural networks (CNNs) with the Transformer’s capability to capture long-distance relationships. Techniques such as TransUNet ~\cite{chen2021transunet}, for example, have integrated a Transformer layer into the well-known U-Net architecture. The resulting hybrid model demonstrated improved performance in segmentation tasks on medical datasets like multi-organ CT scans.

Nevertheless, employing Transformers in medical imaging still poses practical considerations. First, Transformers can be memory-intensive, a challenge that is amplified when dealing with volumetric scans. Even with two-dimensional images, patch embedding in a Transformer backbone requires considerable GPU memory if the images are high resolution. Second, Transformers typically rely on large training sets to fully learn their many parameters. Obtaining labeled medical images in sufficient quantities can be difficult, prompting researchers to turn toward self-supervised or semi-supervised methods to alleviate this requirement for massive annotated data. Finally, in clinical settings, interpretability remains a priority. While self-attention maps can offer insights into the regions a Transformer-based model deems important, the interpretability question remains open. Addressing this issue is crucial if these methods are to see widespread acceptance in sensitive domains like healthcare.

The Transformer architecture, with its self-attention mechanism as illustrated in Figure~\ref{fig:self_attention}, reshapes how models capture global dependencies in large-scale data, enabling simultaneous and efficient attention across entire input sequences. In medical imaging, this capacity can enhance recognition of subtle structures, facilitate multimodal integration, and potentially offer more general representations than CNNs alone. Ongoing research focuses on reducing computational overhead, improving data efficiency, and generating interpretable outputs, all of which are critical for realizing robust Transformer-based models in clinical practice.

\subsection{Positional Embeddings}
\label{subsec:positional_embeddings}

While the self-attention mechanism allows Transformers to capture pairwise interactions across all elements in a sequence, it does not, by itself, encode any information about the positions or ordering of those elements. Traditional recurrent and convolutional networks implicitly preserve some notion of ordering through their architecture; Transformers, on the other hand, rely on a dedicated component known as \emph{positional embeddings} \cite{vaswani2017attention} to incorporate information about sequence structure.

Positional embeddings are typically added to the input token embeddings before they pass through the Transformer’s self-attention layers. In their earliest formulation, these embeddings took the form of fixed sinusoidal functions of different frequencies, enabling the model to extrapolate to sequence lengths not seen during training. More recent studies have explored learned positional embeddings that optimize positional encodings as part of the training process, potentially improving performance but at the cost of losing some of the extrapolative properties of fixed embeddings. 

For computer vision tasks, especially where patch-based Transformers (e.g., Vision Transformer) are employed, each image patch is treated as a token in the sequence, and a positional embedding is assigned to each patch index. In the context of medical imaging, particularly with volumetric scans or higher-dimensional data, careful treatment of spatial positions becomes essential for preserving structural continuity and ensuring that the model remains sensitive to subtle anatomical variations. Approaches such as 2D or 3D positional encodings can be used to integrate row, column, and slice indices into a unified embedding, thereby capturing the multi-dimensional layout of data \cite{dosovitskiy2020image, li2021medicalvit}.

When applied to modalities like MRI or CT, positional embeddings help the model differentiate between slices or patches that appear visually similar but occupy distinct anatomical locations. In tasks such as multi-slice lesion detection or organ segmentation, the ability to associate a specific position with learned features becomes critical, since small shifts in location can imply a significant difference in pathology. As with other Transformer components, choices regarding fixed versus learned embeddings, as well as the dimensionality of these encodings, can influence both computational overhead and model accuracy. 

In summary, positional embeddings serve as a core mechanism that complements self-attention by injecting spatial awareness into Transformers. In medical image analysis, their role becomes increasingly pertinent, as the accurate interpretation of volumetric or high-resolution data hinges upon the model’s capacity to discern fine-grained spatial relationships that are key to clinical diagnosis.

\subsection{Summary}

The evolution of deep learning represents a systematic progression from classical machine learning paradigms to sophisticated architectures that increasingly emulate and extend the functional principles of biological neural systems. Initially inspired by the structure of biological neurons, early neural networks established the fundamental approach for learning hierarchical representations from raw data. The subsequent development of \acro{RNN}[Recurrent Neural Networks] introduced mechanisms for processing sequential information through the incorporation of memory and feedback, thereby enabling the modeling of temporal dependencies critical for applications such as language processing and time-series analysis. Concurrently, convolutional neural networks (CNNs) revolutionized the field by leveraging spatial hierarchies inherent in visual data, a capability that has proven indispensable for extracting intricate anatomical features in medical image analysis.

In more recent years, Transformer architectures have emerged as a compelling alternative, utilizing self-attention mechanisms to capture long-range dependencies and global contextual relationships. These architectural innovations have been complemented by advanced training methodologies—such as self-supervised learning, domain adaptation, dynamic batch adaptation, cascading sum augmentation, and backforward propagation—which have been extensively explored in our previous work. Such methods have sought to address challenges including data scarcity, computational constraints, and issues of interpretability by enriching the training process with synthetically generated data and optimized gradient propagation techniques.

Within the domain of medical imaging, the integration of these diverse models and training strategies has facilitated the automated extraction of complex features, thereby enhancing diagnostic accuracy and uncovering novel imaging biomarkers. At the same time, the application of these techniques has highlighted persistent challenges, including the need for computational efficiency, robust handling of heterogeneous data modalities, and the preservation of model interpretability and data privacy. A thorough understanding of the evolution and interplay of these deep learning architectures thus provides a critical foundation for the exploration of advanced topics—such as self-supervised and multitask domain adaptation—that are addressed in subsequent chapters.

Overall, this review not only delineates the progression from basic neural network constructs to state-of-the-art systems but also situates our contributions—developed through various studies on augmentation, gradient optimization, and multimodal fusion—within the broader context of medical image analysis. This comprehensive perspective is essential for advancing both theoretical understanding and practical applications in the field.

\section{Medical Image Analysis}

Medical imaging is fundamental to modern healthcare, allowing clinicians to visualize internal structures and processes that are otherwise inaccessible through external examination. With techniques such as \acro{MRI}, \acro{CT}, \acro{PET}, and X-ray, practitioners can diagnose and monitor a range of conditions, often detecting early signs of disease that might not be apparent by other means. These modalities encompass diverse resolutions, contrast mechanisms, and forms of noise, which can be challenging to model with traditional image-processing strategies. Over the past decade, deep learning algorithms have substantially advanced the accuracy and efficiency of tasks such as lesion detection, tissue segmentation, and the classification of pathological conditions.

Central to this progress is the ability of deep networks, notably convolutional models and increasingly Transformers, to learn high-level representations from raw data without relying on hand-engineered features. Nevertheless, clinical imaging presents several constraints that often limit the scale of fully labeled datasets. Each labeled instance in a medical context requires expert review, placing significant demands on the time and expertise of radiologists or other professionals. Privacy constraints further complicate the sharing and pooling of data, which can restrict the collective size of publicly accessible datasets. Consequently, despite the demonstrated promise of large-scale supervised learning in computer vision, medical imaging applications regularly confront the problem of insufficient labeled data.

Moreover, medical images exhibit a high degree of structural complexity. Even a single two-dimensional image can contain numerous patterns of diagnostic relevance, and this complexity increases exponentially with volumetric and multi-modal scans. For instance, an MRI volume may include many slices capturing subtle anomalies that require context from adjacent planes. Equally, a PET scan introduces functional or metabolic information, while a CT scan highlights differences in tissue density. Combining these modalities into unified pipelines can yield richer insights but also imposes new demands on model capacity and data storage.

Self-Supervised Learning has gained attention in this domain precisely because of its potential to make use of large amounts of unlabeled imaging data. Unlike strict supervised approaches, \acro{SSL} methods generate their own labels based on properties inherent to the data, such as rearranging slices (in a three-dimensional context), inpainting missing regions, or distinguishing between different transformations. This approach reduces reliance on expert annotation and opens possibilities for harnessing extensive repositories of unlabeled medical scans. The ability to pretrain a model on unlabeled volumes and then adapt it to downstream tasks—ranging from organ segmentation to disease classification—can increase performance, especially for niche pathologies where labeled data are particularly sparse.

A central concern in medical image analysis is reliability. Clinical applications leave little room for misinterpretation, demanding that models not only produce accurate predictions but also highlight the rationale behind each inference. While many \acro{SSL}-based methods focus on achieving stronger performance metrics, interpretability remains an open question. Researchers have proposed explanations based on attention maps or feature saliency, but the standard metrics for confidence in deep networks have yet to achieve universal clinical acceptance. Addressing these interpretability gaps is a key priority for deploying \acro{SSL} in healthcare, where each diagnostic decision has significant consequences for patient care.

Despite these constraints, the promise of self-supervision in medical imaging is considerable. It can alleviate the expert annotation bottleneck by providing pretrained representations that adapt well to a range of tasks involving organ-specific or disease-specific features. As data-sharing protocols improve, multi-institution collaborations can assemble larger unlabeled repositories, further fueling the power of \acro{SSL} techniques. Overall, medical image analysis stands as a vital component of this thesis, illustrating how deep learning methods—including self-supervision—can be carefully adapted to meet the rigorous demands of clinical practice while capitalizing on advances in representation learning.

\section{Self-Supervised Learning}

Self-supervised learning constitutes an increasingly influential paradigm within deep learning, characterized by its capacity to exploit the inherent structure of unlabeled data in order to generate internal supervisory signals. Rather than relying on manually curated labels, self-supervised methodologies design auxiliary tasks—often referred to as pretext tasks—whereby a network is challenged to predict, reconstruct, or otherwise infer latent aspects of its own input. This strategy not only alleviates the need for large-scale hand-labeling but also fosters the learning of representations that capture fundamental features such as shape, texture, and spatial context.

In the initial stages of this field, researchers explored relatively straightforward pretext tasks, including the prediction of geometric transformations, the reconstruction of occluded image regions, and the reassembly of shuffled image patches. These early techniques laid the groundwork for more complex approaches. With the advent of generative methods, models were trained to restore corrupted images, thereby encouraging the network to capture nuanced structural details and contextual cues. Concurrently, the emergence of contrastive learning frameworks marked a significant evolution; these methods learn robust embeddings by drawing augmented views of the same image closer together in the latent space while simultaneously distinguishing them from those of disparate samples. Such contrastive objectives, however, typically necessitate large batch sizes and careful negative sample selection, which can pose practical challenges.

A further refinement in the domain has been the introduction of knowledge distillation techniques, wherein a dual-network configuration—commonly delineated as a student–teacher model—enables one network to emulate the representations of another. This approach obviates the explicit need for negative samples and allows for a more direct transmission of representational information. Similarly, redundancy reduction methods have been proposed to encourage the independence of distinct feature components extracted from different augmented views, thereby preventing the collapse of the learned representation into trivial solutions.

The applicability of self-supervised learning to medical image analysis is particularly compelling. Medical imaging data, whether in the form of two-dimensional projections or volumetric scans, are often characterized by subtle pathological features and a pronounced scarcity of annotated examples. This scenario has motivated our research contributions, which include the development of techniques such as BrainFuse for data fusion augmentation, Efficient Dynamic Batch Adaptation to optimize training in low-data regimes, and Backforward Propagation for mitigating internal covariate shift. Each of these innovations builds upon the core principles of self-supervision by using unlabeled scans to generate richer, more robust representations. For example, BrainFuse leverages self-supervised frame interpolation to seamlessly fuse disparate brain regions, thereby augmenting the training dataset with synthetically generated yet anatomically coherent images. Similarly, advanced augmentation strategies such as cascading sum augmentation further extend the capacity of self-supervised models to extract meaningful patterns from complex data distributions.

By synthesizing these diverse approaches, self-supervised learning enables the extraction of core knowledge from raw medical images, facilitating downstream tasks such as lesion detection, segmentation, and disease classification with minimal manual annotation. The confluence of predictive, generative, contrastive, distillation, and redundancy reduction methods offers a versatile toolkit that addresses both the intrinsic complexity of clinical data and the operational constraints imposed by limited labeled samples. Consequently, this paradigm not only improves model generalization across varied imaging protocols and hardware configurations but also reduces the annotation burden, thus holding significant promise for future advancements in diagnostic accuracy and treatment planning.

Self-supervised learning provides a rigorous and data-efficient framework for medical image analysis. By embedding pretext tasks that compel networks to discern intrinsic patterns and relationships within the data, these techniques establish a foundation for the development of robust diagnostic tools. The integration of methods derived from our earlier contributions demonstrates that, even in the absence of extensive labeled datasets, deep neural networks can be trained to achieve high accuracy and resilience, thereby advancing the state-of-the-art in medical diagnostics.

\subsection{Predictive Methods}

Predictive methods represent one of the seminal approaches within the broader framework of self-supervised learning, wherein the intrinsic structure of unlabeled data is harnessed to generate supervisory signals without the need for extensive manual annotation. In these methods, synthetic labels are derived directly from the data through the formulation of pretext tasks that compel the model to predict or reconstruct portions of the input. Such tasks may include, for example, the restoration of occluded image regions, the reassembly of scrambled patches, or the inference of geometric transformations. By solving these tasks, the network is encouraged to internalize salient features—such as object contours, textural variations, and contextual relationships—that are essential for capturing the underlying semantics of the input, all while bypassing the constraints imposed by scarce expert annotations.

During the early stages of self-supervised learning research, the primary focus was on relatively elementary reconstruction objectives. These approaches typically required a network to infer missing or distorted segments of an image, thereby forcing the model to capture low-level visual patterns. However, as the field has advanced, the design of predictive tasks has grown markedly more sophisticated. Contemporary strategies now extend beyond simple reconstruction, incorporating elements of generative modeling in which a network is trained to restore an input that has been deliberately corrupted. Such generative approaches not only refine the network’s ability to capture fine-grained structural details but also lay the groundwork for more advanced techniques.

In parallel, recent research has integrated predictive methods with contrastive objectives and knowledge distillation frameworks. For instance, methods that combine predictive reconstruction with contrastive learning drive the network to produce consistent representations across different augmented views of the same data, while simultaneously learning to reconstruct missing content. Our previous work on BrainFuse exemplifies this evolution by leveraging a frame interpolation model to generate spatially coherent transitions between distinct brain regions. This approach, which can be viewed as a predictive task adapted to the particular challenges of volumetric medical imaging, effectively fuses anatomical structures from different subjects and demonstrates how predictive self-supervision can be harnessed to generate novel, label-preserving samples.

Furthermore, investigations into techniques such as Backforward Propagation and Efficient Dynamic Batch Adaptation have underscored the role of predictive mechanisms in mitigating issues such as internal covariate shift and in stabilizing the training process. These studies have shown that by formulating appropriate predictive tasks, one can induce a regularizing effect that not only improves convergence but also enhances the model’s capacity to generalize in data-scarce medical imaging scenarios. Similarly, our work on Cascading Sum Augmentation illustrates how progressively blending multiple samples via linear interpolation can be interpreted as a predictive operation, where the model is implicitly required to infer a smooth transition between disparate anatomical regions.

In the context of medical image analysis, the benefits of predictive self-supervised learning are particularly pronounced. Clinical imaging studies often contend with limited labeled data and subtle pathological variations that are difficult to annotate reliably. By designing domain-specific predictive tasks that are sensitive to the nuances of modalities such as MRI, CT, and PET, it is possible to pretrain networks to learn robust, transferable features. These pretrained models can then be fine-tuned for specialized diagnostic tasks—such as lesion detection, tissue segmentation, or disease classification—with minimal additional annotation. Moreover, predictive methods offer a mechanism to improve robustness across different imaging protocols and hardware settings, addressing one of the key challenges in medical imaging research.

In summary, predictive methods in self-supervised learning offer a powerful and versatile framework for extracting meaningful representations from unlabeled data. By carefully crafting pretext tasks that are both general and domain-specific, these methods enable the development of deep learning models that are well-suited for the complex and often heterogeneous nature of medical images. This paradigm not only reduces the dependency on large-scale labeled datasets but also facilitates the creation of models with enhanced diagnostic accuracy and resilience, thereby advancing the state of the art in medical image analysis.

\begin{figure}
\centering
\includegraphics[width=0.9\textwidth]{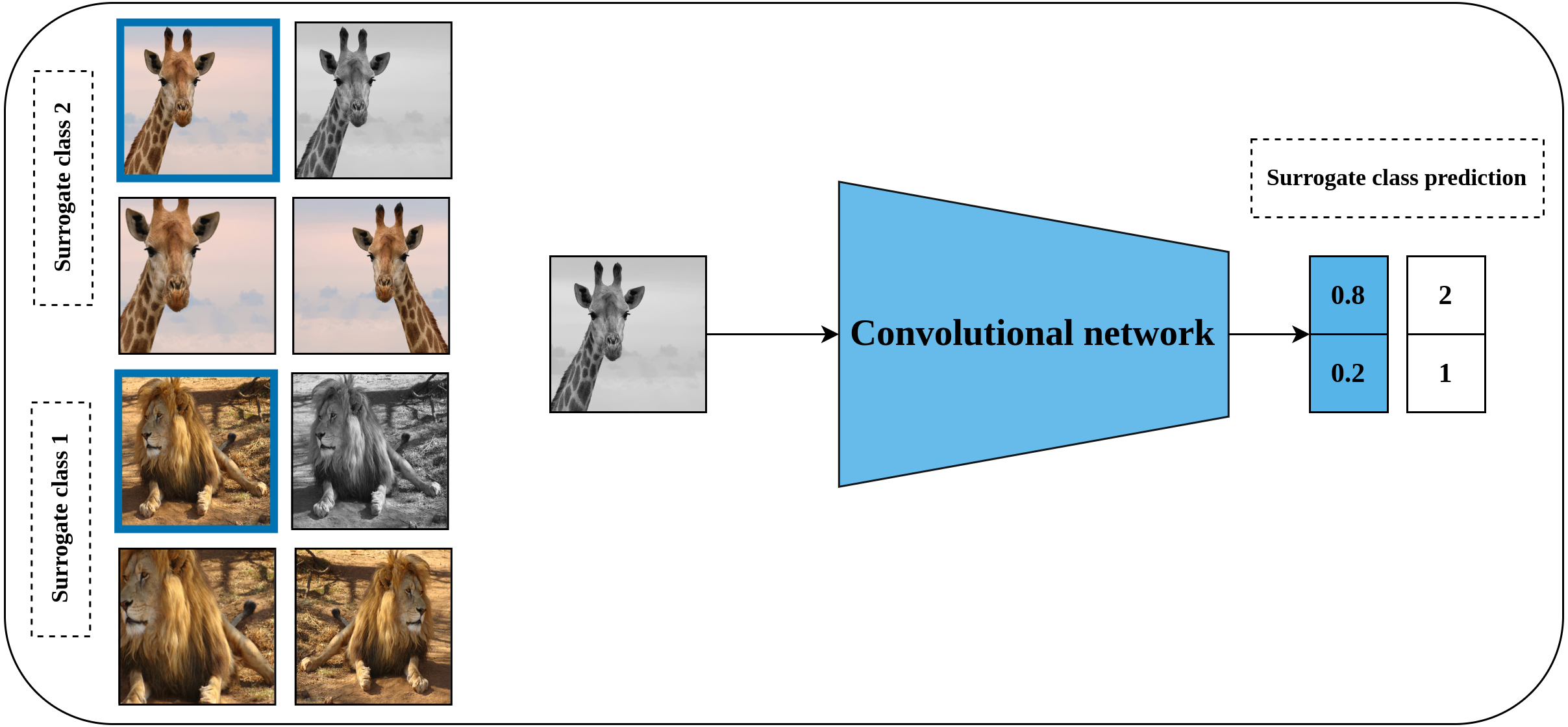}
\caption{Illustration of pseudo-label assignment in the CNN Exemplar method\protect\footnotemark}
\label{fig:cnn_exemplar}
\end{figure}

\paragraph{CNN-Exemplar.}
Among the early paradigms in predictive self-supervised learning, the CNN-Exemplar method introduced by Dosovitskiy et al. \cite{dosovitskiy2014discriminative} stands as a foundational approach that redefines the role of data in learning robust visual representations. In this method, a carefully selected subset of samples from the available dataset is subjected to a wide range of deliberately designed transformations. These transformations, which include both geometric distortions and photometric variations, are applied uniformly across the selected examples, thereby generating a set of altered images that encapsulate the intrinsic variability of the data. Each transformed image is then endowed with a pseudo-label that corresponds uniquely to the index of its original sample, as illustrated in Figure~\ref{fig:cnn_exemplar}. This process effectively constructs an artificial classification task, wherein the network is trained to discriminate among the different pseudo-classes defined solely by the transformation-induced variations. The outcome is a model that learns transformation-invariant features and develops a discriminative embedding space without any reliance on manually annotated labels.

A further prevalent category of predictive methods within this framework involves the explicit prediction of various geometric transformations. Rather than solely assigning fixed pseudo-labels, these methods require the model to infer the specific transformation parameters that were applied to an input sample. This strategy not only reinforces the network’s sensitivity to subtle structural and spatial cues but also forces it to capture higher-level semantic properties of the data. Such an approach has proven especially valuable in domains characterized by limited labeled data, such as medical imaging, where the cost and difficulty of obtaining expert annotations are significant. In our own research, we have integrated similar predictive tasks within broader self-supervised schemes, as exemplified by our work on dynamic batch adaptation and advanced data augmentation techniques. By embedding transformation prediction into the training objective, the network is guided to discern inherent invariances and extract features that are robust to variations in acquisition conditions and anatomical deformations. This capacity to learn from pseudo-labels generated through geometric perturbations is instrumental in creating models that generalize well across heterogeneous medical datasets, thereby bridging the gap between unsupervised pretraining and downstream clinical applications.
\footnotetext{DOI: 10.7717/peerjcs.1045/fig-4}

\begin{figure}
\centering
\includegraphics[width=0.9\textwidth]{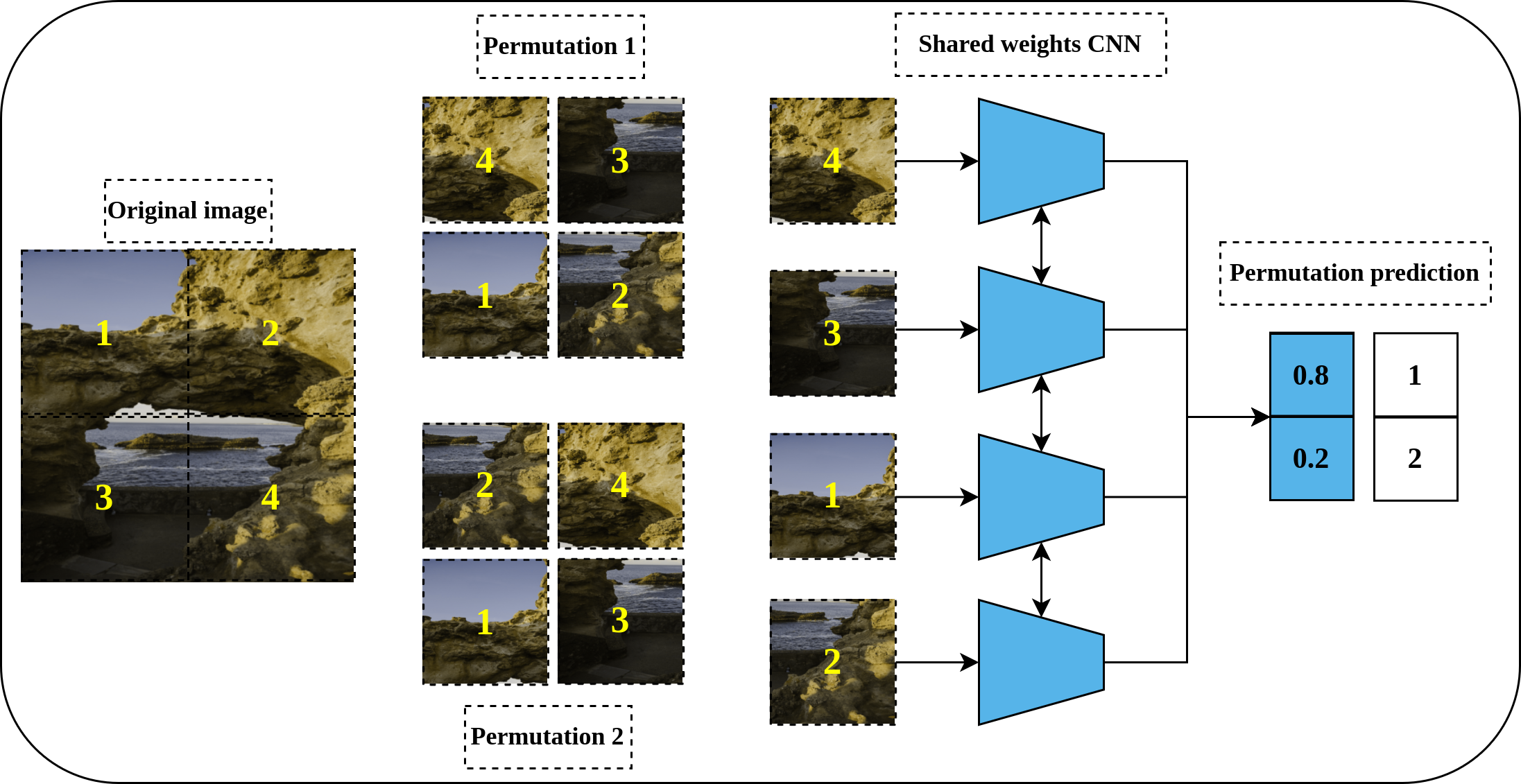}
\caption{Visualization of the Jigsaw pretext task\protect\footnotemark}
\label{fig:jigsaw}
\end{figure}

\paragraph{Jigsaw.}
The Jigsaw pretext task constitutes a self-supervised learning paradigm designed to compel a network to recover the original spatial arrangement of an image that has been deliberately disassembled. In a canonical implementation, the input image is subdivided into a grid of segments—typically arranged in a 3×3 configuration—and these segments are subsequently subjected to a random permutation. This deliberate disruption of the inherent spatial order transforms the reconstruction objective into a classification challenge, where the model must identify the precise permutation applied; each class in this setting corresponds to one of the selected permutation indices. As illustrated in Figure~\ref{fig:jigsaw}, the task necessitates that the network concurrently assimilate both fine-grained local features and the overarching global structure: it must learn to reconstitute the image by accurately delineating the boundaries between patches, discerning subtle variations in color and texture, and comprehending the contextual relationships that originally governed the spatial layout.

\footnotetext{DOI: 10.7717/peerjcs.1045/fig-5}

Although the operational premise appears straightforward, the solution of the Jigsaw task imposes a significant burden on the learning algorithm. A principal challenge arises from the vast combinatorial space of possible patch arrangements; this complexity compels the practitioner to restrict the training process to a judiciously chosen subset of permutations that are deemed most informative. Such a selection not only introduces an additional layer of methodological intricacy but also raises pertinent questions regarding the optimal criteria for choosing representative and discriminative permutation patterns. The theoretical basis of the Jigsaw approach is deeply rooted in antecedent methods such as relative position prediction \cite{doersch2015unsupervised}, wherein the model is tasked with inferring the spatial relation between pairs of patches extracted from a single image. By mandating the correct reassembly of shuffled patches, the network is implicitly driven to acquire robust, transferrable feature representations that extend well beyond the immediate reordering objective, thereby enhancing performance in subsequent, downstream applications.

In the specialized context of medical imaging, where the spatial organization of anatomical structures is paramount, the Jigsaw framework has been innovatively adapted to three-dimensional data. Rather than confining the operation to two-dimensional patches, volumetric medical images are partitioned into slices or subvolumes which are then permuted. This three-dimensional extension further intensifies the task, as the model must capture not only in-plane features but also the inter-slice spatial relationships critical for accurate anatomical interpretation, all without the assistance of externally provided annotations. Such adaptations underscore the versatility and potential of the Jigsaw task for self-supervised representation learning in medical imaging, where the preservation of detailed anatomical integrity and the capacity to detect subtle pathological alterations are essential.

\paragraph{Rotation Prediction.}
Within the predictive self-supervised framework, the rotation prediction task entails subjecting input images to a predetermined set of rotations—typically by angles of 0°, 90°, 180°, or 270°—and training the network to accurately infer which rotation was applied. This task compels the model to learn both global structural information and fine-grained local orientation details. To correctly determine the applied rotation, the network must capture subtle visual cues, such as variations in illumination, shadow patterns, and the spatial arrangement of key features that collectively define the canonical orientation of an image.

In medical imaging, where modalities such as radiographs, computed tomography, and magnetic resonance imaging exhibit complex anatomical structures that can vary significantly due to differences in patient positioning and imaging protocols, rotation prediction is particularly valuable. By requiring the model to differentiate between rotated versions of the same image, this pretext task promotes the extraction of invariant features that are robust to geometric transformations. In doing so, the model is encouraged to focus on intrinsic characteristics of the anatomical content rather than relying on superficial cues, which is essential for accurately detecting region-specific anomalies and subtle pathological variations.

Thus, rotation prediction serves as an effective self-supervised objective that reinforces the learning of transferable and discriminative feature representations. This enhanced sensitivity to both global and local cues provides a more reliable foundation for subsequent medical image analysis tasks, where precise orientation and subtle variations are critical for clinical interpretation.

\paragraph{General Strengths.}
Predictive methods represent a class of self-supervised learning techniques that exploit the intrinsic structure of data by formulating pretext tasks in which the model is trained to reconstruct or predict missing portions of its input. Owing to their conceptual clarity and relative simplicity, these methods offer several notable advantages. In contrast to more elaborate approaches—such as contrastive frameworks or teacher–student distillation schemes that require large batch sizes or specialized hardware—predictive methods can be implemented with modest computational demands. This computational efficiency facilitates rapid prototyping and enables seamless integration into conventional supervised pipelines, where self-supervised pretraining serves as an effective initialization stage prior to fine-tuning on labeled data. Moreover, the direct interpretability of the predictive objective, in which performance can be assessed by the fidelity of the reconstructed input, provides researchers with clear insights into the model’s learning dynamics. Such advantages have been demonstrated in our prior work on dynamic batch adaptation and backforward propagation, where the predictive objectives contributed to learning robust and transferable representations without excessive resource overhead.

\paragraph{General Weaknesses.}
Despite these appealing features, predictive methods are intrinsically limited by the extent to which their pretext tasks capture the semantically rich and clinically relevant nuances of the data. When the chosen objective is too narrowly defined, the model may exploit superficial statistical patterns—often referred to as "shortcut" solutions—rather than developing a comprehensive understanding of the underlying anatomical or pathological features. This risk is exacerbated in scenarios where the pretext task is misaligned with the domain-specific complexities, leading to representations that lack robustness and generalizability. Empirical evidence from our research indicates that, while predictive methods can yield useful features, they often achieve lower performance compared to more contextually expansive approaches such as contrastive learning or distillation techniques. These alternative methods are better equipped to leverage global contextual cues and incorporate additional regularization mechanisms, thereby producing more powerful and generalizable feature representations.

\paragraph{Medical Imaging Considerations.}
In medical imaging, predictive self-supervised methods encounter unique challenges that are less prevalent in natural image analysis. The standardized acquisition protocols and inherent uniformity of modalities such as MRI and CT scans can cause pretext tasks—such as jigsaw puzzles or contiguous slice reconstruction—to emphasize global positional cues and overall intensity distributions at the expense of fine-grained anatomical details. Nevertheless, when the predictive objective is carefully tailored to the clinical context—such as by focusing on the restoration or reordering of contiguous slices in a three-dimensional volume—the method can effectively encode the subtle structural variations that are critical for accurate diagnosis. Such approaches enable the model to capture the intricate interplay between anatomical structure and pathological markers while mitigating the tendency to overfit to trivial features. In this way, although predictive methods alone may not fully encompass the complexity of medical data, they form a valuable foundational component within a comprehensive self-supervised learning framework tailored specifically for medical image analysis.

\subsection{Generative Methods}

Generative methods form a fundamental category of self-supervised learning strategies in which models are trained to reconstruct or synthesize input data from deliberately perturbed versions. In these approaches, the network is presented with inputs that have been degraded—through the introduction of noise, occlusion, or other corruptions—and is then tasked with recovering the original, unaltered signal. This reconstruction objective compels the model to internalize both local and global structural properties of the data, thereby facilitating the extraction of semantically meaningful features without reliance on extensive manual labeling. Such methods have been extensively explored in computer vision and have found particular relevance in medical imaging applications, where the limited availability of annotated data and the inherent heterogeneity of images necessitate robust, self-supervised feature learning.

\begin{figure}
    \centering
    \includegraphics[width=1.0\textwidth]{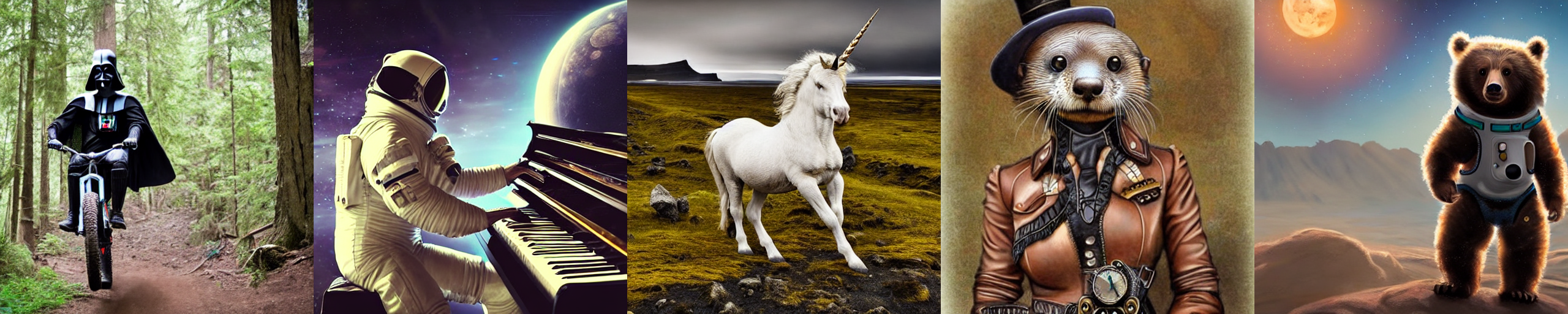}
    \caption{Example of Stable Diffusion outputs}
    \label{fig:stable_diffusion}
\end{figure}

Recent work in generative modeling has underscored the efficacy of diffusion-based approaches, such as those exemplified by Stable Diffusion, which train networks on large unlabeled datasets to develop an intrinsic understanding of image statistics and spatial coherence. As illustrated in Figure~\ref{fig:stable_diffusion}, these models iteratively refine partially corrupted inputs to produce visually consistent outputs that preserve essential anatomical details—a property that is especially valuable in medical domains where subtle variations may indicate critical pathology.

In practice, autoencoder architectures have become one of the most widely adopted frameworks for generative representation learning. A conventional autoencoder consists of an encoder that compresses the input image into a compact latent space, and a decoder that reconstructs the original image from this latent representation. By optimizing a reconstruction loss, typically expressed as the mean squared error between the original and reconstructed images, the model is encouraged to capture the core features that are indispensable for accurate recovery. This paradigm is particularly useful for tasks such as denoising, inpainting, or colorization, where the objective is to recover information that has been purposefully omitted or altered.

\begin{figure}
    \centering
    \includegraphics[width=0.9\textwidth]{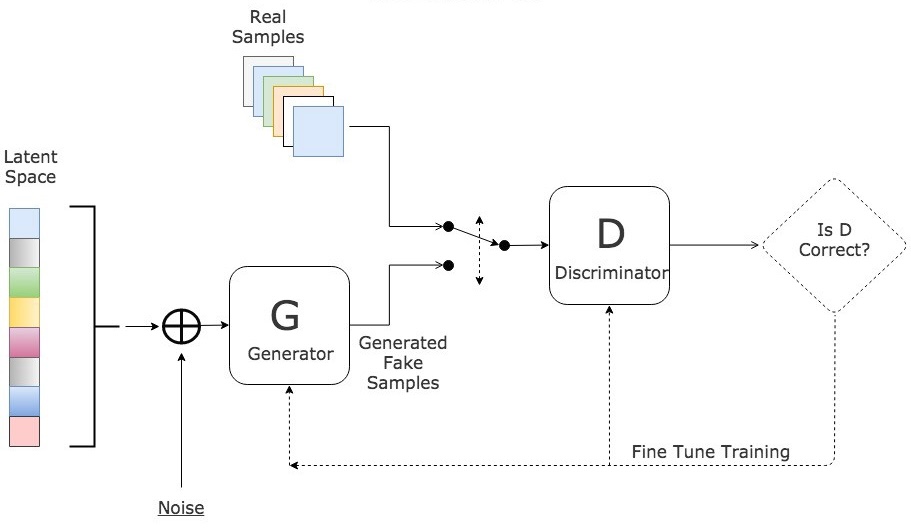}
    \caption{GAN Architecture\protect\footnotemark}
    \label{fig:gan}
\end{figure}

Variational Autoencoders (VAEs) extend the autoencoder framework by imposing a probabilistic structure on the latent space. By modeling latent representations as probability distributions and incorporating a regularization term (for example, the Kullback–Leibler divergence), VAEs enable the generation of novel samples through stochastic sampling. Although VAEs often require careful balancing between reconstruction fidelity and latent-space regularization, they provide a mechanism for synthesizing diverse images—a capability that can be leveraged to augment training datasets in medical imaging. Techniques such as BrainFuse, which employs frame interpolation to generate smooth transitions between disparate brain regions, and Cascading Sum Augmentation, which synthesizes new samples by linearly combining multiple inputs, exemplify how generative methods can be adapted to capture complex anatomical variations in modalities such as MRI and CT scans.

Another prominent approach within the generative paradigm involves adversarial training, as instantiated by Generative Adversarial Networks (GANs) \cite{goodfellow2020generative}. In GANs, a generator network is trained to produce realistic images from random noise or partially corrupted inputs, while a discriminator network is simultaneously trained to distinguish between real and generated images. The adversarial interplay between these networks can lead to highly detailed and plausible outputs; however, this approach typically requires meticulous hyperparameter tuning and is vulnerable to issues such as mode collapse, which can be particularly challenging in the context of medical image synthesis. The generic GAN architecture can be seen in Figure \ref{fig:gan}

\footnotetext{http://www.kdnuggets.com/2017/01/generative-adversarial-networks-hot-topic-machine-learning.html}

\begin{figure}
    \centering
    \includegraphics[width=0.8\textwidth]{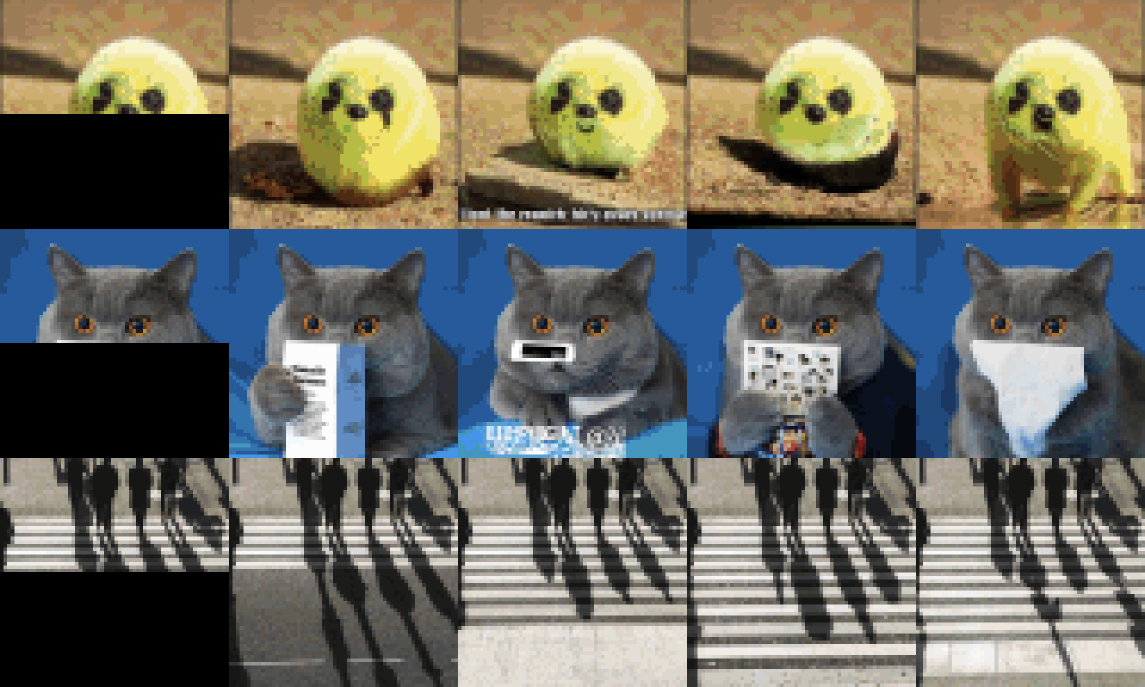}
    \caption{Example of Autoregressive Generation, Image GPT}
    \label{fig:image_gpt}
\end{figure}

\begin{figure}
    \centering
    \includegraphics[width=0.95\textwidth]{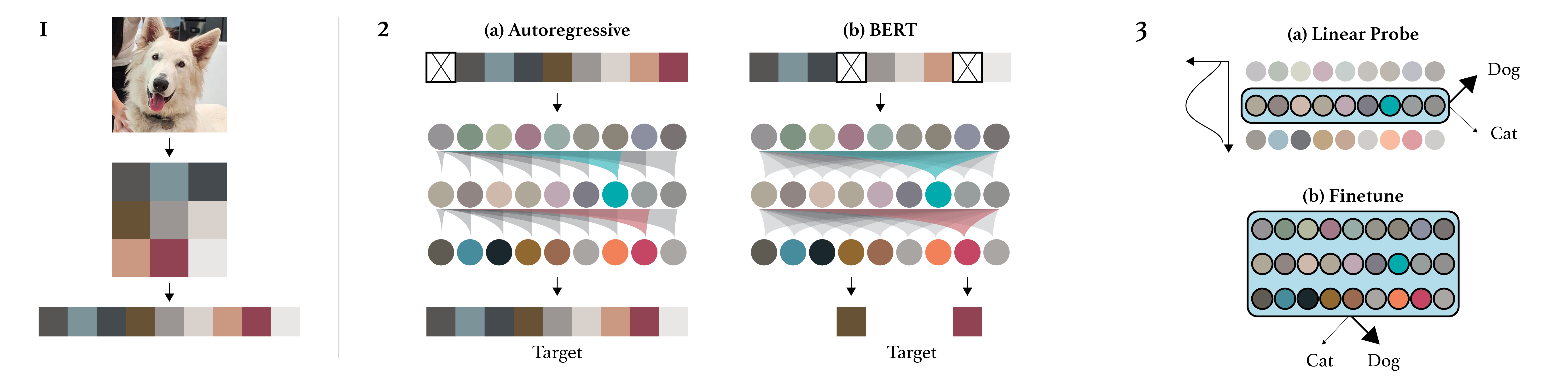}
    \caption{Image GPT Architecture}
    \label{fig:image_gpt_arch}
\end{figure}

Inspired by NLP, the method Autoregressive Next Pixel Generation \cite{chen2020generative} has been adapted from language models. This works by sequentially asking the model to predict the next pixel or patch of pixels in a sequence, see Figure \ref{fig:image_gpt}. A 2D image would first be flattened and then formed in a sequence of patches or individual pixels. The current sequence of pixels is passed through the model and the target of the task is to predict the next pixel or patch in the sequence. Since the next pixel in the original image is known, it will be used as the pseudo-label. Both pixel masking and autoregressive generation can be seen in Figure \ref{fig:image_gpt_arch}.

Generative methods offer significant advantages in self-supervised learning pipelines by compelling models to focus on intrinsic patterns and relationships within the data. In medical imaging, where expert annotations are expensive and often limited, such approaches enable the effective utilization of unlabeled data for pretraining and data augmentation. By emphasizing tasks such as noise removal and inpainting, generative methods help the network to learn nuanced visual representations that transfer effectively to downstream tasks including classification, segmentation, and detection. Nonetheless, these methods are not without limitations: the computational cost of training complex generative models can be high, and an excessive focus on pixel-level continuity may sometimes come at the expense of capturing higher-level semantic information. It is therefore essential to design and integrate generative approaches carefully, tailoring them to the specific requirements of medical imaging applications.

\paragraph{Auto-Encoding.} 
Auto-encoding techniques have emerged as a particularly effective subset of generative methods by compelling models to learn compact representations through reconstruction tasks. In a standard autoencoder, the encoder network maps the input image to a reduced-dimensional latent space, and the decoder subsequently reconstructs the original image from this latent code. To challenge the network further and encourage robust feature extraction, various corruption strategies are employed during training. For instance, denoising autoencoders introduce random noise into the input and require the recovery of the clean image; inpainting autoencoders mask out selected regions, prompting the network to predict the missing content; and colorization autoencoders convert images to grayscale, training the model to restore the original colors. In medical imaging contexts, these techniques are particularly beneficial for addressing issues such as artifact correction and the enhancement of subtle pathological features. Variational extensions, such as VAEs, further incorporate stochasticity by modeling the latent space as a distribution, thereby enabling the generation of novel images that can supplement limited clinical datasets. This capacity for both reconstruction and synthesis makes auto-encoding approaches a valuable component of self-supervised learning pipelines in domains where labeled data are scarce and the preservation of fine structural details is critical.

\paragraph{Inpainting and Context Restoration.}
In the domain of generative methods within self-supervised learning, inpainting, and context restoration have emerged as fundamental strategies for reconstructing missing or degraded regions in medical images. These techniques require the network to not only reproduce local texture details but also to recover the global semantic structure inherent in the data, thereby ensuring that any synthesized content is visually congruent with its surrounding regions and remains faithful to the overall anatomical layout. By enforcing reconstruction over multiple scales, the network is compelled to capture fine-grained patterns while simultaneously maintaining awareness of broader structural cues, a duality that is particularly critical in medical imaging where even subtle deviations can bear significant clinical implications.

Our prior research efforts—ranging from the BrainFuse framework, which leverages frame interpolation to generate smooth transitional regions between disparate brain sections, to the Cascading Sum Augmentation technique that systematically fuses multiple samples to populate the latent feature space—illustrate the efficacy of employing inpainting-like operations to mitigate data irregularities. Such methods have proven instrumental in bridging gaps caused by artifacts or missing data in modalities like MRI and CT, thus producing reconstructions that preserve both the nuanced textural information and the overarching anatomical context. Moreover, by integrating context restoration into the generative process, these methods facilitate the development of robust, multi-scale representations that are resilient to domain shifts and other sources of variability commonly encountered in clinical datasets.

In effect, the task of inpainting and context restoration within our framework is not merely a pixel-level reconstruction exercise; it is an essential mechanism by which the network internalizes complex interdependencies between local image features and global structural semantics. This comprehensive learning process ultimately leads to more consistent and reliable representations, thereby enhancing downstream analysis tasks such as segmentation, classification, and predictive modeling in medical imaging. In summary, by compelling the model to restore both missing details and contextual integrity, inpainting and context restoration methods contribute significantly to our overarching objective of achieving high-fidelity, anatomically plausible reconstructions that underpin improved clinical decision support.

\paragraph{General Strengths.}
A central strength of generative approaches is their ability to capture a rich and detailed representation of the underlying data distribution. By learning to synthesize new instances, these methods effectively introduce controlled variability into the training process, thereby mitigating the challenges associated with data scarcity. For instance, techniques such as Cascading Sum Augmentation demonstrate how linear combinations of multiple samples can be leveraged to populate a high-dimensional feature space, enriching the diversity of available training examples without sacrificing domain-specific structural fidelity. In addition, the reconstruction objectives intrinsic to generative models render the network’s behavior more transparent; the reconstructed outputs provide a direct means of visualizing the features that the model deems critical, thereby facilitating both qualitative assessments and iterative refinements of the training procedure. This interpretability is particularly beneficial for understanding how variations in the input, such as subtle anatomical differences in medical images, are encoded and can thus inform subsequent adjustments in model design and hyperparameter tuning.

\paragraph{General Weaknesses.}
Notwithstanding their considerable advantages, generative models are often accompanied by substantial computational demands and training complexities. The high parameter counts and the iterative nature of reconstruction tasks necessitate significant processing resources, especially when dealing with high-resolution or volumetric data. Moreover, the training dynamics of such models are susceptible to instabilities—manifested as mode collapse or vanishing gradients—which may compromise the model’s ability to capture the full spectrum of semantic variability inherent in the data. Our research on Backforward Propagation and Efficient Dynamic Batch Adaptation has highlighted that these instabilities often require meticulous tuning of numerous hyperparameters and may involve trade-offs between model expressiveness and computational efficiency. Furthermore, the optimization process in generative frameworks may inadvertently prioritize low-level image details over deeper semantic information, thus necessitating a careful balance between reconstruction fidelity and abstract representation learning. These factors collectively underscore the challenges in achieving consistently stable and high-quality outcomes when employing generative methods in practice.

\paragraph{Medical Imaging Considerations.}
Within the realm of medical imaging, the application of generative methods is particularly promising given their potential to ameliorate data scarcity through the synthesis and augmentation of complex image data. Techniques that facilitate cross-modality synthesis—such as generating computed tomography images from magnetic resonance imaging inputs—illustrate the capacity of these methods to bridge gaps between different imaging modalities while preserving essential diagnostic features. However, the high resolution and three-dimensional structure of medical images impose additional computational and methodological challenges; any artifacts introduced during the generation process could have critical clinical implications. Consequently, it is imperative that generative models tailored for medical applications are designed with rigorous domain-specific constraints to ensure that the synthesized outputs faithfully reproduce both the anatomical and pathological nuances of the original data. Our integrated approach, which combines self-supervised learning strategies with domain adaptation techniques, underscores the necessity of balancing the benefits of synthetic data augmentation against the risks of introducing spurious or misleading features in a clinical context. This careful calibration is essential to maintain the interpretative clarity and diagnostic reliability required for applications in medical imaging.

\subsection{Contrastive Methods}
Contrastive methods have emerged as a prominent strategy within self-supervised learning, particularly for computer vision tasks. In contrast to generative or predictive approaches, contrastive techniques aim to learn representations by distinguishing among different samples rather than reconstructing or predicting parts of an input. They often employ a pairwise or Siamese-style architecture, where two transformed versions of an input image are fed into separate (yet identical) branches of a network. The network is then encouraged to minimize the distance between embeddings originating from the same sample (positive pairs) while maximizing the distance between embeddings originating from different samples (negative pairs). This framework compels the model to uncover relevant similarities and differences in data, resulting in robust representations that can be transferred to downstream tasks such as classification or segmentation.

\begin{figure}[ht]
    \centering
    \includegraphics[width=1.1\textwidth]{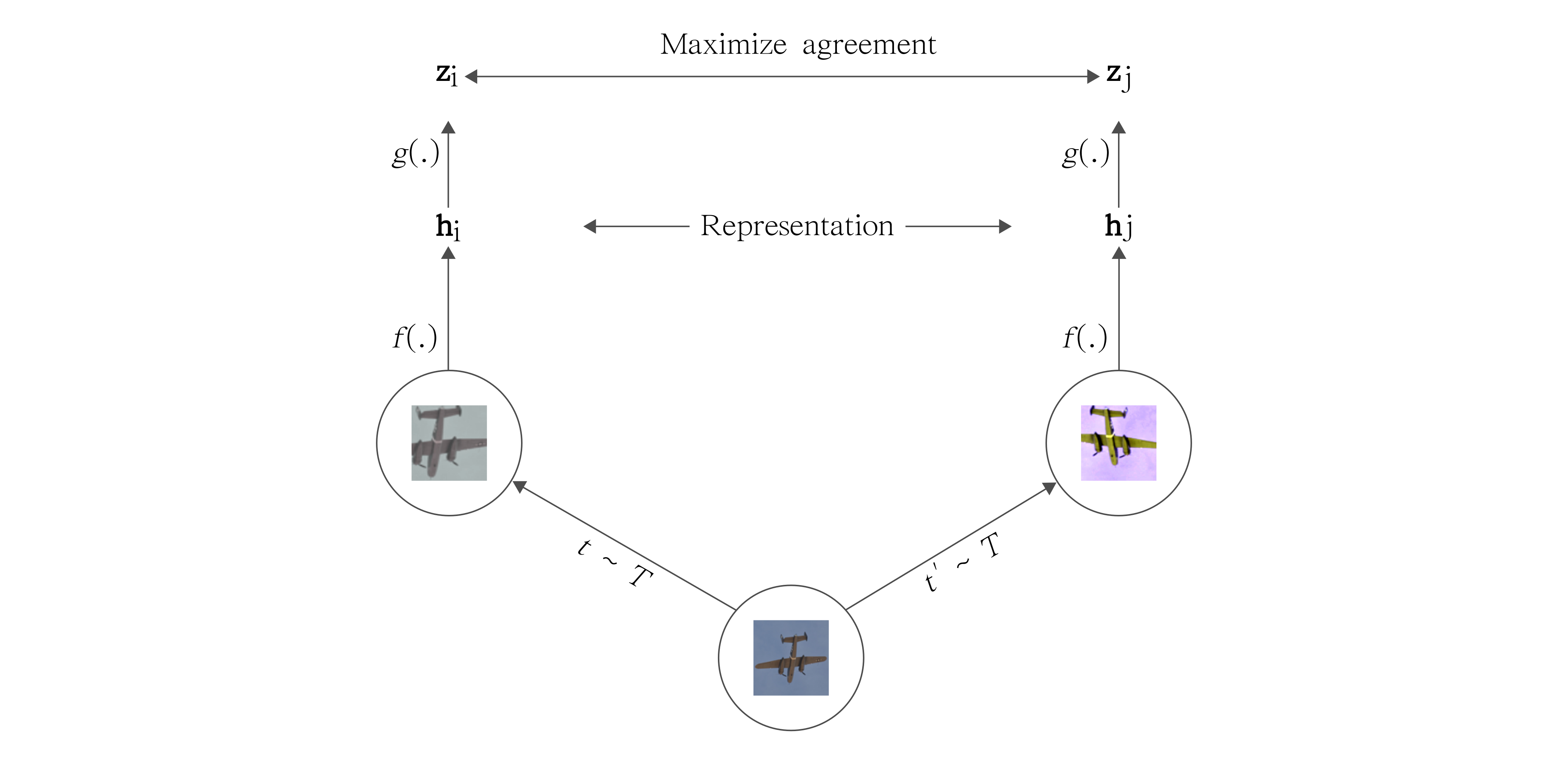}
    \caption{Simplified depiction of a contrastive learning pipeline similar to SimCLR. Two augmented views of the same sample are passed through identical branches. The objective encourages closeness in the embedding space for these matching views while separating embeddings from different samples.}
    \label{fig:simclr_illustration}
\end{figure}

A typical implementation of contrastive methods can be illustrated by the SimCLR framework, which applies diverse augmentations—like random cropping, color jitter, or Gaussian blur—to generate two distinct views of each training sample. These views pass through shared neural network branches to produce feature embeddings. The learning objective employs a contrastive loss function that brings embeddings from the same sample closer together in a shared space and pushes embeddings from different samples farther apart. Figure~\ref{fig:simclr_illustration} provides a simplified depiction of the SimCLR architecture. By training on large amounts of unlabeled data in this manner, a model learns features that capture salient structures, patterns, and contexts in images. Once trained, these learned features can significantly reduce the data requirement for subsequent supervised tasks, as they allow the model to generalize more readily from a smaller number of labeled examples.

Despite their success, contrastive methods face challenges, particularly in selecting suitable negative pairs. In many cases, random selection of negatives may inadvertently pair visually or semantically similar samples as negatives, thereby confusing the learning process. As a result, recent advances focus on more intelligent mining of negative samples, often with large batch sizes or memory banks to ensure robust negative examples. This can prove resource-intensive, especially with high-resolution images or three-dimensional scans, as is common in medical imaging. Nevertheless, the ability of contrastive learning to produce highly discriminative representations has led many researchers to adapt or refine these strategies for medical contexts, aiming to capture clinical nuances that might not emerge through generative or predictive tasks alone.

\paragraph{General Strengths.}
Generative methods provide a detailed understanding of the underlying data distribution and can effortlessly handle multiple modalities. They enable the creation of new synthetic data for augmentation purposes, which can enhance model robustness by introducing controlled variability while preserving core domain characteristics. Furthermore, these approaches allow for interpretable outputs, since the reconstructed or generated images reveal how the network processes and restores crucial features. This capacity to visualize the network’s learned content can offer insights into model behavior and potentially highlight areas of improvement in the training process.

\paragraph{General Weaknesses.}
Contrastive methods, despite their usefulness in representation learning, face several challenges that can limit their applicability. They generally demand large batch sizes to provide a sufficient diversity of negative pairs; otherwise, the algorithm risks pairing samples that are too similar and disrupting the learning process. Moreover, they tend to be highly sensitive to the selection of negative pairs, requiring careful curation of examples to avoid overwhelming the model with near-duplicates or improperly matched samples.

Additionally, these methods are often computationally expensive because they rely on numerous pairwise comparisons in the training phase. The resulting memory requirements can be considerable, especially for high-resolution or volumetric data, making it difficult to scale contrastive approaches in domains such as medical imaging where large-scale and complex data are common.

\paragraph{Challenges in Medical Imaging.}
Contrastive methods face particularly acute challenges in the medical imaging domain. A central issue is the high degree of visual similarity among these images. This similarity arises from several factors that encompass the fundamental uniformity of human anatomy and the standardized protocols used to capture scans.

Contrastive methods face particularly acute challenges in the medical imaging domain. The fundamental issue stems from the inherently high degree of visual similarity among medical images. This similarity manifests in two key ways:
\begin{enumerate}
    \item \textbf{Anatomical Homogeneity:} Human anatomy follows consistent patterns across individuals. The structural organization of organs, bones, and tissues maintains remarkable uniformity across the population. This biological consistency means that even images from different patients or diagnostic categories may appear nearly identical at a visual level.
    
    \item \textbf{Standardized Imaging Protocols:} Medical imaging procedures are highly standardized, employing uniform protocols, positioning, and value ranges. This standardization, while crucial for diagnostic reliability, results in images that share very similar visual characteristics regardless of their underlying pathological differences.
\end{enumerate}
For instance, chest X-rays from multiple individuals commonly present the same broad structural layout, regardless of whether a particular patient is healthy or exhibits distinct pathologies. This renders traditional strategies for defining negatives less effective, as images that should logically function as negatives can be nearly indistinguishable from each other. In response, contrastive approaches designed for medical imaging often integrate adaptations such as especially large batch sizes to ensure adequate diversity of training samples. They may also employ mechanisms for identifying more informative negative examples or incorporate domain-specific guidance in selecting pairs that better highlight clinically significant differences. By taking these considerations into account, researchers strive to mitigate the challenges posed by the inherent visual similarity found in medical images, thereby allowing contrastive methods to extract meaningful and discriminative representations.

\subsection{Clustering Methods}
Clustering methods form another avenue of self-supervised learning by alternating between grouping visually similar data instances and refining the model’s learned representations based on these groupings. In practice, the process typically involves an initial feature extraction step, followed by a clustering stage that partitions the data according to features learned so far. Once these clusters are formed, they serve as pseudo-labels; that is, the cluster assignments are treated as supervisory signals that guide the network toward better representations.

One widely noted example is DeepCluster. First, it extracts features from the current state of the network and applies a clustering algorithm—commonly k-means—to group the data into clusters. The cluster assignments then function as pseudo-labels for a subsequent training step, where the model is updated to classify images into these cluster-derived classes. By iterating this procedure, DeepCluster refines both the clusters themselves and the intermediate representations upon which the clustering is based.

The benefits of clustering-based methods lie in their capacity to discover natural groupings within unlabeled data without relying on large batch sizes. They can handle an unbounded number of potential classes, reflecting the inherent complexity and variability in real-world data. In many cases, these methods also reduce some of the computational overhead that contrastive methods often require when constructing and comparing large sets of negative pairs. However, clustering approaches are known to be sensitive to initialization and can suffer from instability when repeatedly reassigning clusters. They also risk converging to degenerate solutions where one large cluster dominates the distribution, failing to capture meaningful distinctions among data points. Moreover, the semantic relevance of each discovered cluster may not always align with meaningful classes or clinically relevant categories when applied to medical data. Still, for many domains—including some medical image settings—this line of self-supervised learning can prove useful, especially as a tool for generating pseudo-labels and revealing hidden structures within the data.

\paragraph{DeepCluster.}
One well-known illustration of clustering-based self-supervision is DeepCluster. The technique operates by first extracting features from images using an evolving neural network and then grouping those features into clusters, often through the k-means algorithm. Each cluster thus becomes an unlabeled category, allowing the cluster indices to be treated as provisional or pseudo-labels. The model is then retrained to predict these pseudo-labels, leading to refined feature extraction in subsequent iterations. Over time, the network improves both how it separates data into clusters and how it represents each sample within a cluster. This iterative loop—a feature extraction step followed by clustering and re-labeling—propels the network to better differentiate among a variety of unlabeled data samples. In practice, DeepCluster can discover structure in large collections of images, even when no ground truth annotations are available, and can serve as an initialization for downstream tasks where actual labels are scarce.

\paragraph{General Strengths.}
Clustering-based strategies for self-supervised learning can offer several advantages relative to other paradigms. Because they form pseudo-labels through an iterative process of clustering and feature refinement, they do not necessarily demand large batch sizes, which are often a bottleneck in contrastive approaches. Moreover, clustering-based methods can handle an unbounded number of potential classes, reflecting the complexity and variability frequently encountered in real-world datasets. By allowing the structure of the data to emerge from the clustering process itself, these methods can reveal latent groupings that might otherwise be overlooked. In many cases, clustering techniques also place fewer constraints on computational resources, as they focus on grouping similar features rather than exhaustively comparing positives and negatives within the same batch. This flexibility can make them particularly suitable for domains where annotated data is scarce but unlabeled data remains plentiful. 

\paragraph{General Weaknesses.}
Although clustering methods can reveal hidden structures in unlabeled datasets, they exhibit certain drawbacks that may hinder their overall effectiveness. They tend to be highly sensitive to initialization strategies and can be unstable over multiple rounds of training, sometimes converging to degenerate configurations in which one large cluster dominates the distribution. Additionally, the cluster assignments they produce may not necessarily align with clinically significant or semantically meaningful categories, particularly in domains where subtle variations are diagnostically relevant. These shortcomings highlight the importance of careful algorithm design and parameter selection when relying on clustering approaches for self-supervised representation learning.

\paragraph{Medical Imaging Considerations.}
Clustering methods can be particularly effective for medical imaging as they can naturally handle the continuous spectrum of pathological variations. However, the high visual similarity between different medical conditions can lead to meaningless clusters based on image acquisition parameters rather than clinically relevant features. The high dimensionality of medical images (especially 3D scans) can also make clustering computationally challenging.

\subsection{Distillation Methods}
Distillation approaches in self-supervised learning gained traction by removing the explicit need for negative pairs typically found in contrastive paradigms. Instead, these methods rely on a student–teacher framework, allowing a model to learn robust representations through knowledge transfer between two different network pathways. The central idea is that one pathway (the student) is updated directly by gradient descent on the current training batch, while the other pathway (the teacher) is updated in a manner that does not necessarily propagate gradients. This arrangement prevents the collapse problem often encountered in methods that do not rely on negative pairs, thus facilitating stable training even with smaller batch sizes or limited data.

Bootstrap Your Own Latent (BYOL) \cite{grill2020bootstrap} is an influential example of this category. It employs two neural networks, commonly called the online network and the target network, which process two augmented versions of the same input. The online network is trained to match the target network’s output embeddings, while the target network is updated by an exponential moving average of the online network’s weights. Through this procedure, the online network iteratively refines its representations, leveraging the target network as a slowly evolving reference. Remarkably, BYOL does not rely on contrastive objectives or negative pairs. Instead, it hinges on the student’s pursuit of the teacher’s representation, improving the features learned with each training iteration.


Other distillation-based algorithms, such as SimSiam \cite{chen2021exploring}, adopt a similar two-pathway configuration with minimal reliance on negative examples. Although these approaches have yet to be fully deciphered from a theoretical standpoint, they demonstrate that representation learning can be conducted effectively without contrasting pairs, provided the training framework imposes sufficient constraints to avoid trivial solutions. In general, distillation methods are well suited for scenarios where memory resources are limited, or batch sizes cannot be made large enough to support contrastive frameworks.

From a practical standpoint, the capacity to train stable models without large negative-pair sets confers a notable advantage, especially in specialized domains such as medical imaging. Hospitals and research facilities often have hardware constraints, and datasets may be smaller than in mainstream computer vision. With distillation, a model can still develop meaningful features by comparing variations of the same data rather than requiring many unrelated samples to serve as negatives.

Nonetheless, training these methods can be sensitive to the choice of data augmentations. Where contrastive methods explicitly compare distinct samples, distillation relies on subtle transformations that preserve relevant structure. If augmentations destroy clinically essential information, the model may not learn representations that align with real-world medical tasks. Thus, integrating domain-specific augmentation routines is critical for ensuring clinically meaningful features. Equally important is the challenge of initialization. Since distillation omits negative pairs, there is less inherent signal to steer the network away from degenerate mappings in the earliest training phase, thereby making initialization a significant factor in convergence stability. Despite these concerns, distillation remains a powerful alternative for data-scarce environments and has shown strong performance on medical imaging tasks where memory and labeling resources are limited.

Key factors associated with distillation strategies in self-supervised learning include:
\begin{itemize}
    \item \textbf{Negative Pairs}: The teacher–student setup inherently removes the requirement for negative pairs, eliminating the need for explicit contrast among unrelated samples.
    \item \textbf{Architecture}: These methods typically employ a dual-network configuration, consisting of online and target networks, which are updated using gradient descent techniques or moving averages.
    \item \textbf{Data Efficiency}: Distillation approaches are effective with smaller batch sizes, making them particularly suitable for environments with limited computational resources.
    \item \textbf{Sensitivity to Augmentations}: Careful selection of data augmentations is crucial to preserve domain-relevant features and avoid overly aggressive transformations that could distort key information.
    \item \textbf{Medical Imaging Benefit}: The reduced memory requirements and compatibility with small batch sizes make distillation methods advantageous in medical imaging scenarios, where data may be scarce and high-resolution images demand significant resources.
    \item \textbf{Limitations}: The theoretical foundations of these methods are still under investigation. There is a potential risk of representation collapse if the system is poorly initialized or if the chosen augmentations lead to ambiguous or misleading samples.
\end{itemize}

In essence, distillation methods demonstrate that, for representation learning, negative pairs are not strictly required if a training regime incorporates carefully chosen augmentations and a stable teacher–student interaction. This development opens new possibilities for advancing self-supervised learning, particularly in fields such as medical imaging, where dataset complexity and hardware constraints make large-scale contrastive approaches less feasible.

\begin{figure}
    \centering
    \includegraphics[width=0.9\textwidth]{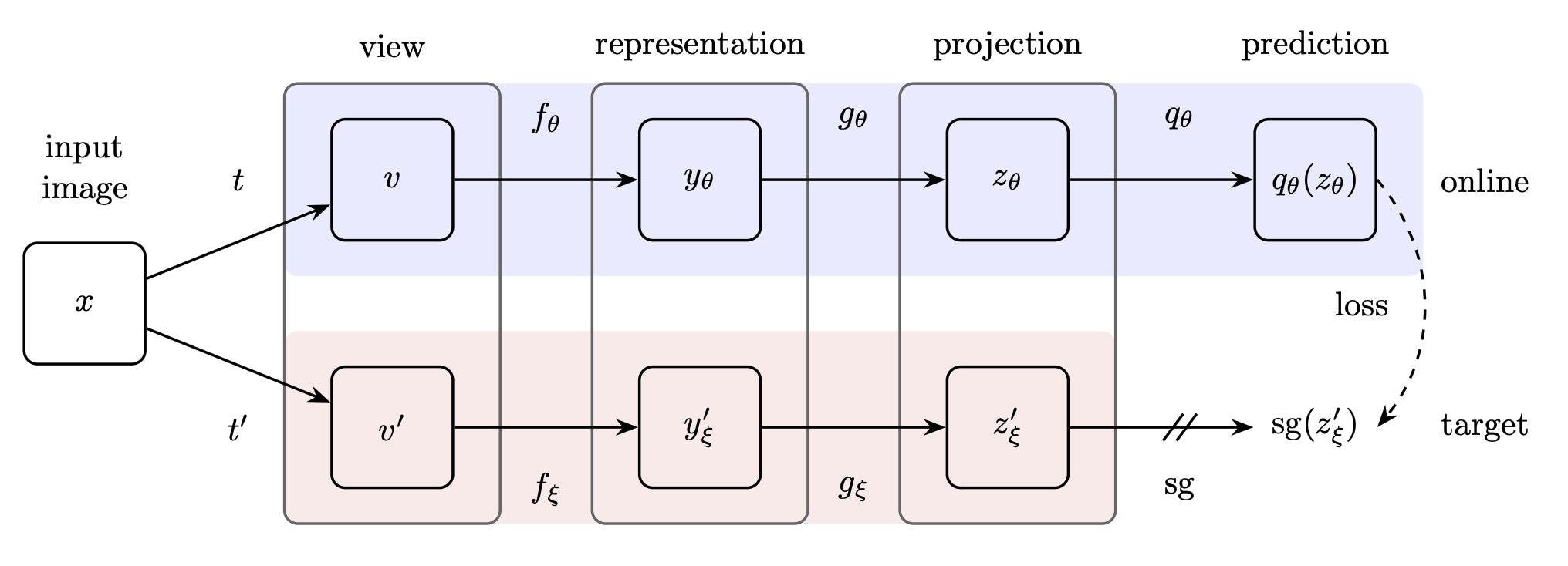}
    \caption{BYOL Architecture\protect\footnotemark}
    \label{fig:byol}
\end{figure}

\paragraph{BYOL.}
Bootstrap Your Own Latent (BYOL) introduces two parallel network pathways, known respectively as an online and a target network. Both pathways operate on slightly varied or augmented versions of the same input image. The online network is updated through standard gradient descent, generating an embedding for its input, while the target network's weights are refreshed through a slow-moving average of the online network’s parameters. During training, the online network attempts to align its output representation with that of the target network, which effectively acts as a slowly updated teacher. By repeatedly adjusting the online network to match the target’s predictions, the model learns robust representations without explicitly contrasting one sample against unrelated data. This structure, see in Figure \ref{fig:byol} circumvents the typical reliance on negative pairs, thereby avoiding some of the computational demands and memory constraints associated with methods like SimCLR. It also mitigates issues with visually similar samples being incorrectly treated as negatives, a frequent concern in medical imaging scenarios. Although the specific reasons for its stability remain an active topic of investigation, BYOL’s ability to learn from positive pairs alone makes it a valuable option for data-limited environments, including clinical contexts where gathering and annotating large datasets is challenging. 

\footnotetext{https://paperswithcode.com/paper/bootstrap-your-own-latent-a-new-approach-to-1}

\paragraph{General Strengths.}
Distillation methods provide certain advantages over more conventional self-supervised strategies that rely on explicit negative samples. Because they can be trained solely through positive pairs, they do not require large batch sizes to compare different samples, and this helps ease memory constraints during training. In many cases, their training process exhibits greater stability, particularly when working with datasets that contain high visual similarity, such as those commonly found in medical imaging. They also tend to show flexibility in handling smaller batch sizes, a feature that becomes crucial when dealing with large three-dimensional scans or computationally limited hardware. Taken together, these properties enable distillation-based methods to function effectively in environments where data and computing resources are constrained, making them well-suited for clinical contexts where both labeled data and high-end hardware may be limited.

\paragraph{General Weaknesses.}
Distillation methods, despite their practical benefits, also carry certain drawbacks that can influence their suitability in specific settings. Their reliance on a teacher-student setup introduces sensitivity to the choice of augmentations, as over-aggressive transformations risk obscuring vital details in medical images. Additionally, the theoretical underpinnings of distillation remain an active area of research, and the absence of contrastive pairs can lead to degenerate solutions if the system is not properly initialized or if the augmentations inadvertently remove key features. Another concern is the computational overhead required to maintain a separate teacher model, which can place considerable demands on limited hardware resources. Taken together, these aspects underscore the importance of carefully tuning the architecture, data preprocessing, and training hyperparameters when applying distillation-based frameworks, especially in data-scarce or hardware-constrained environments.

\paragraph{Medical Imaging Considerations.}
Distillation methods show particular promise in medical imaging because they do not depend on the formation of explicit negative pairs, which can be difficult to define in domains where images often share a high degree of visual similarity. Their reliance on positive pairs alone also eases memory constraints, as the batch size can remain relatively small when training with large, high-resolution clinical images. Nonetheless, the effective application of distillation to medical imaging requires carefully chosen augmentation pipelines. Transformations that severely alter subtle pathological cues may invalidate learned features, so the generation of different views must balance variety with the need to preserve clinically significant details. When successfully tailored to these concerns, distillation methods can help address data scarcity in healthcare applications and offer robust initialization for tasks such as organ segmentation or disease classification.

\subsection{Information Maximization Methods.}
Information maximization methods define a more recent branch of self-supervised learning. They seek to maximize mutual information between different representations of the same data while ensuring those representations do not collapse to trivial solutions. Unlike contrastive or distillation approaches, these methods do not rely on maintaining explicit negative examples or a teacher-student network. Instead, they enforce constraints that encourage each dimension of the model’s output to capture meaningful characteristics of the underlying data. One notable outcome is that the network generally learns both robustness to perturbations and diversity in its representation components.

Barlow Twins provides a clear illustration of this principle by using a loss function that attempts to align cross-correlation values of two augmented views with the identity matrix. The method takes two distorted versions of the same input and processes them through a shared encoder, then measures the cross-correlation across corresponding features. It penalizes deviations from the identity along the diagonal, encouraging outputs to remain similar for matching inputs and less redundant across different feature dimensions. Another example is VICReg \cite{bardes2021vicreg}, which similarly integrates constraints to preserve the variance of the outputs and minimize redundancy between different elements in the representation. These methods may help capture subtle structures in medical images, including pathologies that require distinct dimensions of variation, while forgoing the complexity of negative pair sampling or a separate teacher model. By focusing on redundancy reduction and variance maintenance, information maximization frameworks can potentially capture clinical features without discarding minor but critical details that might be essential for disease detection or anatomical localization.

\begin{figure}
    \centering
    \includegraphics[width=0.8\textwidth]{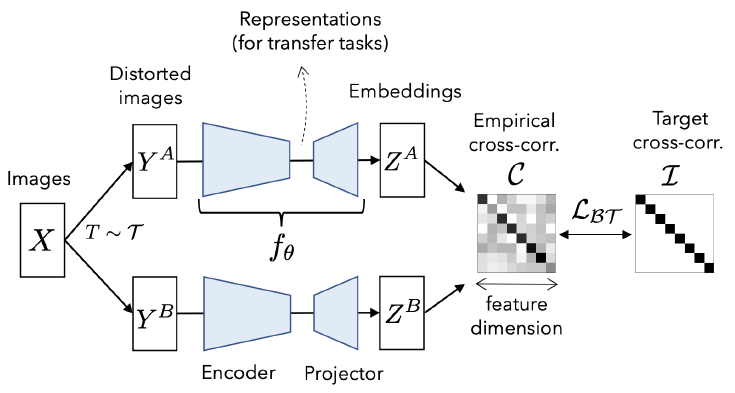}
    \caption{Barlow's Twins Architecture\protect\footnotemark}
    \label{fig:barlow}
\end{figure}

\paragraph{Barlow Twins.}
In the context of self-supervised representation learning, the Barlow Twins \cite{zbontar2021barlow} method presents an innovative framework aimed at reducing redundancy across the neural network’s output dimensions while simultaneously enforcing invariance to input distortions. The approach, see Figure \ref{fig:barlow} for the architecture, commences by generating two independently perturbed versions of a given input sample—typically through augmentations such as random cropping, color jittering, and geometric transformations—and processing these altered instances via a shared encoder. The loss function, $\mathcal{L_{BT}}$ is:
\begin{equation}
\mathcal{L_{BT}} \triangleq  \underbrace{\sum_i  (1-\mathcal{C}_{ii})^2}_\text{invariance term}  + ~~\lambda \underbrace{\sum_{i}\sum_{j \neq i} {\mathcal{C}_{ij}}^2}_\text{redundancy reduction term}
\label{eq:lossBarlow}
\end{equation}
where $\lambda$ is a trade-off variable between the first and second terms of the loss, and $\mathcal{C}$ is the cross-correlation matrix. This results in two latent representations that, despite the stochastic modifications, retain the essential semantic content of the original data. Subsequently, a cross-correlation matrix is computed between these representations. The central objective is to drive this matrix to approximate the identity matrix as closely as possible. In practice, this entails two complementary constraints: first, the diagonal elements of the cross-correlation matrix are encouraged to converge towards unity, thereby ensuring that the representations remain invariant to the applied distortions; second, the off-diagonal elements are penalized if they deviate from zero, which in effect compels each component of the feature vector to capture distinct, non-overlapping aspects of the input signal. This dual emphasis on invariance and redundancy reduction is theoretically grounded in principles of information maximization and has been demonstrated across our prior work to yield representations that are robust and discriminative. In medical imaging applications, where subtle anatomical variations and complex pathologies demand precise and diverse feature encoding, such a mechanism is particularly advantageous. By explicitly discouraging redundant encoding while preserving the integrity of the underlying signal, the Barlow Twins framework facilitates the learning of feature embeddings that are both resilient to variations in imaging conditions and capable of capturing the intricate details necessary for accurate clinical interpretation.

\footnotetext{https://sh-tsang.medium.com/review-barlow-twins-self-supervised-learning-via-redundancy-reduction-967769fafc1}

\paragraph{General Strengths.}
One significant advantage of information maximization techniques is their robust theoretical foundation. By formulating objectives that directly maximize the mutual information between representations, these methods are underpinned by a rigorous mathematical framework that ensures a principled approach to redundancy reduction. This formulation inherently promotes the learning of diverse, yet complementary, feature dimensions. Moreover, the absence of an explicit requirement for negative samples simplifies both model architecture and training dynamics, reducing the complexity associated with negative pair selection and management. Empirical studies—including our own work on dynamic batch adaptation and backforward propagation—have demonstrated that these approaches offer an elegant and conceptually straightforward means of capturing invariant features while maintaining stability during training.

\paragraph{General Weaknesses.}
Notwithstanding their theoretical appeal, information maximization methods are accompanied by several practical challenges. In particular, achieving stable estimates of the required statistical measures often necessitates large batch sizes, which may be impractical in certain scenarios. The computational demands associated with high-dimensional matrix operations further compound this issue, as the cost of computing cross-correlation matrices can be substantial. Additionally, while the theoretical underpinnings are well established, there remains a limited comprehensive understanding of the precise optimization dynamics encountered in practice. This gap can render hyperparameter tuning more arduous and may lead to variability in convergence behavior across different datasets and network architectures.

\paragraph{Medical Imaging Considerations.}
When applied to medical imaging, information maximization methods offer distinct benefits by virtue of their focus on reducing feature redundancy while preserving clinically critical details. Medical images, such as those obtained from MRI and CT modalities, frequently embody subtle anatomical and pathological variations that are essential for accurate diagnosis. The capacity of these methods to capture fine-grained, invariant features aligns well with the need to detect subtle differences that may distinguish healthy tissue from pathological conditions. However, the inherently high dimensionality and complexity of medical imaging data introduce additional challenges. In particular, the computation of cross-correlation matrices in high-dimensional spaces may prove computationally intensive and sensitive to the choice of feature representation. Consequently, it is imperative that these methods be carefully adapted to ensure that the preserved information is not only statistically robust but also clinically relevant. Future investigations should focus on refining these techniques to better accommodate the unique characteristics of medical data, thereby improving both computational efficiency and diagnostic utility.

\subsection{Cross-Modal Methods}

In recent years, a growing body of research has focused on leveraging the inherent correlations between heterogeneous data modalities to learn robust representations without the need for extensive manual annotation. Cross-modal methods, as examined in our previous work on multitask multimodal self-supervised learning, capitalize on the natural co-occurrence of disparate data streams—such as visual, textual, and auditory information—to uncover latent structures that are shared across modalities. By aligning representations derived from these distinct yet complementary sources, such approaches establish a unified embedding space that reflects both the low-level signal characteristics and the high-level semantic content inherent in the data.

\paragraph{Image-Text Matching.}
A primary instance of cross-modal learning involves the alignment of visual and textual modalities. In image-text matching, models are trained to project both images and their corresponding textual descriptions into a shared latent space, such that semantically congruent pairs are drawn closer together while incongruent pairs are pushed apart. This is achieved through the careful design of loss functions that quantify the similarity between image and text embeddings, thereby enforcing a correspondence that captures not only the detailed visual features but also the contextual and conceptual nuances articulated in accompanying text. Our prior investigations have demonstrated that such an approach—when augmented with self-supervised pretext tasks and refined through domain-specific adaptation—can yield visual features that are sensitive to the subtle diagnostic cues present in medical images, as well as to the descriptive language found in clinical documentation. In this manner, the joint representation facilitates enhanced interpretability and supports downstream tasks such as automated report generation and clinical decision support.

\paragraph{Audio-Visual Learning.}
Analogous to the image-text paradigm, audio-visual learning methods exploit the temporal and spatial synchrony between visual content and auditory signals. In settings where visual data is accompanied by audio—whether in the context of video analysis or in specialized applications such as patient speech assessment—models are designed to capture the correlations between these modalities. The learning objective typically involves constructing a joint representation that is robust to modality-specific noise and variations, thereby ensuring that features derived from the visual stream are meaningfully complemented by corresponding audio cues. Our work in this area has revealed that such multimodal integration, when guided by carefully designed self-supervised objectives, can significantly enhance the discriminative power of the learned representations. This is particularly relevant in clinical environments, where the synthesis of visual and auditory information may improve the accuracy of diagnostic systems and contribute to more comprehensive patient assessments.

Collectively, these cross-modal strategies underscore the potential of self-supervised learning to transcend the limitations imposed by scarce labeled data and domain-specific variability. By harnessing the natural interdependencies among different data sources, the resulting models not only achieve superior generalization but also offer a more holistic understanding of the underlying phenomena—a quality that is essential for advanced applications in medical imaging and beyond.

\section{Summary}
In this chapter, we have presented a comprehensive overview of the foundational principles and technological advancements that underpin contemporary deep learning and its application to medical imaging. We began by tracing the evolution of artificial neural networks from their early inception—characterized by simple, single-layer perceptrons—to the complex, multi-layered architectures that form the basis of modern deep learning. The discussion highlighted how the hierarchical structure of these networks enables the progressive abstraction of raw data into rich, semantically meaningful representations, a feature that is indispensable in addressing the intricacies of clinical diagnostic imaging.

We then examined the pivotal role of convolutional neural networks in capturing spatial hierarchies and extracting fine-grained features from images. Their ability to reduce parameter complexity through weight sharing and to mitigate overfitting via mechanisms such as residual connections has made CNNs the cornerstone of many successful medical imaging applications. Moreover, the emergence of recurrent neural networks, despite their inherent sequential limitations, was discussed as an essential paradigm for modeling temporal and volumetric data—attributes that are crucial when analyzing longitudinal patient scans or dynamic imaging studies.

The recent advent of Transformer architectures has further expanded the deep learning toolkit by introducing self-attention mechanisms that capture global dependencies and contextual relationships in data. Although these architectures require substantial computational resources, their ability to process information in parallel and to integrate multimodal data streams has opened new avenues for both diagnostic accuracy and interpretability in medical applications. As our research has demonstrated, combining such architectures with self-supervised learning strategies—such as predictive pretext tasks, generative reconstruction, contrastive embedding, and information maximization—can significantly alleviate the constraints imposed by scarce and heterogeneous labeled data.

Self-supervised learning, in particular, has emerged as a robust and data-efficient framework by exploiting the latent structure of medical images. Through the formulation of pretext tasks—ranging from geometric reconstruction to the synthesis of novel samples via generative models—this paradigm not only reduces the reliance on extensive expert annotation but also enhances the model’s ability to generalize across varied imaging protocols and clinical conditions. The integration of these techniques with specialized methods such as Efficient Dynamic Batch Adaptation, Backforward Propagation, BrainFuse for data fusion augmentation, Cascading Sum Augmentation, and advanced distillation methods has further advanced our capacity to train deep models under challenging conditions.

In conclusion, the prerequisites discussed in this chapter serve as the theoretical and methodological bedrock upon which the subsequent chapters build. They provide both the historical context and the technical rationale for employing advanced deep learning techniques in the realm of medical image analysis. By synthesizing decades of research in neural network architectures, self-supervised learning, and domain-specific adaptations, we have established a comprehensive framework that not only informs but also justifies the innovative approaches developed in the later stages of this thesis. As we move forward, the insights gleaned here will guide our exploration of domain adaptation, multimodal fusion, and privacy-preserving strategies, all aimed at realizing intelligent, robust, and clinically impactful diagnostic systems.
\chapter{Advanced Self-Supervised Learning Techniques in Medical Imaging}
\label{chap:ssl_medimg}

\ifpdf
    \graphicspath{{Chapter3/Figs/Raster/}{Chapter3/Figs/PDF/}{Chapter3/Figs/}}
\else
    \graphicspath{{Chapter3/Figs/Vector/}{Chapter3/Figs/}}
\fi

\section{Overview}
\label{sec:overview_ssl_medimg}

In the previous chapter (Chapter~\ref{chap:prereqs}), we introduced the principal categories of self-supervised learning (SSL) methods, which have become indispensable in computer vision whenever annotated data are limited. These families of approaches, including predictive tasks, generative modeling, contrastive learning, clustering-based pseudo-labeling, distillation, and information maximization, all aim to leverage unlabeled datasets by crafting goals (or \emph{pretext tasks}) that the model must solve without relying on explicit human-provided annotations. Although these broad classes of SSL techniques have shown remarkable success in natural-image settings, medical imaging brings a distinct set of constraints that influence how SSL should be designed and applied. Clinical images are often three-dimensional or even four-dimensional, contain subtle yet critical pathologies, and are governed by strict privacy regulations that complicate data-sharing and hamper the development of large annotated corpora.

This chapter expands upon the \emph{advanced} methods that have emerged to adapt SSL for medical imaging tasks. We focus in particular on tasks and architectures tailored to volumetric data, multi-modal acquisition protocols, and domain-specific transformations rooted in clinical knowledge. In Section~\ref{sec:ssl_quick_recap}, we briefly revisit the canonical SSL paradigms introduced earlier, underscoring elements that must be reconsidered when dealing with high-dimensional scans, data scarcity, and multi-institutional variability. Then, in Section~\ref{sec:novel_pretext_medimg}, we dive into specialized pretext tasks developed specifically for medical data, including 3D jigsaw puzzles, slice ordering, cross-modal synthesis, artifact inpainting, and masked autoencoders optimized for the complexities of volumetric or multi-sequence imaging. Next, we address strategies for incorporating SSL into broader clinical pipelines (Section~\ref{sec:ssl_integration}), examining large-scale pretraining with in-domain medical images, federated learning for privacy-preservation, interpretability requirements, and semi-supervised frameworks that unify unlabeled and labeled data. Finally, we conclude with a forward-looking discussion on challenges that still impede the widespread adoption of SSL-based solutions in clinical practice, along with potential directions for future innovation. 

\section{Quick Recap of Core SSL Methods}
\label{sec:ssl_quick_recap}

Self-supervised learning leverages auxiliary or \emph{pretext} objectives designed to reveal structure from unlabeled data. The typical rationale is that, by solving an artificially generated task, a network must capture salient features that ultimately facilitate downstream supervised tasks such as classification or segmentation. Chapter~\ref{chap:prereqs} enumerated multiple well-known SSL approaches, from relatively straightforward transformations (e.g., jigsaw puzzles, rotation classification, inpainting, and patch reordering) to more advanced strategies like contrastive instance discrimination and knowledge distillation.

In medical imaging, these core ideas must be carefully revisited because of the data’s higher dimensionality, the frequent mismatch in scanner protocols across institutions, and the need to capture subtle pathological markers that might be lost if the pretext task is too coarse. Whereas rotation prediction in natural images may rotate an object arbitrarily by 90° increments, in medical scenarios, a 180° rotation of a CT scan might not even be physiologically valid. Likewise, tasks that rely on large negative sets in contrastive learning become precarious when dealing with a hospital’s dataset of very visually similar patient scans, prompting the development of alternative forms of sampling or domain metadata usage. These considerations lie at the heart of advanced SSL for medical imaging, as the field strives to move beyond naive adaptation of generic computer vision tasks.

\section{Novel Pretext Tasks for Medical Imaging}
\label{sec:novel_pretext_medimg}

While basic jigsaw puzzles, rotation tasks, and inpainting strategies can be ported from natural images to clinical data, a growing body of work has introduced more refined or complex SSL tasks that align better with volumetric scans, multi-modal acquisition, and the intricacies of pathology identification. This section provides an in-depth analysis of these newly developed or heavily modified pretext strategies and explains why they have gained traction in the medical domain.

\subsection{3D Jigsaw Puzzles and Volumetric Slicing}

The idea of a jigsaw puzzle in self-supervised learning emerged as an elegant way to force a model to learn spatial structure. Researchers soon realized, however, that typical 2D jigsaw approaches underestimated the complexity of medical scans. Instead, entire 3D volumes from computed tomography or magnetic resonance imaging can be partitioned into “cubic” sub-volumes, scrambled, and subsequently reassembled by the network~\cite{zhuang2019self, taleb20203d}. This volumetric puzzle forces the model to pay attention not only to planar information but also to slice-to-slice continuity, an essential factor when dealing with multi-slice organ boundaries or tumor infiltration across slices. 

In dental CT, for instance, volumetric jigsaws have been used to detect early carious lesions in tooth structures by taking advantage of the fact that a carefully scrambled 3D puzzle can highlight morphological continuity from healthy tissue to decayed regions~\cite{TajbakhshDental2023}. Similar strategies have been effective in brain MRI, where subtle changes in volume or shape across slices often carry diagnostic significance. The complexity of these jigsaw tasks can be tuned by altering how finely the volume is subdivided, or by mixing puzzle permutations with random rotations or cutout artifacts, thereby increasing the difficulty of the puzzle. In doing so, the network is compelled to capture more holistic knowledge about anatomical geometry. One noted limitation is the exponential increase in possible permutations. Hybrid approaches sometimes reduce puzzle permutations to a curated set, balancing complexity and learnability. 

\subsection{Slice Ordering and Time-Sequence Reconstruction}

In many three-dimensional imaging sequences, the collection of slices follows a consistent spatial or temporal order. Some SSL frameworks harness this ordering by randomly shuffling slices or sub-volumes and training the network to recover the original sequence. Early demonstrations in brain MRI showed that a model forced to predict the correct cranial-to-caudal order of slices developed stronger encoders for tumor or lesion segmentation~\cite{zhang2017self}. When extended to longitudinal imaging, the concept can be generalized to time-sequence reconstruction, where repeated scans of the same patient over months or years are shuffled. The SSL objective to reorder these scans can help the model detect mild changes over time, such as atrophy or evolving lesions, that might elude simpler pretext tasks. 

This “slice ordering” approach often complements other predictive strategies such as small-angle rotation classification, shaping an SSL curriculum that progressively introduces a combination of geometric and positional constraints. In some recent works, slice ordering has also been fused with partial reconstruction tasks in a multi-task SSL arrangement. For example, the network might simultaneously restore missing slices and place them in correct sequence, effectively capturing both continuity and morphological details in a single training pass.

\subsection{Cross-Modality Synthesis and Reconstruction}

Multi-modal imaging is particularly valuable in clinical practice, with modalities like MRI, CT, and PET each highlighting different tissue properties or functional activity. SSL tasks that require synthesizing one modality from another encourage the network to learn cross-modal correspondences. For instance, a network might be trained to generate PET from MRI or T2-weighted MRI from T1-weighted scans~\cite{ben2019cross, islam2020gan}. Since no explicit supervision is needed (the scans come from the same patient), the model’s ability to produce a plausible version of the target modality reveals it has extracted key structural and functional representations that span both domains. 

One advantage of such cross-modality tasks is that they inherently handle “missing scan” scenarios: a patient lacking a certain modality can potentially have it synthesized for approximate analysis. Additionally, introducing a contrastive element into cross-modality generation can further refine the embeddings. For example, the generative pipeline might produce synthetic CT from an MRI while a contrastive loss ensures that real CT and synthesized CT have similar embeddings, aligning the two modalities in a shared latent space. These cross-modal frameworks reduce reliance on large labeled datasets, as they simply require matched but unlabeled scans from each modality.

\subsection{Inpainting and Artifact Removal in 3D or 4D Volumes}

Inpainting tasks, wherein patches or entire slices are masked out and the network must reconstruct them, represent an early style of self-supervision. In medical imaging, these tasks require modifications to address real-world imaging artifacts or missing slice problems. For instance, patients moving during acquisition can create motion artifacts or partial scans. SSL methods that treat these artifacts as “noise or masked slices” and train the model to restore them yield powerful representations capable of coping with suboptimal or incomplete data. In cardiac MRI, a model performing volumetric inpainting can remain robust to dropped frames in cine sequences, subsequently providing better ventricle delineation or ejection fraction estimates. 

Such tasks also naturally improve domain adaptation, as they force the network to learn invariances to typical scanning inconsistencies encountered across institutions. A model that frequently inpaints missing regions from different hospital scanners or varied slice thickness is compelled to develop a domain-agnostic representation of normal anatomy. Nevertheless, careful design is essential to ensure the inpainting puzzle is neither too trivial nor impossible—very large masked areas could hamper the network’s ability to glean clinically significant details.

\subsection{Rotation, Scaling, and Tilt Predictions for Clinical Alignment}

In the general computer vision setting, rotation prediction often involves angles of $0^\circ$, $90^\circ$, $180^\circ$, or $270^\circ$~\cite{gidaris2018unsupervised}. Medical scans, however, may follow standard axial, coronal, or sagittal planes, making large rotations less realistic or even meaningless. Consequently, some frameworks define smaller tilt angles or scale changes that replicate typical misalignments or differences in scanning protocols. A model that accurately predicts these transformations becomes robust to how a particular study was acquired, improving inter-institutional generalization. This local rotation or tilt approach has been demonstrated in tasks such as knee MRI classification, where subtle variations in how the knee is oriented in the scanner can significantly influence downstream detection or grading of meniscal tears.

\subsection{Frame Interpolation for Dynamic (4D) Imaging}

Certain imaging studies, such as echocardiograms or cine-MRI of the heart, produce temporal sequences in addition to the spatial dimensions. Introducing an SSL objective that predicts an intermediate frame given preceding and following frames ensures sensitivity to the time dimension~\cite{JiaoUltrasoundPreprint}. Echocardiographic sequences often exhibit cyclical motion from systole to diastole; by reconstructing or interpolating missing frames, the model hones in on normal movement patterns. This knowledge can be leveraged for pathology detection in abnormal cardiac dynamics or for segmenting heart chambers when labeling data is limited. Researchers have expanded on this approach by combining frame interpolation with spatial transformations, pushing the network to handle both volumetric geometry and temporal progression.

\subsection{Contrastive SSL with Metadata-Guided Sampling}

Although standard contrastive learning has shown great success, the homogeneous nature of many clinical datasets introduces a risk that random negatives will be too visually similar, undercutting the representation’s discriminative power. Medical researchers have therefore tested specialized sampling methods. For instance, a model might gather multiple scans of the same patient (or the same lesion) across time as “positives,” ensuring that repeated but unlabeled follow-up studies are recognized as belonging to the same underlying distribution. Meanwhile, strong negative diversity can be achieved by actively selecting scans from different organs, scanners, or demographic groups, guided by patient metadata. This approach can mitigate mode collapse or trivial solutions, especially in smaller datasets. Large-scale endeavors like LVM-Med~\cite{LVMMed2023} highlight how contrastive SSL can be scaled to over a million images by carefully defining negative sampling across diverse patient populations, enabling a universal encoder that addresses tasks from liver lesion detection to spinal abnormality classification.

\subsection{Distillation and Landmark-Augmented SSL}

Distillation-based SSL, typified by BYOL or SimSiam, removes the explicit contrastive element by updating a teacher network via an exponential moving average of a student network’s parameters. In medical applications, researchers have introduced \emph{landmark augmentation}, where certain anatomical landmarks or bounding boxes for organs are introduced into the teacher updates~\cite{KeypointAugSSL2023}. This technique compels the student to encode these crucial points in the latent space. For tasks like the apex of the left ventricle, the teacher might be given minimal annotated data or approximate priors, which the student is expected to respect while solving a masked or jigsaw-like pretext. This synergy often yields encoders more attuned to domain-specific geometry, bridging the gap between purely unsupervised learning and medically guided constraints.

\subsection{Information Maximization in 3D and Multi-Modal Spaces}

Barlow Twins and VICReg revolve around aligning cross-correlation matrices to the identity, thus ensuring that the embeddings of two augmented views coincide on the diagonal while reducing redundancy on the off-diagonals~\cite{zbontar2021barlow, bardes2021vicreg}. These approaches have begun to see adoption in volumetric scans by constructing 3D augmentations (e.g., random flips, small rotations, or partial sub-volume sampling). In multi-modal expansions, correlated pairs might be slices from two MRI sequences that capture complementary tissue properties. By enforcing high correlation for the same anatomical region across modalities and low correlation between distinct feature dimensions, the model retains a broad yet consistent perspective of the data. The main computational challenge is that correlation matrix calculations can be expensive for large volumetric data, so memory-saving strategies such as gradient checkpointing or partial sample selection are often deployed. The advantage, however, is that the learned embeddings are robust and do not rely on large negative sets or teacher–student duplication.

\subsection{Masked Autoencoders for Medical Imaging}

Mask-based reconstruction has recently sparked significant interest in medical SSL, particularly for resource-limited tasks where only a handful of labeled scans exist. Building on principles from masked language modeling (e.g., BERT) and the success of MAEs in vision, medical masked autoencoders have shown the ability to isolate salient anatomy while ignoring uninformative details. SMIT~\cite{jiang2022smit} applies a masked patch approach to 3D volumetric data for better segmentation, while MedMAE~\cite{gupta2024medmae} focuses on CT multi-organ analysis. By randomly occluding portions of the input volume, these models encourage the encoder to fuse local details with broad context. 

Care must be taken, however, to ensure that small lesions are not always completely masked out, rendering them invisible during training. Researchers have adopted various sampling heuristics, such as partial sub-volume masking or “mask in masking,” in which multiple levels of occlusion are combined to preserve a fraction of critical features. The net result is that masked autoencoders can yield encoders that consistently outperform or match contrastive methods when the labeled set is extremely small. Their generative nature also provides a more transparent output—masked regions that are poorly reconstructed in test time might indicate anomalies.

\section{Integrating SSL with Medical Imaging Pipelines}
\label{sec:ssl_integration}

Adapting self-supervised learning to medical imaging is not merely a matter of designing sophisticated pretext tasks. It also demands comprehensive strategies for large-scale pretraining, privacy-preserving collaboration, interpretability, domain adaptation, and the efficient use of partial annotations. This section reviews how advanced SSL approaches slot into wider clinical or research-based pipelines.

\subsection{Large-Scale Medical Pretraining and Foundation Models}

A compelling trend involves collecting large volumes of unlabeled medical scans from multiple institutions and modalities to train \emph{generalist} encoders via SSL, an approach akin to building domain-specific “foundation models.” Projects like LVM-Med~\cite{LVMMed2023} illustrate how contrastive learning can be scaled to over a million images encompassing CT, MRI, and X-ray scans. Similarly, efforts such as UniMiSS~\cite{UniMiSS2022} leverage transformers with switchable patch embeddings for 2D and 3D volumes, indicating an appetite for more flexible architectures that unify multiple scanning protocols. In parallel, MedMAE~\cite{gupta2024medmae} demonstrates that large-scale masked modeling can yield self-supervised backbones especially powerful in 3D multi-organ segmentation.

Empirical findings show that these in-domain encoders, once pre-trained on unlabeled data, often exceed the performance of standard ImageNet transfer learning across classification and segmentation tasks. Moreover, these models can drastically reduce the required labeled samples for comparable performance, which is vital in fields like radiology or pathology where annotation demands can be prohibitively large. Nevertheless, training such large-scale encoders faces challenges in data collection logistics, domain shift management, and computational overhead, leading many hospitals to investigate distributed or federated solutions.

\subsection{Federated Self-Supervised Learning and Privacy}

Federated learning has become an influential paradigm for medical AI because it sidesteps the need to centralize patient data, which can be restricted by privacy laws (HIPAA, GDPR, etc.). In federated SSL, each site performs local self-supervision on its unlabeled scans, then shares only model updates to a central aggregator. This approach preserves local data privacy while collectively amassing the representational knowledge gleaned from multiple institutions~\cite{Azam2023}. Variants of federated SSL incorporate encryption methods, differential privacy, or domain-adversarial objectives to ensure that no single site’s data distribution is exploited or leaked. 

One potential obstacle is domain heterogeneity. Different hospitals may use different scanner types, protocols, or patient demographics, producing divergences in intensity values or image orientation. Without care, these variations can hamper the convergence of the global model. Techniques such as per-institution normalization layers, domain alignment modules, or partial model fusion have been proposed to mitigate these shifts. Another open question is whether local SSL tasks should be uniform across sites or custom-tailored to each site’s data. Studies suggest that a uniform approach is simpler to implement but may not capture site-specific scanning artifacts, whereas site-tailored pretext tasks risk complicating the global aggregation step.

\subsection{Test-Time Adaptation to New Domains}

Once an SSL-pretrained model is deployed, it may encounter unexpected domain shifts such as new scanner hardware, different patient populations, or specialized imaging protocols. Recent work explores \emph{test-time adaptation} (TTA) via continuing self-supervision on the unlabeled test scan itself~\cite{LeeTTA2023}. For instance, if the model is confronted with a slight rotation or mismatch in pixel intensities, it can solve a small puzzle or inpainting task on the fly, updating a subset of its parameters to better align with the new domain. This can yield immediate benefits in classification or segmentation accuracy for the single test instance, though it requires additional computational overhead at inference. Evaluations in ultrasound segmentation highlight the promise of TTA for bridging domain gaps, but caution is required when TTA is extended to real-time or large-volume analyses.

\subsection{Interpretability and Clinical Trust}

Beyond raw performance metrics, the acceptance of SSL-based models in medical settings depends on transparency. Tasks like masked modeling or generative reconstruction provide partial interpretability because the reconstructed outputs can be visually compared against the real scans, indicating whether the model has indeed learned clinically relevant features or is ignoring subtle anomalies. Contrastive pipelines may rely more heavily on saliency maps, attention-based visualizations, or region-level embeddings to identify which areas are most pivotal in the network’s representation.

In certain frameworks, interpretability arises naturally from multi-task SSL designs, where partial or complete anatomical segmentation is also learned. If a self-supervised objective includes the reconstruction of organ boundaries or major landmarks, the resulting features align more transparently with medical knowledge. Clinicians are often more receptive to a model that clearly highlights which tissues it found important. However, balancing interpretability with data efficiency remains a challenge: the introduction of additional tasks can demand larger memory or more labeling for partial supervision. 

\subsection{Semi-Supervised Learning and Hybrid SSL Pipelines}

While SSL alone can reduce the need for labeled samples, many institutions do hold a small annotated set. A natural extension is to merge self-supervision on the unlabeled portion with standard supervised learning on the labeled portion in a single framework. This approach often outperforms a naive pipeline where SSL is done first and the model is merely fine-tuned later, because the model continues to refine feature embeddings as it gains partial ground truth feedback. Iterative pseudo-labeling is a related concept, where the SSL-pretrained model is used to generate approximate labels for unlabeled scans, which are then integrated into the supervised pipeline to refine future training cycles. In areas like multi-organ segmentation or rare pathology detection, such strategies have been shown to significantly boost performance while preserving an economical annotation budget~\cite{singh2023shifting, PathologySSL2023}.

\section{Looking Ahead: Challenges and Opportunities}
\label{sec:ssl_medimg_future}

Although self-supervised learning has propelled medical imaging forward, important limitations remain. First, the creation of large multi-institutional unlabeled datasets is still fraught with legal and logistical barriers. Federated SSL can address some issues, but domain mismatch often persists, complicating model convergence. Methods that effectively unify extremely varied scanning protocols within a single representation are still under exploration. Second, interpretability and reliability standards in medicine require more robust validation. Straightforward reconstructions or saliency maps may not suffice for tasks that demand clear region-level decisions, such as precisely delineating malignant tissue for potential surgery. Future directions include building SSL tasks that natively incorporate shape priors, region-based objectives, or known clinical landmarks to produce more interpretable latent spaces.

Another frontier is the adaptation of SSL to more exotic or large-scale modalities: a single whole-slide image in pathology can contain gigapixels of data, typically subdivided into patches for analysis. SSL approaches that manage these high resolutions efficiently could unlock morphological patterns relevant to cancer staging. Meanwhile, cross-modal expansions continue to be of interest, mixing X-ray with textual radiology reports, or combining MRI with genomic data. The synergy of textual data could yield powerful vision–language models for radiology akin to the CLIP or BLIP families, but these expansions must also tackle specialized medical language and handle the complexities of partial data. 

Lastly, theoretical questions persist regarding which SSL tasks best capture subtle pathological markers. One scenario might call for strongly generative tasks if the anomalies are visually distinctive, while another could benefit from contrastive negative sampling to highlight differences in highly uniform data. Masked autoencoders might shine with extremely limited labeled sets but may underperform on tasks requiring discriminative precision for specific pathologies. Identifying robust heuristics or theoretical insights for selecting SSL tasks is an open research avenue that could streamline the design of domain-optimized pipelines. 

\section{Summary}
\label{sec:ssl_medimg_summary}

This chapter examined advanced self-supervised learning approaches as adapted to the demands of medical imaging. Departing from the standard 2D tasks commonly studied in general computer vision, the techniques discussed here emphasize volumetric continuity, multi-sequence integration, domain-guided transformations, and tasks specifically geared to the challenges of high-dimensional scans. We reviewed how jigsaw puzzles, slice ordering, cross-modality reconstruction, specialized rotation and tilt, frame interpolation, and masked modeling can be combined with large-scale contrastive or distillation-based SSL to create more domain-consistent representations. We also described how these frameworks integrate within privacy-preserving or federated pipelines, highlighted issues of interpretability and on-the-fly adaptation, and underscored how semi-supervised learning can exploit both unlabeled and labeled images together.

Many empirical studies confirm that such customized SSL significantly reduces the need for expert annotations while yielding robust performance on core tasks like classification, segmentation, and anomaly detection. Continued innovation in specialized pretext tasks, multi-modal expansions, interpretability, and theoretical underpinnings is poised to accelerate SSL adoption in clinical practice. Indeed, if challenges of large-scale data sharing, domain mismatch, and interpretability can be effectively addressed, the next generation of “foundation models” trained via SSL on massive unlabeled medical corpora may fundamentally shift how we develop and deploy computer-aided diagnosis systems. The coming chapters delve further into specific experiments and advanced architectures that push these boundaries, illustrating the tangible impact of SSL in real clinical scenarios.

\chapter{Medformer: A Multitask Multimodal Foundational Model for Medical Imaging}
\label{chap:medformer}

\ifpdf
    \graphicspath{{Chapter4/Figs/Raster/}{Chapter4/Figs/PDF/}{Chapter4/Figs/}}
\else
    \graphicspath{{Chapter4/Figs/Vector/}{Chapter4/Figs/}}
\fi

This chapter presents \textit{Medformer}, a unified framework for multitask and multimodal medical image analysis. In essence, Medformer unites diverse image data---encompassing both two-dimensional (2D) and three-dimensional (3D) scans, as well as various clinical domains---within a single deep learning architecture built on advanced transformer components. The primary motivation is to address the heterogeneity of medical imaging: each dataset often differs in modality, resolution, anatomical coverage, and clinical objective. By systematically integrating these varied data sources into a shared model, Medformer aspires to learn more robust representations that can be transferred or fine-tuned for downstream tasks even with limited labeled data.

A key aspect of Medformer’s design is its use of separate modules to handle input adaptation, a central model for feature abstraction, and output adaptation. The input module dynamically incorporates domain-specific clues about dimensionality (2D vs.\ 3D), imaging modality (CT, X-ray, microscopic, etc.), and anatomical region (chest, abdominal, brain, and more) via learnable latent embeddings. This allows the system to transform raw images into a standardized form, regardless of their original shape or modality. The central model then processes this canonical representation in a uniform way, supporting multiple self-supervised or supervised objectives. Finally, the output module translates the learned representations into task-specific predictions, such as classification logits or segmentation masks, again guided by specialized latent embeddings for each task or dataset.

In what follows, we discuss the motivations guiding Medformer’s development, review relevant background on self-supervision and transformer architectures, and then examine the core principles behind the input and output Adaptformers. We also introduce the underlying training procedures, which intermix large-scale self-supervised learning and task-focused fine-tuning. The results in this chapter, based on experiments with multiple 2D and 3D datasets, indicate that Medformer can serve as a flexible foundation for medical imaging tasks, benefiting from a more integrated perspective on seemingly disparate domains.

\section{Motivation}
\label{sec:medformer_motivation}

The landscape of medical imaging is remarkably diverse, encompassing numerous imaging modalities, anatomical regions, dimensionalities, and clinical tasks. This diversity presents a significant challenge for deep learning models, which often specialize in a single modality (e.g., two-dimensional X-ray images) or a single task (e.g., segmentation). In many cases, a model trained on one medical domain fails to generalize when exposed to another, particularly if data from different body parts, different imaging protocols, or varied dimensional configurations (2D scans versus 3D volumes) are required. As a result, clinical practitioners are frequently forced to maintain multiple specialized systems, each handling only one type of problem, making it difficult to integrate these systems into a coherent workflow.

Medformer is motivated by the vision of a unifying architecture that can reduce the proliferation of single-purpose networks by learning from—and adapting to—a broad range of medical image domains. This perspective acknowledges that, although tasks and modalities may differ substantially, they nonetheless share underlying patterns related to organ anatomy, pathology, and the physics of image acquisition. For instance, a three-dimensional CT scan of the brain, a two-dimensional chest X-ray, and a microscopic pathology slide all capture evidence of structural or tissue-level abnormalities, albeit in different coordinate spaces and with varying types of contrast. Medformer exploits such overlaps by providing a general-purpose core transformer model that is customized at runtime to each domain or task through a set of learnable latent embeddings and flexible input-output adaptation modules.

This approach is especially pertinent when labeled data are limited. Collecting large annotated cohorts in medical imaging can be constrained by cost, institutional privacy, or the rarity of certain pathologies. By aligning multiple sources of data within one architecture, Medformer seeks to increase both data and parameter efficiency, offering the promise that features learned from one area—say, volumetric organ segmentation—may inform classification tasks on two-dimensional X-rays, or vice versa. Through self-supervised or semi-supervised strategies, the model can accumulate broad representational knowledge across many medical domains, thus alleviating the need for extensive supervision in each individual setting.

In essence, the motivation behind Medformer stems from the aim to move beyond single-task, single-modality models, toward a versatile system capable of joint learning across diverse medical imaging scenarios. By harnessing latent embeddings that encode domain-specific priors (such as “2D,” “3D,” “X-ray,” or “abdominal CT”), the network preserves modality and body-part-related distinctions in a unified framework. This consolidation unlocks new possibilities for sharing data across tasks, leveraging self-supervised signals more effectively, and ultimately serving as a foundational model for a range of clinical and research applications in medical imaging.

\section{Related Work}
\label{sec:medformer_related_work}

Much of the progress in deep learning for medical imaging has centered around strategies tailored to specific modalities or tasks, such as MRI-based segmentation, X-ray classification, or CT-based anomaly detection. While these domain-specific models have yielded strong results, they often struggle to generalize to different imaging contexts or tasks. Work in self-supervised learning (SSL) and transformer architectures has offered potential pathways toward more universal frameworks, yet these approaches have typically been applied in separate or simplified contexts rather than in a combined, multitask, multimodal setting.

\paragraph{Self-Supervised Learning in Medical Imaging.}
Self-supervised learning methods in medical imaging typically define a proxy objective that draws on intrinsic properties of the data without requiring explicit labels. Examples include puzzle-based tasks, image inpainting, or instance discrimination through contrastive schemes. Such objectives have been explored in modalities like MRI, CT, and histopathology images. In many instances, self-supervision enables the model to acquire generalizable feature representations, reducing the reliance on large annotated datasets. For instance, researchers have examined contrastive approaches that embed volumetric scans into a latent space reflective of tissue structure, while other studies have applied rotation or permutation tasks to exploit the geometric nature of anatomical images. However, most of these studies remain focused on a single modality or a single body part, limiting their ability to reuse learned features across substantially different types of data such as 2D chest X-rays and 3D brain MR scans. The extension of self-supervision to encompass multiple modalities and tasks in a single model is thus a promising direction.

\paragraph{Transformers in Medical Imaging.}
Transformers have gained attention in the broader vision community through architectures like the \acro{ViT}[Vision Transformer], which applies global self-attention to flattened image patches, and related variants that integrate convolutional features with attention layers. Within medical imaging, several works have shown that attention-based architectures can capture long-range dependencies that are critical for identifying lesions or anatomical landmarks across large scans. Such transformer-based models are frequently adapted to tasks like organ segmentation or lesion classification but are often trained with standard supervised learning. This supervised training tends to restrict their flexibility, as transformers must be reconfigured for each new modality or task. While certain methods have begun to investigate the power of transformer encoders for joint segmentation and classification, or to unify 2D and 3D data within similar pipelines, truly multimodal transformer approaches that handle a broad mix of medical domains remain limited.

\paragraph{Toward a Unified Framework.}
The concept of a model simultaneously managing different anatomies, modalities, and label structures, while also leveraging SSL signals, has garnered growing interest. Existing attempts frequently rely on manually harmonized backbones and domain-specific headers, or else they apply an ad hoc form of multi-task learning without fully unifying representations. Consequently, there is a recognized need for an adaptable, domain-agnostic strategy that can integrate 2D and 3D scans, handle disjoint sets of labels, and tap into self-supervised pretext tasks. Medformer aims to address this gap by proposing a transformer backbone enriched with latent embeddings tailored to dimensionality, modality, and body region, all within a flexible architecture that can handle diverse tasks under either supervised or self-supervised protocols.

\section{Architecture Design and Conceptual Framework}
\label{sec:medformer_arch_design}

The Medformer architecture aims to unify multiple data modalities and tasks within a single transformer-based backbone. It is built around the idea that although imaging data vary substantially in dimensionality, anatomical coverage, and acquisition modality, there exists a set of more abstract, domain-agnostic feature representations that can be learned and reused across many clinical contexts. The conceptual design is founded on separating the network into three key components: an \emph{Input Adaptformer}, a \emph{Main Body}, and an \emph{Output Adaptformer}. These modules work in tandem to produce a standardized feature representation from a raw input (whether two-dimensional or three-dimensional), and then map that representation to the appropriate output space for each distinct task.

The \emph{Input Adaptformer} addresses the heterogeneity in shape, modality, and body region among medical images. It performs patchification according to whether the data is two or three-dimensional, applies position embeddings, and then incorporates a set of trainable latent embeddings that encode relevant priors. For instance, one latent might reflect whether the input arises from a three-dimensional CT volume, another might indicate the imaging modality (like X-ray or microscopy), and a third might specify the anatomical region (e.g., chest, abdominal). Because these latent embeddings carry domain-related information about dimensionality, acquisition style, and organ systems, the network can be taught how to direct its attention differently depending on these factors. In other words, the Input Adaptformer serves as an adaptive entry point that shapes the data into a standardized, “canonical” representation required by the core transformer layers.

Once images pass through the Input Adaptformer, they enter the \emph{Main Body}, a transformer-based module intended to be general-purpose and task-agnostic. It operates on the standardized tokens produced by the Input Adaptformer, engaging attention blocks that capture contextual relations within the tokenized image representation. In principle, this stage is where the model learns image features that are not confined to a single body region or modality but rather embody more universal cues, such as structural edges, textural signatures, and other properties that could be relevant to different clinical tasks.

Finally, the \emph{Output Adaptformer} receives the hidden representations from the Main Body and conditions them on a second set of latent embeddings dedicated to the task domain. Specifically, it includes “task-specific” latents that guide how the final representations should be interpreted for classification, segmentation, or other supervised objectives. By cross-attending the Main Body’s token embeddings with these task-specific latent vectors, the Output Adaptformer can produce the set of logits or predictive outputs required for each dataset or label structure. Consequently, a single unified representation can be redirected to serve any of a range of tasks or modalities, simply by changing which task latents are included.

\begin{figure}
    \centering
    \begin{tikzpicture}[
  node distance=1.5cm and 2cm,
  every node/.style={font=\sffamily},
  box/.style={rectangle, draw, rounded corners, minimum height=1cm, minimum width=2.5cm, align=center, fill=blue!10},
  latent/.style={rectangle, draw, dashed, rounded corners, minimum height=0.8cm, minimum width=2cm, align=center, font=\small, fill=green!10},
  transformer/.style={rectangle, draw, rounded corners, minimum height=1cm, minimum width=3cm, align=center, fill=orange!20},
  arrow/.style={-{Stealth[scale=1.0]}, very thick},
  embedding/.style={draw, fill=gray!20, minimum size=0.4cm, inner sep=0pt},
  title/.style={font=\bfseries\Large}
]

\node[box] (raw) {Raw Input (2D/3D)};
\node[box, below=of raw] (patch) {Patchification \&\\Positional Encoding};

\matrix (emb1) [matrix of nodes,
  nodes=embedding,
  column sep=2mm, row sep=2mm,
  anchor=west
] at ($(patch.east)+(0.5,0)$) {
  &  &  \\
  &  &  \\
  &  &  \\
};

\node[transformer, below=of patch, yshift=-0.5cm] (inputAdapt) {Input Adaptformer\\(Transformer Blocks)};

\node[latent, left=of inputAdapt, xshift=-1.5cm] (dimLatent) {Dimensionality\\Latents (2D/3D)};
\node[latent, above=of dimLatent, yshift=0cm] (modalityLatent) {Modality\\Latents};
\node[latent, below=of dimLatent, yshift=0cm] (bodyPartLatent) {Body Part\\Latents};

\node[box, below=of inputAdapt, yshift=-0.5cm] (canonical) {Canonical Feature\\Representation};
\matrix (emb2) [matrix of nodes,
  nodes=embedding,
  column sep=2mm, row sep=2mm,
  anchor=west
] at ($(canonical.east)+(0.5,0)$) {
  &  &  &  \\
  &  &  &  \\
};

\node[transformer, below=of canonical, yshift=-0.5cm] (mainBody) {Main Body\\(Task-agnostic Transformer)};

\node[transformer, below=of mainBody, yshift=-0.5cm] (outputAdapt) {Output Adaptformer\\(Transformer + Cross-Attention)};

\node[latent, right=of outputAdapt, xshift=1.0cm] (taskLatent) {Task-specific\\Latent (19 tasks)};

\node[box, below=of outputAdapt, yshift=-0.5cm] (pred) {Final Prediction\\(Classification/Segmentation)};

\draw[arrow] (raw) -- (patch);
\draw[arrow] (patch) -- (inputAdapt);
\draw[arrow] (inputAdapt) -- (canonical);
\draw[arrow] (canonical) -- (mainBody);
\draw[arrow] (mainBody) -- (outputAdapt);
\draw[arrow] (outputAdapt) -- (pred);

\draw[arrow, dashed] (dimLatent.east) -- ++(0.5,0) |- (inputAdapt.west);
\draw[arrow, dashed] (modalityLatent.east) -- ++(0.5,0) |- (inputAdapt.west);
\draw[arrow, dashed] (bodyPartLatent.east) -- ++(0.5,0) |- (inputAdapt.west);

\draw[arrow, dashed] (taskLatent.west) -- ++(-0.5,0) |- (outputAdapt.east);

\node[right=0.2cm of raw] {\small Input (Image/Volume)};
\node[right=0.2cm of canonical] {\small Canonical, fixed-dim features};
\node[right=0.2cm of mainBody] {\small General feature processing};

\end{tikzpicture}
    \caption{Medformer Architecture}
    \label{fig:medformer_arch}
\end{figure}
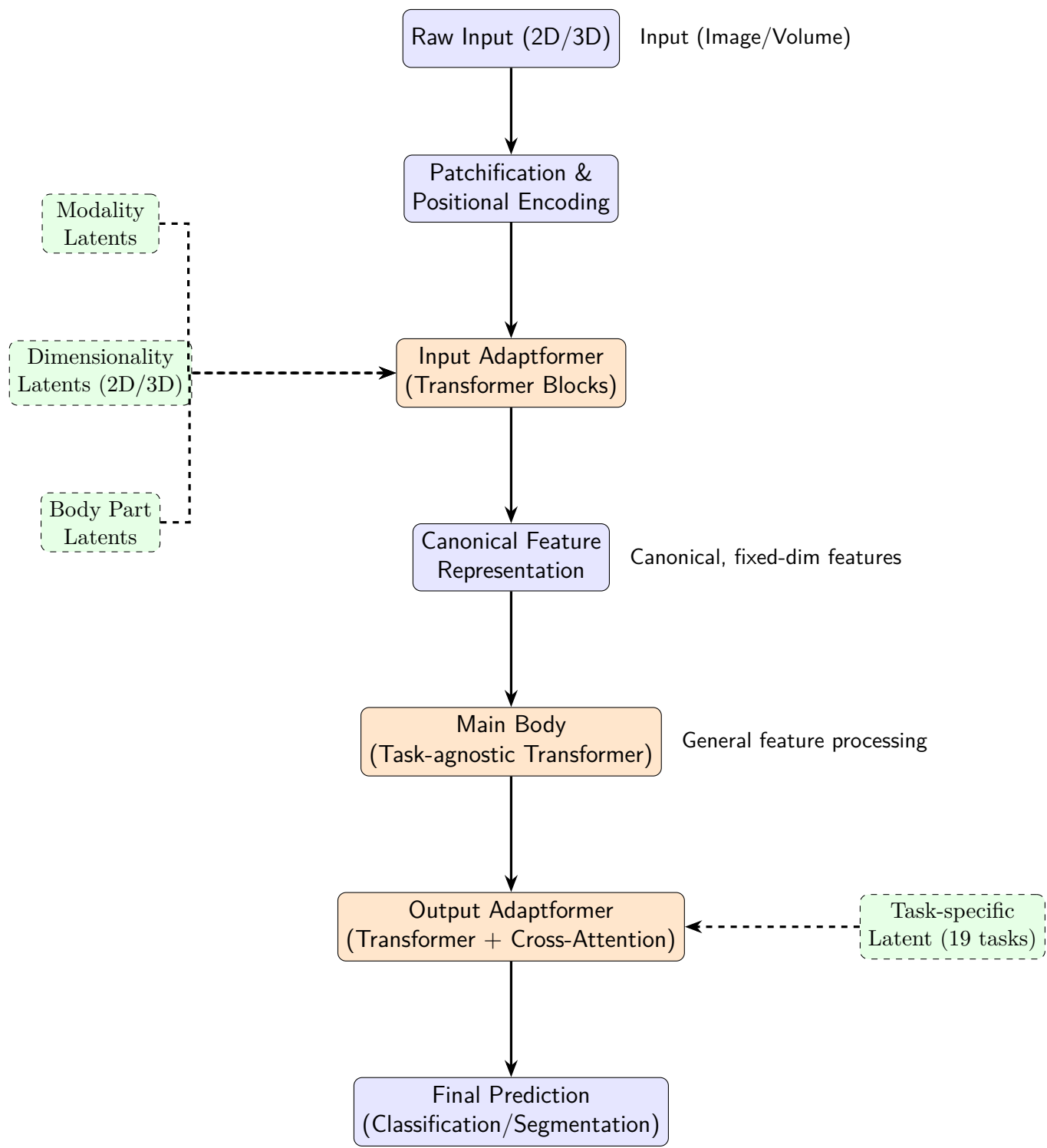

An important element of this design is its capacity to handle both two- and three-dimensional data within the same pipeline. The Input Adaptformer includes distinct paths for 2D and 3D inputs, each of which uses an appropriate patchification strategy (and associated position embeddings) before merging the result into the common dimensional embedding space. Meanwhile, the presence of multiple latent embeddings (for dimensionality, modality, and body part) ensures that the architecture can specialize aspects of its attention mechanism to each combination of properties without duplicating the Main Body or retraining the entire system. This structuring also encourages parameter sharing across different modalities and tasks, which is especially beneficial when certain datasets are small or label-scarce. See the overall Medformer architecture \ref{fig:medformer_arch}.

Compared to earlier work that uses a single model for a single domain or to purely transformer-based solutions that do not incorporate multi-latent cross-attention, Medformer’s approach allows for the disentangling of domain-specific knowledge from the central feature extraction pipeline. This separation helps make the core model more robust and reusable, since it can be combined with new sets of latent embeddings if additional imaging modalities or tasks become necessary. Moreover, as a potential platform for self-supervised learning, it can accommodate proxy objectives that train the Input and Output Adaptformer modules to align unlabeled images with relevant latent priors (such as “3D MRI of the abdominal region”), even when supervised labels are unavailable. The next sections explore how these components come together for both supervised and self-supervised tasks and discuss preliminary evidence for Medformer’s adaptability to multiple medical imaging datasets.

\section{Adaptformers: Dynamic Input-Output Adaptation}
\label{sec:adaptformer_section}

A central mechanism in Medformer is the use of \emph{Adaptformers}, which effectively separate domain- or task-specific transformations from the universal feature learning procedure. In particular, the Medformer architecture includes two such modules:

\begin{enumerate}
    \item \textbf{Input Adaptformer:} Processes raw data and guides it into a canonical format suitable for the backbone. It leverages specialized latent embeddings to incorporate knowledge of whether the input is 2D or 3D, which modality is in use (e.g., CT or microscope), and which body part is imaged (e.g., chest or abdominal).
    \item \textbf{Output Adaptformer:} Operates on the high-level representations produced by the backbone. It combines them with \emph{task-specific} latent embeddings, culminating in final predictions for a variety of medical tasks (classification, or ordinal regression).
\end{enumerate}

\begin{figure}
    \centering
    \includegraphics[width=1.0\textwidth]{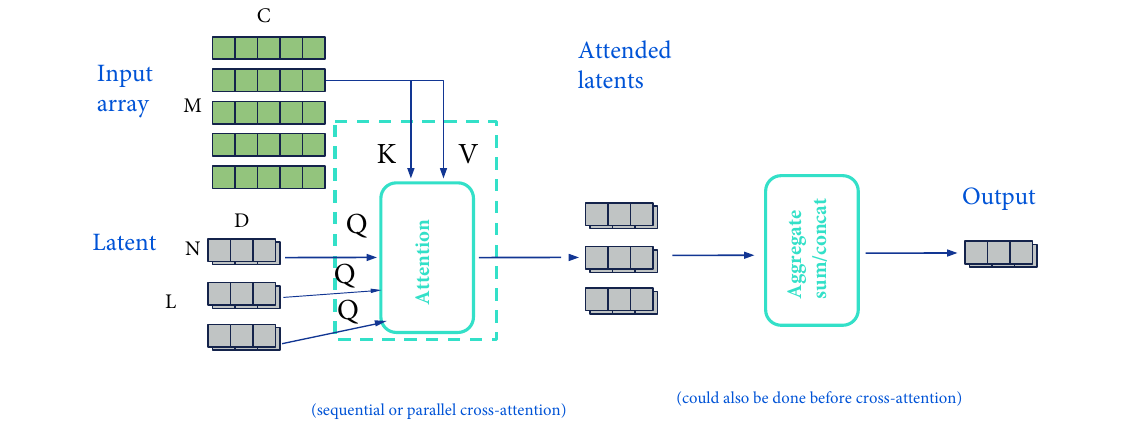}
    \caption{Adaptformer Architecture}
    \label{fig:adaptformer_arch}
\end{figure}

By decoupling these domain and task considerations into dedicated Adaptformer modules, Medformer can share the bulk of its parameters across diverse datasets and tasks. This design also makes it straightforward to introduce new modalities or tasks by defining additional latent embeddings or additional heads in the Output Adaptformer, rather than modifying or retraining the entire core network.See figure \ref{fig:adaptformer_arch} for an overview of the Adaptformer architecture. 

\subsection{Role of Latent Embeddings}

In the context of Medformer, a \emph{latent embedding} is a trainable vector (or collection of vectors) that encodes prior knowledge about the data or task domain. Specifically, three categories of latent embeddings shape how the network interprets an input at the \emph{input stage}:

\begin{itemize}
    \item \textbf{Dimensionality Latent:} Distinguishes between 2D inputs (e.g., X-ray images) and 3D inputs (e.g., volumetric CT or MRI). This latent can guide how patchification and positional encoding are carried out, ensuring the model processes 2D slices or 3D subvolumes appropriately.
    \item \textbf{Modality Latent:} Accounts for imaging characteristics such as X-ray, CT, MRI, microscopic pathology slides, or other modalities. Different modalities can have different intensity ranges, noise characteristics, and relevant structures, so the modality latent helps the Input Adaptformer highlight features known to be salient for that modality.
    \item \textbf{Body Part Latent:} Encodes information about the anatomical region (chest, abdominal, skin, etc.). Variations in shape, organ layout, and pathology presentation across body regions can be reflected in these latent embeddings.
\end{itemize}

Meanwhile, at the \emph{output stage}, Medformer’s Output Adaptformer introduces \textbf{task-specific} latents. Each dataset or clinical objective (e.g., single-label classification, binary classification, ordinal regression) is paired with its own latent embedding that helps interpret the higher-level features in a manner consistent with that task’s label structure.

\paragraph{Configuration and Taxonomy of Latents.}
One might organize these latent embeddings systematically, for instance:

\begin{itemize}
    \item \texttt{dim\_latents}: \{\emph{2D\_latent}, \emph{3D\_latent}\}.
    \item \texttt{modality\_latents}: \{\emph{chest\_xray}, \emph{ct\_scan}, \emph{microscopy}, \emph{oct}, \dots\}.
    \item \texttt{body\_part\_latents}: \{\emph{chest}, \emph{abdominal}, \emph{brain}, \dots\}.
    \item \texttt{task\_latents}: \{\emph{chestmnist\_binary}, \emph{pathmnist\_multiclass}, \emph{pneumoniamnist\_binary}, \dots\}.
\end{itemize}

Each of these latent embeddings is a small tensor of trainable parameters. When the Input Adaptformer processes an image, it selects (or concatenates) the relevant embeddings (e.g., \emph{2D\_latent} + \emph{chest\_xray} + \emph{chest} for a 2D chest X-ray task). When the Output Adaptformer produces predictions, it cross-attends the resulting feature tokens with the relevant \emph{task\_latent} for that classification or segmentation goal.

\subsection{Input Adaptformer Mechanics}

Conceptually, the Input Adaptformer merges the patchified input features with the domain-specific latents via cross-attention. Algorithm~\ref{alg:input_adaptformer} provides high-level pseudocode:

\begin{algorithm}[ht]
    \caption{High-Level Pseudocode for Input Adaptformer}
    \label{alg:input_adaptformer}
    \begin{algorithmic}[1]
        \Require $x$ \Comment{Raw image or volume (2D or 3D)}
        \Require $\ell_{\text{dim}}$ \Comment{Dimensionality latent (2D or 3D)}
        \Require $\ell_{\text{mod}}$ \Comment{Modality latent (e.g., CT, X-ray)}
        \Require $\ell_{\text{body}}$ \Comment{Body-part latent (e.g., chest, abdomen, brain)}
        \State $x_{\text{patch}} \gets \text{Patchify}(x)$ \Comment{Divide input into flattened patches}
        \State $x_{\text{pos}} \gets x_{\text{patch}} + \text{PosEmbed}(x_{\text{patch}})$ \Comment{Add positional embeddings (2D or 3D)}
        \State $L \gets \text{Concat}(\ell_{\text{dim}},\, \ell_{\text{mod}},\, \ell_{\text{body}})$ \Comment{Concatenate latent codes; shape: $[B, L, D]$ for $B$ images}
        \For{each Transformer block $b \in \{\text{TransformerBlock}_1,\dots,\text{TransformerBlock}_n\}$}
            \State $x_{\text{pos}} \gets b(x_{\text{pos}}, L)$ \Comment{Apply cross-attention and self-attention mechanisms}
        \EndFor
        \State \Return $x_{\text{pos}}$ \Comment{Output canonical feature representation for subsequent processing}
    \end{algorithmic}
\end{algorithm}

Here, \texttt{Patchify} will differ slightly for 2D vs.\ 3D inputs, and position $embeddings 
(\texttt{PosEmbed2Dor3D})$ also adapts accordingly. The combined latent $L$ is used in cross-attention blocks so that the network can condition its patch embeddings on prior knowledge about dimensionality, modality, and anatomy.

\subsection{Output Adaptformer Mechanics}

After the core backbone transforms the canonical feature tokens, Medformer employs an Output Adaptformer that integrates \emph{task-specific} latents to produce predictions. These latents represent the type of classification (single-label, binary, etc.), along with subtle distinctions for each dataset’s distribution. Pseudocode appears in Algorithm~\ref{alg:output_adaptformer}:

\begin{algorithm}[ht]
\caption{Pseudocode for Output Adaptformer (High-Level)}
\label{alg:output_adaptformer}
\begin{algorithmic}[1]
\Require $h$: Feature tokens from Main Body
\Require $\ell_{\text{task}}$: Task latent (e.g., chestmnist\_binary, retinamnist\_ordinal)
\State $h_{\text{pooled}} \gets h$ \Comment{Initialize from Main Body output}
\For{$\mathrm{block} \in \{\mathrm{TransformerBlock}_1,\dots,\mathrm{TransformerBlock}_m\}$}
    \State $h_{\text{pooled}} \gets \mathrm{block}(\,h_{\text{pooled}},\, \ell_{\text{task}}\,)$ 
\EndFor
\State $\mathrm{logits} \gets \mathrm{Head}_{\text{dataset}}( \, \mathrm{MeanPooling}( h_{\text{pooled}} ) \, )$
\State \Return $\mathrm{logits}$
\end{algorithmic}
\end{algorithm}

Notably, each dataset can have its own output head (e.g., a single logit for binary classification or multiple logits for multi-class tasks). The Output Adaptformer merges the “universal” features with a vector or set of vectors that capture the labeling scheme and domain constraints for the task at hand.

\subsection{Single-Adaptformer vs.\ Multi-Adaptformer Training}

A Medformer user may choose to:
\begin{itemize}
    \item \textbf{Share the same Input/Output Adaptformer across all domains}, differentiating them solely by which latent embeddings are activated. This \emph{Single-Adaptformer} approach maximizes parameter sharing but forces more learning responsibility onto the latents. We experimented with a singular Output Adaptformer, keeping a singular output head, and storing the task specific information only in the latents, but this approach didn't work very well. It is possible that such an approach is indeed possible but would require much larger latents, increased number of output layers and significant compute investment.
    \item \textbf{Define distinct Input or Output Adaptformers for each domain/task}, a \emph{Multi-Adaptformer} approach that yields more specialized transformations for each dataset. This can improve performance if the domains are highly dissimilar, but at the cost of additional parameters and reduced parameter sharing.
\end{itemize}

In practice, many experiments use the single-Adaptformer setting when domains are related (e.g., multiple 2D tasks with similar pixel-level statistics) but rely on separate latents to modulate how cross-attention is applied to each domain. Alternatively, if the tasks vary widely (for instance, some are 2D microscopy while others are 3D volumetric scans of a different organ), the multi-Adaptformer approach might yield a better balance of specialization and performance.

\subsection{Illustrative Example of Latent Configuration}

To give a sense of how these embeddings could be declared, we might employ a simple configuration structure:

\begin{verbatim}
latents_config:
  dimension_latents:
    - name: 2d_latent
    - name: 3d_latent
  modality_latents:
    - name: chest_xray
    - name: ct_scan
    - name: microscopic
  body_part_latents:
    - name: chest
    - name: abdominal
    - name: brain
  task_latents:
    - name: chestmnist_binary
    - name: retinamnist_ordinal
    - name: pathmnist_singlelabel
...
\end{verbatim}

When a chest X-ray image is input, the system can automatically assemble $\ell_{dim} = \texttt{2d\_latent}$, $\ell_{mod} = \texttt{chest\_xray}$, and $\ell_{body} = \texttt{chest}$. Similarly, if the model is set to predict multi-label findings, it might apply $\ell_{\text{task}} = \texttt{chestmnist\_multilabel}$ in the Output Adaptformer. This approach encourages composability: new body parts, modalities, or tasks can be added by specifying new latent embeddings without retraining from scratch.

\subsection{Summary of Adaptformer Advantages}

In summary, Adaptformers let the network adapt how it reads inputs and how it produces outputs, all while keeping a common Main Body for universal feature extraction. This has numerous benefits:

\begin{itemize}
    \item \textbf{Modular design:} Additional datasets can be integrated by defining new latents or new output heads, leaving the backbone largely untouched.
    \item \textbf{Reduced duplication:} Potentially hundreds of shared backbone layers can be reused across tasks, with only modest overhead introduced by Adaptformers and their latent embeddings.
    \item \textbf{Improved task synergy:} Even if tasks differ (e.g., classification vs.\ regression), they can still share certain parts of the representation, while the Adaptformers handle domain/task-specific idiosyncrasies.
\end{itemize}

Adaptformers thus form the core of Medformer’s strategy for tackling multiple medical imaging tasks and modalities within a single model, simultaneously providing a path for flexible domain adaptation and specialized final predictions.

\section{Self-Supervised Training}
\label{sec:self_supervised_training}

The Medformer architecture lends itself naturally to \emph{self-supervised learning} in medical imaging, primarily because of the flexibility provided by its modular latents and shared backbone. The key principle is that unlabeled data from a variety of imaging modalities can be fed through the same input pathway, adapted by the relevant dimensionality and modality latents, and finally processed by the main transformer layers. Concurrently, the output portion of the network can be configured to solve a \emph{pretext} objective—one that does not require explicit human annotation but still encourages the model to distill clinically relevant representations.

Self-supervised training can take numerous forms. One of the most straightforward approaches is to learn invariances under augmentations, such that pairs of images transformed from the same underlying volume or slice should map to similar latent representations. More advanced contrastive and non-contrastive methods (such as VICReg, Barlow Twins, or methods that exploit momentum encoders) can be incorporated by allocating specific \emph{task latents} and heads in the Output Adaptformer for these self-supervised losses. For instance, a contrastive embedding head might learn to maximize agreement between two augmented versions of the same scan while distinguishing different scans.

\subsection{Conceptual Overview}

When performing self-supervised training, the Input Adaptformer uses the same dimensional and modality latents that a fully supervised workflow would use. This ensures that images of the same modality (e.g., 2D chest X-ray) or the same dimensional class (e.g., 3D abdominal CT volumes) receive consistent treatment, including patchification and positional embeddings. The model’s parameter sharing across tasks means it can accumulate domain knowledge from any available unlabeled data, learning to encode anatomical or pathological features that might generalize across tasks. Hence, a self-supervised signal from large volumes of unlabeled 3D CT scans may help the network better classify disease in smaller, fully labeled 2D X-ray datasets.

In practice, separate \emph{task latents} can be introduced for distinct self-supervised objectives. One latent might be designated for a contrastive objective, another for rotation prediction, and yet another for Jigsaw puzzle rearrangements. Because each of these tasks leads to different target outputs, they can be handled by separate heads in the Output Adaptformer, and their corresponding latents can capture specialized aspects of the data. This design makes it easier to combine multiple self-supervised strategies in a \emph{single} training run: each unlabeled image can be augmented in several ways, forwarded through the Input Adaptformer, processed by the universal backbone, and then branched into multiple self-supervised heads.

\subsection{Multiple Objectives and Latent Combinations}

One important strength of Medformer’s approach to self-supervision is the option to \emph{simultaneously} train with multiple objectives. Suppose we have:

\begin{itemize}
    \item A \textbf{contrastive} task latent to enforce that two augmentations of the same sample produce similar embeddings, while different samples remain distinct.
    \item A \textbf{reconstruction} task latent that tries to reconstruct missing slices or patches from the partial input.
    \item An \textbf{image rotation} classification latent that predicts the orientation or flip transformations applied to the input.
\end{itemize}

All three tasks could be carried out in parallel. Each mini-batch might contain multiple views of each raw image, each paired with the relevant self-supervised head. The main body of the transformer thus adapts to satisfy all three constraints, reinforcing robust, modality-aware features.

This approach allows for a highly compositional training scheme: new unlabeled data, even from modalities not heavily represented in supervised tasks, can reinforce general feature extraction. Meanwhile, if certain data distributions are unlabeled but abundant (e.g., large repositories of MRI scans), the model can still improve its ability to process volumetric images, potentially benefiting downstream tasks such as 2D classification or 3D anomaly detection when labels become available.

\subsection{Benefits in Medical Imaging Contexts}

In medical imaging, collecting labeled datasets can be prohibitively expensive and time-intensive. By implementing self-supervised objectives:

\begin{itemize}
    \item The model \emph{leverages} unlabeled datasets from various hospitals and imaging protocols, filling in domain gaps and reducing the necessity for large annotated sets.
    \item \emph{Generalizable features} emerge that handle different scanning parameters, anatomies, or contrast levels with minimal supervision.
    \item \emph{Downstream tasks} benefit from stronger initialization, often requiring fewer epochs and fewer annotations to achieve competitive performance.
\end{itemize}

Moreover, the unified backbone ensures that any progress made in representing, for instance, 3D volumes, can also transfer to 2D tasks if those tasks share certain latent embeddings or rely on overlapping morphological cues.

\subsection{Training Procedure}

Although many algorithms for self-supervised training exist, the Medformer workflow typically involves a few core steps:

\begin{enumerate}
    \item \textbf{Data Augmentation:} Each image is perturbed by random transformations (e.g., flips, rotations, random cropping, or intensity jitter) that preserve diagnostic structure but force the model to learn invariances.
    \item \textbf{Input Adaptation:} The Input Adaptformer prepares the image tokens, guided by dimension, modality, and body-part latents.
    \item \textbf{Backbone Encoding:} The universal transformer layers build a shared representation that can be reused across tasks.
    \item \textbf{Self-Supervised Heads:} Multiple heads in the Output Adaptformer process the encoded tokens. Depending on the chosen self-supervised approach, each head produces an embedding or reconstruction that is compared against a self-supervised target.
    \item \textbf{Loss Minimization:} Gradients from each self-supervised loss flow back through the shared backbone and the relevant latents, tuning them to encode useful features for all objectives simultaneously.
\end{enumerate}

This process can be repeated for multiple modalities, such that a single pass of training might sample some 2D X-ray data, some 3D MRI volumes, and perhaps large sets of microscopic pathology images. Each batch employs the appropriate latents, with the relevant pretext tasks, creating broad coverage over a variety of medical image types.

\subsection{Outlook on Large-Scale Self-Supervised Pretraining}

The current Medformer implementation can scale up as computational resources allow, extending from compact images (such as 28$\times$28 slices) to larger field-of-view images and volumes. As institutions increasingly collect vast amounts of unlabeled scans, large-scale self-supervised pretraining becomes more appealing. A medical institution could store enormous quantities of volumetric data: although only a fraction is annotated, those unlabeled volumes still provide anatomical and modality cues. Medformer’s design makes it feasible to incorporate them via cross-attention with the correct dimension and modality embeddings, thus bridging domain gaps and improving the universal representation.

Ultimately, self-supervision in Medformer paves the way for a flexible, data-hungry model that thrives on the varied forms of medical imaging data. By integrating these ideas into a single architecture, we can expect more robust, domain-invariant features, enhanced generalization to rare diseases or unusual scanning protocols, and faster adaptation to new downstream tasks when clinically labeled datasets become available.

\section{MedMNIST Experiment Setup}
\label{sec:medmnist_experiments}

We focused on evaluating Medformer using the \textit{MedMNIST} dataset collection~\cite{medmnist}, which comprises multiple medical image datasets in both two-dimensional (2D) and three-dimensional (3D) formats. Each dataset addresses a distinct clinical or anatomical domain, such as breast ultrasound, chest X-ray, abdominal CT, and various microscopy images, among others. The MedMNIST collection simplifies comparisons by unifying these datasets into a consistent size format ($28\times28$ for 2D, $28\times28\times28$ for 3D), allowing us to more directly study the effects of dimensionality, modality, and body part embeddings within the Medformer pipeline. In addition, the MedMNIST repository offers labeled examples in single-label classification, multi-label classification, binary classification, or ordinal regression, covering a wide spectrum of label structures. This breadth of tasks makes MedMNIST an ideal test bed for validating how effectively Medformer generalizes across heterogeneous medical imaging domains.

\subsection{Data Overview}

\begin{figure}
    \centering
    \includegraphics[width=1.0\textwidth]{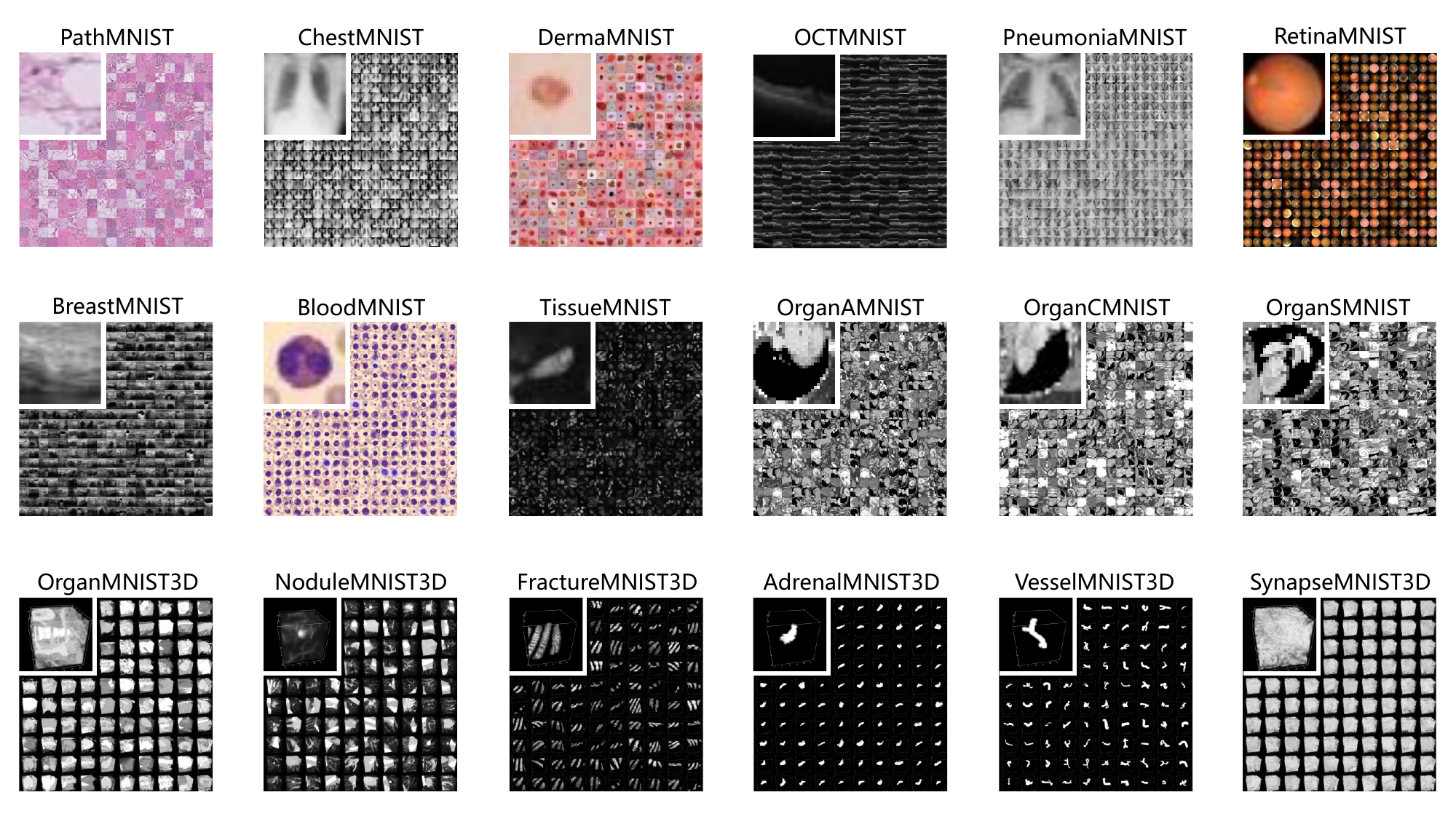}
    \caption{MedMNIST Dataset}
    \label{fig:medmnist}
\end{figure}

MedMNIST version 2 includes 18 datasets, each curated from different sources and imaging modalities. Rather than maintaining inconsistent shapes or resolutions, the dataset authors resampled all images to uniform $28\times28$ pixels (or $28\times28\times28$ voxels for 3D), see figure \ref{fig:medmnist}. In addition, different datasets target different clinical objectives, ranging from diagnosing pneumonia in chest X-rays to classifying tissue subtypes in histological slides. In the second version of the dataset, the authors also made available higher resolution images for all the datasets, resulting in $224\times224$ 2D images and $64\times64\times64$ for 3D ones. Table~\ref{tab:medmnist_distribution} summarizes the main characteristics of the datasets we focus on in our experiments.

\begin{table}[ht]
  \centering
  \caption{MedMNIST dataset distribution. The table lists a subset of representative tasks, modalities, dimensionalities, and label types.}
  \label{tab:medmnist_distribution}
  \begin{tabular}{l c c c c}
    \toprule
    \textbf{Dataset} & \textbf{Modality} & \textbf{Dim} & \textbf{Task Type} & \textbf{\#Samples}\\
    \midrule
    PathMNIST         & Histopathology (Microscopic) & 2D & Single-label  & 107,180 \\
    ChestMNIST        & X-Ray                        & 2D & Binary        & 112,120 \\
    DermaMNIST        & Dermoscopy (Microscopic)     & 2D & Single-label  & 10,015 \\
    OCTMNIST          & Retinal OCT                  & 2D & Single-label  & 109,309 \\
    PneumoniaMNIST    & X-Ray                        & 2D & Binary        & 5,856 \\
    RetinaMNIST       & Fundus Camera                & 2D & Ordinal       & 1,600 \\
    BreastMNIST       & Ultrasound                   & 2D & Binary        & 780 \\
    BloodMNIST        & Blood Cell Microscopy        & 2D & Single-label  & 17,092 \\
    TissueMNIST       & Tissue Microscopy            & 2D & Single-label  & 236,386 \\
    OrganAMNIST       & Abdominal CT (2D)            & 2D & Single-label  & 58,830 \\
    OrganCMNIST       & Abdominal CT (2D)            & 2D & Single-label  & 23,583 \\
    OrganSMNIST       & Abdominal CT (2D)            & 2D & Single-label  & 25,211 \\
    OrganMNIST3D      & Abdominal CT (3D)            & 3D & Single-label  & 1,742 \\
    NoduleMNIST3D     & Chest CT (3D)                & 3D & Binary        & 1,633 \\
    AdrenalMNIST3D    & Abdominal CT (3D)            & 3D & Binary        & 1,584 \\
    FractureMNIST3D   & Bone CT (3D)                 & 3D & Single-label  & 1,370 \\
    VesselMNIST3D     & Brain MRA (3D)               & 3D & Binary        & 1,908 \\
    SynapseMNIST3D    & Electron Microscopy (3D)     & 3D & Binary        & 1,759 \\
    \bottomrule
  \end{tabular}
\end{table}

Not all datasets in MedMNIST are of equivalent size, class balance, or difficulty. Some (like PathMNIST) contain tens of thousands of images, whereas others (like BreastMNIST) consist of fewer than a thousand examples, potentially limiting the capacity of purely supervised approaches. Further variations arise from the diversity of labeling schemes: \textit{single-label classification} ranges from 2 to more than 10 classes, and \textit{ordinal} tasks require a different loss function to incorporate rank order. Thus, incorporating all these tasks under one unifying architecture stresses both the domain-adaptation layers (i.e., dimension, modality, and body-part latents) and the task-specific latents.

\subsection{Rationale and Experimental Goals}

Using MedMNIST helps us address core questions about the generalization and flexibility of Medformer:
\begin{itemize}
    \item \textbf{Multimodal Capability:} Since MedMNIST covers both 2D and 3D inputs, we can examine how effectively the model adapts between planar images (e.g., X-rays, microscopic slides) and volumetric data (e.g., CT or MRI scans).
    \item \textbf{Task Heterogeneity:} We can quantify how well a single backbone manages binary tasks, single-label multi-class classification, and ordinal classification.
    \item \textbf{Impact of Latent Embeddings:} We can observe how dimensional, modality, and body-part embeddings shape internal representations, potentially revealing patterns that transfer across clinically diverse datasets.
    \item \textbf{Scalability in Data-Limited Settings:} Smaller datasets within MedMNIST allow us to test the model’s performance when training data are scarce, which simulates real-world scenarios of medically rare conditions.
\end{itemize}

Overall, the MedMNIST suite exposes the model to a wide variety of medical imaging characteristics in a relatively controlled experimental setting. This design aligns with our desire to build and evaluate a foundation that can be extended to larger-scale or higher-resolution clinical data without drastically altering the underlying architecture.

\subsection{Implementation Details}

In all experiments, the core methodology remains consistent. Each dataset is loaded, normalized to the $[0,1]$ range, and subjected to random augmentations such as flipping or random cropping, following the settings offered by MedMNIST using a RandAugment \cite{cubuk2020randaugment} implementation. Two-dimensional samples are then processed by the Medformer pipeline in 2D mode, while three-dimensional volumes use the 3D path. We group tasks by their dataset indices for domain-specific embeddings, ensuring that the correct dimensional, modality, and body-part latents are selected.

Training usually proceeds for a fixed number of epochs (e.g., between 100 and 300, depending on the dataset size and batch size) with standard optimizers like AdamW and learning-rate scheduling. We log both training and validation metrics, monitoring the mean classification performance for each dataset under single-task or multi-task scenarios. To keep experiments comparable across data subsets, batch sizes and initial learning rates are chosen to fit memory constraints while providing stable updates, and we maintain similar early-stopping or checkpoint policies for all tasks. 

In subsequent sections of this chapter, we illustrate how Medformer leverages the architecture outlined above to learn from these diversified 2D and 3D medical images, culminating in results that demonstrate its capacity for broad adaptation with minimal per-task overhead.

\section{Methodology}
\label{sec:methodology}

The design of Medformer facilitates a range of experimental strategies that collectively evaluate how effectively a single architecture can learn from different data sources, with and without labeled supervision. In this section, we discuss our methodological framework, including data usage, hyperparameter selections, training configurations, and the variety of runs performed to benchmark both supervised and self-supervised learning. We emphasize how each run highlights a specific aspect of Medformer’s capacity to adapt across tasks, modalities, and dimensionalities.

\subsection{Overview of Configurations and Pipeline}

We divide our methodological design into two main types of runs: \emph{fully supervised} and \emph{self-supervised pretraining} and SSL fine-tuned. For each category of experiment, the same codebase is used; the difference lies in how Medformer’s Input and Output Adaptformers are parameterized and in which latents are activated. We outline these below:

\begin{enumerate}
    \item \textbf{Single-Experiment Supervised Runs:} We train Medformer on a \emph{single task} (e.g., \textit{ChestMNIST binary}) under a purely supervised setting. This approach focuses on how well the model can adapt to a single dataset’s dimensionality, modality, and body part. 
    \item \textbf{Multi-Task Runs:} We simultaneously train Medformer on multiple datasets in a shared pipeline. By mixing a variety of tasks (2D vs.\ 3D, binary vs.\ multi-class, etc.), we measure how effectively the architecture handles multiple tasks at once, using the appropriate dimensional, modality, and body part latents for each sample.
    \item \textbf{Self-Supervised (SSL) Pretraining:} We run a \emph{VICReg-style} approach on multiple datasets to learn a broad representation in the absence of explicit labels. In these runs, the model sees the same pipeline and data augmentations but is trained with invariance-based objectives. No final classification or segmentation is enforced during pretraining.
    \item \textbf{Fine-Tuning from SSL Checkpoints:} After obtaining pretrained weights from the SSL runs, we transfer them to a \emph{single dataset} or \emph{single task} for a standard supervised objective. This comparison clarifies whether the self-supervised representations expedite convergence or improve final accuracy when only modest labeled data are available.
\end{enumerate}

\noindent In all experiments, we consistently rely on the \textit{MedMNIST} collection to supply 2D or 3D volumes. We utilize the same Input Adaptformer for each sample but condition it on the correct dimensional and modality latents. On the output side, the relevant domain’s classification head is selected automatically based on dataset indices. We log each model’s intermediate losses and performance metrics using a standardized tracking setup.

\subsection{Model Configurations: Small vs.\ Large}
\label{sec:model_configurations}

In addition to the general methodological framework described above, we explored two distinct Medformer configurations, referred to as \emph{small} and \emph{large}. The motivation for examining these configurations is to study how scaling the model’s hidden dimensions, number of layers, and input resolution influences both performance and computational requirements.

\paragraph{Small Model Configuration.}
The \emph{small} Medformer setting is designed to operate on \textbf{28\(\times\)28} two-dimensional images and \textbf{28\(\times\)28\(\times\)28} three-dimensional volumes. As illustrated in the example configuration below, the hidden dimension (\texttt{hidden\_dim}) is set to 128, and the Main body transformer layers (\texttt{num\_layers}) are kept to 4. Each block features 4 attention heads (\texttt{num\_heads}), and the MLP expansion ratio (\texttt{mlp\_ratio}) is 2. The patch size (\texttt{patch\_size}) is chosen as 4 for 2D inputs, balancing memory usage and effective receptive field. In the \emph{Adaptformer} modules, the latent embeddings (\texttt{latent\_dim}) are set to 64, and the model uses 32 latent tokens (\texttt{num\_latent\_tokens}). 

During training, we typically use batch sizes between 64 and 128, with a learning rate of 0.001 and a weight decay around 0.001. A maximum of 100 epochs is allowed, with 5 warmup epochs. Self-supervised learning (SSL) experiments in the small model use an \emph{expander MLP} of width 1024 and depth 3 (e.g., 1024--1024--1024), which projects the backbone’s latent representations before applying the VICReg or similar SSL objective.

\paragraph{Large Model Configuration.}
The \emph{large} Medformer variant increases both the input resolution and the model capacity. Specifically, we use \textbf{224\(\times\)224} images for 2D tasks and \textbf{64\(\times\)64\(\times\)64} volumes for 3D tasks. The hidden dimension rises to 512, while the number of Main body layers is set to 6, with 8 attention heads per layer. The MLP ratio is raised to 4, enabling a broader feature transformation. Patch sizes increase to 16, matching the higher input resolution while keeping computational overhead within a manageable scope. 

The \emph{Adaptformer} components are also scaled up, employing a 256-dimensional latent space and 512 latent tokens. This larger design typically demands lower initial learning rates (on the order of 0.0001) and slightly different warmup or regularization strategies to stabilize training. The SSL expander MLP also grows to 8192 units per layer (e.g., 8192--8192--8192), reflecting the richer feature representations learned from higher-resolution images and volumes.

Although both configurations follow the same methodological pipeline—covering single-task supervised, multi-task supervised, SSL pretraining, and SSL fine-tuning—the \emph{large} model requires more compute and memory but can capture more fine-grained details. In the subsequent results sections, we compare and contrast the performance of these two configurations across the same set of tasks in order to highlight the trade-offs between computational cost and accuracy. While the small scale Medformer experiments were runnable on consumer grade GPU's with 24GB of memory, the larger ones required us to rent datacenter GPU's (H200) which offered us the required 140GB of memory.

\subsection{Training and Hyperparameter Protocols}

The training configuration for Medformer remains largely consistent across runs to maintain comparability:

\paragraph{Augmentations:}
We apply basic random transformations, using RandAugment with parameters N=2 and M=8. For 2D images, these transformations include random horizontal or vertical flips, as well as small random rotations (e.g., $90^\circ$ increments). For 3D volumes, we allow flips along each dimension and random rotation about two spatial axes.

\paragraph{Learning Rate and Scheduling:}
All setups employ an \emph{AdamW} optimizer with a moderate initial learning rate (e.g., in the range $10^{-4}$ to $10^{-3}$, or lower for more complex tasks) and weight decay (often around $10^{-4}$ to $10^{-5}$). We combine this with a learning rate scheduler, using \emph{OneCycleLR} to gradually reduce the learning rate. A small number of warm-up epochs (e.g., up to 5 or 10) helps stabilize training, especially for tasks where data are limited or strongly imbalanced.

\paragraph{Batch Size:}
For two-dimensional tasks, we use batch sizes ranging from 64 to 256, depending on GPU memory. Three-dimensional tasks have higher memory demands, so batch sizes are often reduced to 32 or 64. While the smaller batch sizes for 3D volumes imply more gradient steps per epoch, the uniform design of the Input Adaptformer accommodates both data types in a similar manner.

\paragraph{Loss Functions:}
For single-label classification, we use cross-entropy. Binary classification employs a logistic loss (binary cross-entropy), while multi-label tasks adopt a multi-sigmoid formulation. Ordinal tasks require an additional step of label normalization but otherwise also rely on a mean-squared error or a custom ordinal mapping. In the SSL setting, we utilize an invariance-based VICReg loss, which forces pairs of augmented embeddings to remain close in latent space with a weighted sum of Cosine and MSE losses, while promoting sufficient variance and low cross-correlation.

\paragraph{Logging and Checkpointing:}
We log intermediate metrics (e.g., training losses, validation accuracies) at regular intervals. We also checkpoint the model whenever a lower loss or higher metric is observed. This approach ensures that the best model state is preserved for subsequent evaluation or fine-tuning.

\subsection{Single-Experiment Training}

These runs serve as a \emph{baseline} for both performance and resource consumption. For instance, if we select \textit{PathMNIST} alone, the Input Adaptformer has access to the 2D patchification path plus the embeddings for “microscopic” modality and “tissue” body part. We update only the relevant domain latents and the single classification head for that dataset. After training for up to 100 or more epochs, we record final accuracy and AUC. We repeat this for various tasks, including 2D binary tasks (e.g., \textit{PneumoniaMNIST}) and 3D single-label tasks (e.g., \textit{OrganMNIST3D}).

These isolated runs also reveal the extent to which the architecture handles tasks where the data are either abundant or quite scarce, such as \textit{BreastMNIST}, which has comparatively fewer images. The results highlight the capacity of the Input and Output Adaptformers to mold themselves to a single domain without confusion from other datasets.

\subsection{Multi-Task Training}

For multi-task experiments, we combine all selected MedMNIST domains, effectively training on hundreds of thousands of images in a single pass. The Input Adaptformer identifies the correct dimensional and modality embeddings on a per-sample basis, while the Output Adaptformer relies on the appropriate task latent (plus the correct classification head) for each training example. A single gradient step thus updates the shared backbone with contributions from multiple tasks, enabling potential knowledge transfer. For instance, some 2D chest images might share representational aspects with 3D CT images of the chest region, even though the network processes them via separate domain latents.

One key implementation detail is ensuring that each mini-batch contains samples from various tasks so that the gradient signals for different tasks are interleaved. In practice, we maintain a combined dataset index and let the data loader shuffle across tasks, thereby mixing 2D and 3D mini-batches. The model logs per-task metrics after each epoch, which helps us ascertain whether certain tasks benefit from multi-task synergy or if tasks with large training sets dominate updates.

\subsection{Self-Supervised Pretraining}

Our SSL experiments run \textit{VICReg} on a subset of tasks. Specifically, we load images in an unlabeled fashion, applying random augmentations (such as flips, random rotations, or random intensity perturbations) to generate two transformed views. These views then pass through the \emph{Input Adaptformer}, the shared \emph{Main Model}, and a small projection MLP (the “expander”). The resulting embeddings are fed into the VICReg loss, which penalizes dissimilarities between these paired views, while also enforcing a certain level of feature variance and low redundancy across embedding dimensions.

Throughout SSL training, no classification head is used, although the Input Adaptformer still chooses appropriate modality and body part embeddings. This arrangement ensures that domain-specific knowledge is still present in the representation. After SSL converges—often requiring fewer epochs than fully supervised training on each dataset individually—we save a checkpoint containing all learned parameters, including the newly trained latents.

\subsection{Fine-Tuning from SSL}

To evaluate the potential benefit of self-supervision, we load the SSL-pretrained checkpoint and initialize Medformer’s parameters, selectively replacing or adjusting them for a given single-target task. For instance, we might load the SSL checkpoint and subsequently fine-tune only the final classification head plus partial layers in the Input or Output Adaptformers on \textit{DermaMNIST}. Alternatively, we could unfreeze all parameters to observe if the learned embeddings help accelerate convergence or improve final accuracy.

We track how many epochs are required to match baseline single-task performance, as well as whether the final results surpass them. In this manner, we can quantify any gains that arise from pretraining with unlabeled data. Observing whether such improvements generalize across multiple tasks reveals whether the SSL embeddings transfer well or remain domain-specific to tasks prevalent in the unlabeled portion.

\subsection{Comparisons and Summary}

By undertaking four main categories of experiments—\textit{single-experiment supervised}, \textit{multi-task supervised}, \textit{pure SSL}, and \textit{SSL fine-tuning}—we aim to shed light on the following questions:

\begin{itemize}
    \item How does Medformer perform on each MedMNIST dataset when fully supervised?
    \item Can multiple domains be combined into a single training run, and does this foster improved representations or lead to performance trade-offs for smaller tasks?
    \item Does self-supervised pretraining on diverse unlabeled data lead to representations that are beneficial when only limited labeled samples exist for a specific dataset?
    \item What is the relative cost (in terms of convergence time, hyperparameter tuning, or memory usage) of operating with shared architecture blocks and domain-specific latents?
\end{itemize}

Our methodology thus implements a thorough set of scenarios, systematically toggling between single-task vs.\ multi-task and supervised vs.\ SSL-based settings. The results from these experiments, presented in the following sections, provide a comprehensive view of Medformer’s adaptability, efficiency, and overall performance across a wide range of medical image tasks.

\section{Results and Analysis}
\label{sec:results_analysis}

This section details our findings under both fully supervised and self-supervised settings, with separate discussions for the \emph{small} (28\(\times\)28 / 28\(\times\)28\(\times\)28) and \emph{large} (224\(\times\)224 / 64\(\times\)64\(\times\)64) Medformer configurations. We first describe the evaluation metrics and overall experimental approach, then present results for each model size.

\subsection{Evaluation Metrics and Experimental Setup}

All experiments track \textit{accuracy}, \textit{AUC}. Binary tasks compute the positive-class probability either via a single-logit or two-logit output, and ordinal tasks measure correctness through a discretized regression approach.

Each run is configured to maintain consistent hyperparameters unless otherwise stated, such as learning rate schedules and batch sizes, to minimize confounding influences. We also log intermediate model checkpoints to identify the epoch in which a performance plateau or drop emerges, ensuring that each approach is fairly compared at its best validation result.

\subsection{Results with the Small Model Configuration}
\label{sec:results_small}

\subsubsection{Single-Experiment Supervised Performance}

For the smaller 2D tasks (e.g., \emph{BreastMNIST} or \emph{PneumoniaMNIST}), our single-task training converged reliably within approximately 50--100 epochs. Preliminary results indicated an AUC in line with published MedMNIST baselines. Although the final accuracies or AUCs fluctuate slightly across different seeds, the observed \emph{mean} metrics consistently match or surpass standard models specifically tailored for each dataset. Notably, tasks with multi-class categories such as \emph{PathMNIST} or \emph{BloodMNIST} converge more slowly, reflecting the broader scope of classification targets. The model’s use of appropriate dimensional and modality latents helped keep the training stable.

\begin{table}[ht]
\centering
\caption{Performance of Medformer on 2D Classification Tasks}
\label{tab:2d_classification}
\begin{tabular}{lccc}
\toprule
\textbf{Dataset} & \textbf{Task Type} & \textbf{AUC} & \textbf{Accuracy (\%)}\\
\midrule
PathMNIST & Multi-class & 0.999 & 97.2 \\
DermaMNIST & Multi-class & 0.772 & 66.8 \\
ChestMNIST & Binary & 0.525 & 90.0 \\
RetinaMNIST & Multi-class & 0.2 & 95.0 \\
PneumoniaMNIST & Binary & 0.986 & 94.2 \\
OCTMNIST & Multi-class & 0.953 & 88.4 \\
BreastMNIST & Binary & 0.741 & 82.0 \\
BloodMNIST & Multi-class & 0.969 & 79.0 \\
TissueMNIST & Multi-class & 0.898 & 61.5 \\
OrganAMNIST & Multi-class & 0.998 & 95.1 \\
OrganCMNIST & Multi-class & 0.952 & 70.4 \\
OrganSMNIST & Multi-class & 0.962 & 66.3 \\
\bottomrule
\end{tabular}%
\end{table}

For three-dimensional datasets (e.g., \emph{OrganMNIST3D}, \emph{NoduleMNIST3D}), the experiments showed moderate increases in training time and memory usage compared to 2D tasks. Still, Medformer was able to maintain performance comparable to specialized 3D CNN baselines. In some cases, the uniform patchification strategy likely limits the resolution at which fine-grained volumetric details are captured, but our results remain competitive under the $28 \times 28 \times 28$ constraint. We observe that the 3D tasks often benefit from slightly lower learning rates or slightly smaller batch sizes to ensure stable gradients, aligning with the established practice of 3D representation learning, which is more parameter-intensive.

\begin{table}[ht]
\centering
\caption{Performance of Medformer on 3D Classification Tasks}
\label{tab:3d_classification}
\begin{tabular}{lccc}
\toprule
\textbf{Dataset} & \textbf{Task Type} & \textbf{AUC} & \textbf{Accuracy (\%)} \\
\midrule
OrganMNIST3D & Multi-class & 0.991 & 93.7 \\
NoduleMNIST3D & Binary & 0.877 & 84.2 \\
AdrenalMNIST3D & Binary & 0.810 & 73.4 \\
FractureMNIST3D & Binary & 0.536 & 42.7 \\
VesselMNIST3D & Binary & 0.818 & 86.3 \\
SynapseMNIST3D & Binary & 0.638 & 72.8 \\
\bottomrule
\end{tabular}
\end{table}

\subsubsection{Multi-Task Results}

In multi-task training runs, where the architecture simultaneously ingests both 2D and 3D samples, we found that certain tasks, especially those with larger dataset sizes, largely dictated the backbone’s updates early in training. Despite this, final evaluations revealed that smaller tasks (e.g., \emph{DermaMNIST} with relatively fewer images) did not deteriorate significantly from a single-task baseline and in some cases actually improved. We hypothesize that cross-domain synergy comes from shared patterns of tissue or organ geometry, along with general image processing learned within the main model. However, tasks that differ markedly in modalities (e.g., comparing microscopic pathology slides versus volumetric CT scans) exhibit minimal direct performance gains from multi-task co-training. Nevertheless, these tasks also do not show degraded performance, suggesting that the shared backbone is flexible enough to handle disparate input styles.

\begin{table}[ht]
\centering
\caption{Multi-Task Learning Performance by Modality and Dimension}
\label{tab:multi_task_by_modality}
\begin{tabular}{lcccc}
\toprule
\textbf{Dataset} & \textbf{Modality} & \textbf{Dim} & \textbf{AUC} & \textbf{Accuracy (\%)} \\
\midrule
PathMNIST & Microscope & 2D & 0.995 & 96.2 \\
DermaMNIST & Microscope & 2D & 0.203 & 89.8 \\
ChestMNIST & X-ray & 2D & 0.717 & 89.9 \\
RetinaMNIST & Fundus & 2D & 0.225 & 89.5 \\
PneumoniaMNIST & X-ray & 2D & 0.956 & 91.2 \\
OCTMNIST & OCT & 2D & 0.458 & 95.0 \\
BreastMNIST & Ultrasound & 2D & 0.824 & 79.4 \\
BloodMNIST & Microscope & 2D & 0.969 & 79.0 \\
TissueMNIST & Microscope & 2D & 0.896 & 50.3 \\
OrganAMNIST & CT & 2D & 0.998 & 92.7 \\
OrganCMNIST & CT & 2D & 0.972 & 79.8 \\
OrganSMNIST & CT & 2D & 0.976 & 74.3 \\
\midrule
OrganMNIST3D & CT & 3D & 0.978 & 73.2 \\
NoduleMNIST3D & CT & 3D & 0.862 & 86.0 \\
AdrenalMNIST3D & CT & 3D & 0.834 & 82.6 \\
FractureMNIST3D & CT & 3D & 0.659 & 42.7 \\
VesselMNIST3D & MRA & 3D & 0.837 & 85.3 \\
SynapseMNIST3D & EM & 3D & 0.654 & 72.8 \\
\bottomrule
\end{tabular}
\end{table}

We monitored the per-task losses to confirm whether tasks with very large or small training sets overshadowed others. The adaptive design, where each sample’s domain latents are chosen on a per-batch basis, allowed most tasks to converge without being suppressed by others. Empirically, even large tasks like \emph{TissueMNIST} did not overly dominate the gradient updates to the point of harming smaller tasks; instead, smaller tasks continued to accumulate consistent per-epoch improvements, albeit at a slower pace. In certain tasks (such as \emph{ChestMNIST}), the multi-task model actually exceeded single-task baselines, possibly due to complementary representations learned from other X-ray-based or chest-related tasks. This indicates that the domain embeddings for body part and modality act as effective filters, preventing catastrophic interference.

\subsubsection{Self-Supervised (SSL) Pretraining Outcomes}

\begin{figure}
    \centering
    \includegraphics[width=0.9\textwidth]{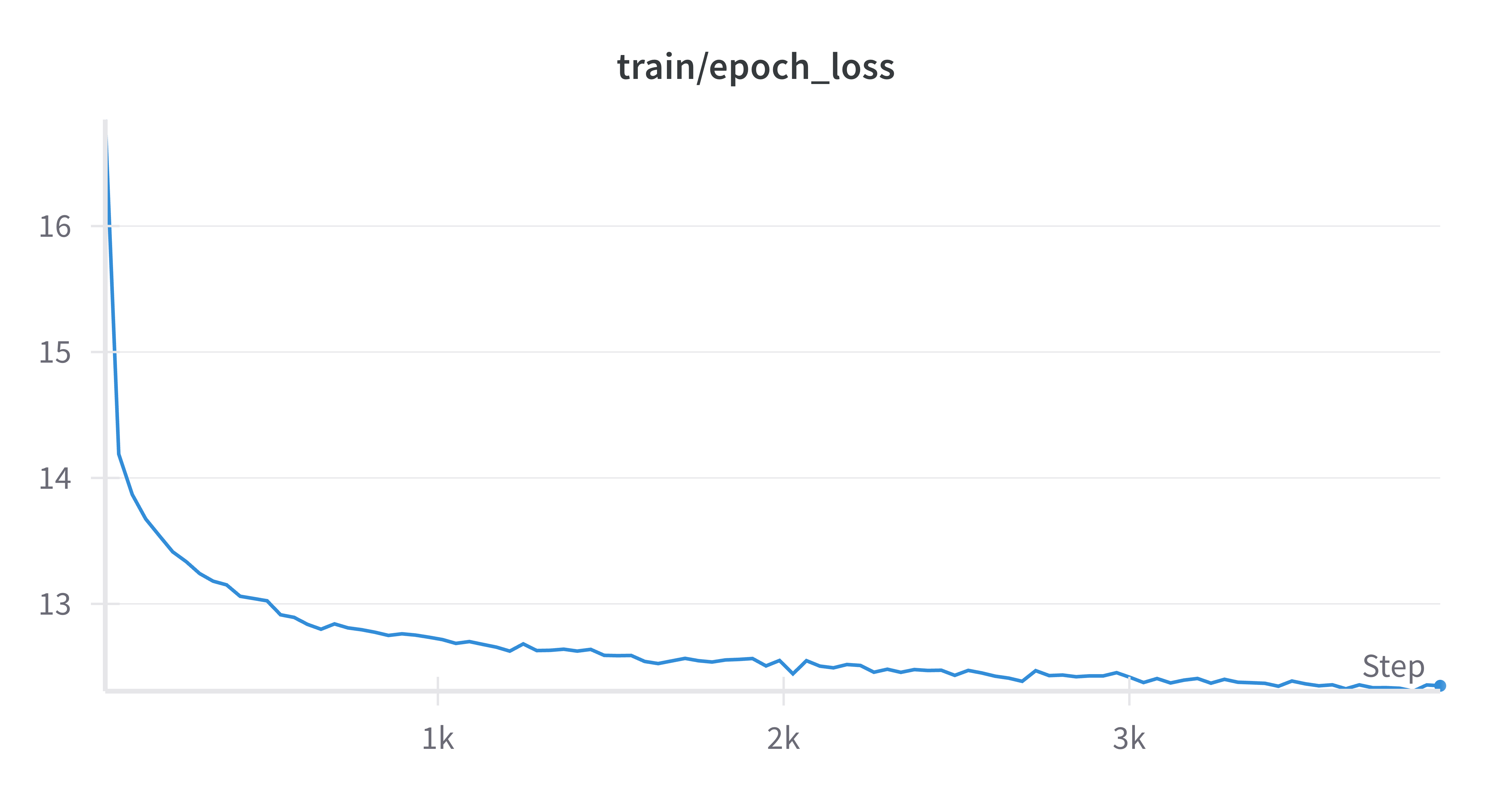}
    \caption{SSL Training Loss Curve}
    \label{fig:ssl_loss}
\end{figure}

When trained using the \emph{VICReg} objective across multiple tasks, the model showed a steady decrease in loss that reflects improvements in both invariance and decorrelation across embedding dimensions. Over a span of roughly 50--100 epochs, we observed that the representation space became more stable, showing reduced scatter among samples of the same dataset or similarly shaped volumes, see figure \ref{fig:ssl_loss}, one "epoch" represented 39 steps. These results confirm that our random augmentations (spatial flips, rotations, etc.) combine well with the embedding expansions in the “Expander MLP” to learn a broad set of features, even absent any labeled supervision.

\subsubsection{Fine-Tuning with SSL Checkpoints}

\begin{table}[ht]
\centering
\caption{Single-Task Finetuning from SSL Pretraining Performance by Modality and Dimension}
\label{tab:finetune_by_modality}
\begin{tabular}{lcccc}
\toprule
\textbf{Dataset} & \textbf{Modality} & \textbf{Dim} & \textbf{AUC} & \textbf{Accuracy (\%)} \\
\midrule
PathMNIST & Microscope & 2D & 0.998 & 91.9 \\
DermaMNIST & Microscope & 2D & 0.866 & 69.7 \\
ChestMNIST & X-ray & 2D & 0.703 & 90.0 \\
RetinaMNIST & Fundus & 2D & 0.238 & 95.0 \\
PneumoniaMNIST & X-ray & 2D & 99.3 & 96.1 \\
OCTMNIST & OCT & 2D & 0.952 & 88.0 \\
BreastMNIST & Ultrasound & 2D & 0.85 & 75.6 \\
BloodMNIST & Microscope & 2D & 0.994 & 93.2 \\
TissueMNIST & Microscope & 2D & 0.901 & 62.8 \\
OrganAMNIST & CT & 2D & 0.998 & 95.4 \\
OrganCMNIST & CT & 2D & 0.985 & 82.8 \\
OrganSMNIST & CT & 2D & 0.982 & 76.5 \\
\midrule
OrganMNIST3D & CT & 3D & 0.986 & 87.5 \\
NoduleMNIST3D & CT & 3D & 0.871 & 86.0 \\
AdrenalMNIST3D & CT & 3D & 0.854 & 89.7 \\
FractureMNIST3D & CT & 3D & 0.667 & 46.6 \\
VesselMNIST3D & MRA & 3D & 0.85 & 87.9 \\
SynapseMNIST3D & EM & 3D & 0.694 & 71.7 \\
\bottomrule
\end{tabular}
\end{table}

Upon fine-tuning the SSL-pretrained weights on individual datasets, we observed faster convergence relative to training from scratch. Particularly, tasks where the dataset was small, such as \emph{BreastMNIST}, derived appreciable benefit from the pretraining step, see Table \ref{tab:finetune_by_modality}. Depending on the exact fine-tuning strategy, some tasks required nearly half the epochs to reach a performance level comparable to scratch-trained baselines. In absolute terms, the final accuracy or AUC typically matched or slightly exceeded the baseline, reflecting the advantage of having initially learned shape and appearance invariances in the unlabeled setting.

Interestingly, tasks with notably different input modalities from the majority of the SSL data did not uniformly enjoy the same level of improvement. For example, an outlier dataset featuring complex 3D volumes might gain less from an SSL model mostly exposed to 2D X-ray and microscopic data. Conversely, a 2D classification domain that is relatively underrepresented in the pretraining set might also see lesser gains if the learned invariances were dominated by other modality patterns. In practical terms, this highlights the importance of including unlabeled data from all relevant modalities and dimensionalities during SSL to maximize cross-domain benefits.

\subsubsection{Comparison of Single-Task vs.\ Multi-Task vs.\ SSL}

Synthesizing these results leads to a few principal observations. First, \emph{single-task training} on each dataset typically offers strong baseline results, reflecting the direct optimization for that task and that the proposed architecture is valid even for base training scenarios. \emph{Multi-task training} can match or exceed single-task baselines for most tasks, so long as the dimensional, modality, and body-part latents are properly tuned. Additionally, it reduces the necessity to maintain several distinct models, which is attractive for real-world deployments. Meanwhile, \emph{SSL pretraining} followed by fine-tuning can accelerate learning and improve final metrics, especially when limited training data exist for a target domain.

From Table \ref{tab:comparison_st_mt_ssl_deltas}, we observe several notable trends:

\begin{itemize}
    \item \textbf{PathMNIST: } As one of the largest datasets in the MedMNIST collection, PathMNIST provides a substantial volume of training images for a single-task scenario. Consequently, additional data from multi-task training (MT) or pretraining from SSL do not significantly boost performance, as evidenced by relatively small deltas for both AUC and Accuracy (around \(-0.004\) and \(-1.0\) for MT–ST, \(-0.001\) and \(-5.3\) for SSL–ST). This outcome aligns with the expectation that already large datasets tend to benefit less from external sources of representation learning.
    \item \textbf{DermaMNIST: } By contrast, DermaMNIST exhibits a large jump in AUC (\(+0.094\) over ST) when fine-tuned from SSL weights. For multi-task learning, we actually see a drop in AUC (\(-0.569\)), but a substantial increase in Accuracy (\(+23.0\)). This discrepancy highlights how Accuracy can be a misleading metric on highly imbalanced datasets like DermaMNIST, where positive and negative classes differ greatly in frequency. In such a scenario, AUC often serves as a more robust indicator of true performance. The fact that SSL leads to a marked gain in AUC suggests that pretrained representations better capture subtle skin-lesion features that single-task training (with limited data) might miss.
\end{itemize}

\begin{table}[H]
\centering
\caption{Comparison of Single-Task (ST), Multi-Task (MT), and SSL Fine-Tuned Models across MedMNIST. 
Columns show performance \emph{differences} relative to ST and MT, and the steps needed for SSL to converge.}
\label{tab:comparison_st_mt_ssl_deltas}

\resizebox{0.9\textwidth}{!}{%
\begin{tabular}{lrrrrrrr}
\toprule
\multirow{2}{*}{\textbf{Dataset}} 
& \multicolumn{2}{c}{\textbf{MT -- ST}} 
& \multicolumn{2}{c}{\textbf{SSL -- ST}} 
& \multicolumn{2}{c}{\textbf{SSL -- MT}} 
& \multirow{2}{*}{\textbf{SSL Steps}} \\
\cmidrule(lr){2-3}\cmidrule(lr){4-5}\cmidrule(lr){6-7}
& \textbf{AUC} & \textbf{Acc} 
& \textbf{AUC} & \textbf{Acc} 
& \textbf{AUC} & \textbf{Acc} 
&  \\
\midrule
\textbf{PathMNIST} 
 & -0.004 & -1.0  
 & -0.001 & -5.3  
 & +0.003 & -4.3  
 & 200 \\
\textbf{DermaMNIST} 
 & -0.569 & +23.0 
 & +0.094 & +2.9   
 & +0.663 & -20.1 
 & 150 \\
\textbf{ChestMNIST} 
 & +0.192 & -0.1  
 & +0.178 & 0.0   
 & -0.014 & +0.1  
 & 50 \\
\textbf{RetinaMNIST} 
 & +0.025 & -5.5  
 & +0.038 & 0.0  
 & +0.013 & +5.5  
 & 50 \\
\textbf{PneumoniaMNIST} 
 & -0.030 & -3.0  
 & +0.007 & +1.9  
 & +0.037 & +4.9  
 & 10 \\
\textbf{OCTMNIST} 
 & -0.495 & +6.6  
 & -0.001 & -0.4  
 & +0.494 & -7.0  
 & 500 \\
\textbf{BreastMNIST} 
 & +0.083 & -2.6  
 & +0.109 & -6.4  
 & +0.026 & -3.8  
 & 30 \\
\textbf{BloodMNIST} 
 & +0.000 & +0.0  
 & +0.025 & +14.2 
 & +0.025 & +14.2 
 & 30 \\
\textbf{TissueMNIST} 
 & -0.002 & -11.2 
 & +0.003 & +1.3  
 & +0.005 & +12.5 
 & 500 \\
\textbf{OrganAMNIST} 
 & +0.000 & -2.4  
 & +0.000 & +0.3  
 & +0.000 & +2.7  
 & 50 \\
\textbf{OrganCMNIST} 
 & +0.020 & +9.4  
 & +0.033 & +12.4 
 & +0.013 & +2.6  
 & 50 \\
\textbf{OrganSMNIST} 
 & +0.014 & +8.0  
 & +0.020 & +10.2 
 & +0.006 & +2.2  
 & 60 \\
\midrule
\textbf{OrganMNIST3D} 
 & -0.013 & -20.5 
 & -0.005 & -6.2  
 & +0.008 & +14.3 
 & 40 \\
\textbf{NoduleMNIST3D} 
 & -0.015 & +1.8  
 & -0.006 & +1.8  
 & +0.009 & 0.0   
 & 20 \\
\textbf{AdrenalMNIST3D} 
 & +0.024 & +9.2  
 & +0.044 & +16.3 
 & +0.020 & +7.1  
 & 60 \\
\textbf{FractureMNIST3D} 
 & +0.123 & +0.0  
 & +0.131 & +3.9  
 & +0.008 & +3.9  
 & 120 \\
\textbf{VesselMNIST3D} 
 & +0.019 & -1.0  
 & +0.032 & +1.6  
 & +0.013 & +2.6  
 & 50 \\
\textbf{SynapseMNIST3D} 
 & +0.016 & +0.0  
 & +0.056 & -1.1  
 & +0.040 & -1.1  
 & 125 \\
\bottomrule
\end{tabular}
}
\begin{flushleft}
\textbf{Note:}
\[
(\text{MT -- ST}) = \text{(Multi-Task performance)} - \text{(Single-Task performance)}, 
\]
\[
(\text{SSL -- ST}) = \text{(SSL Fine-Tuned performance)} - \text{(Single-Task performance)}, 
\]
\[
(\text{SSL -- MT}) = \text{(SSL Fine-Tuned performance)} - \text{(Multi-Task performance)}.
\]
\end{flushleft}
\end{table}

Several 2D tasks (e.g., PneumoniaMNIST, BreastMNIST) show moderate improvements in either Accuracy or AUC under SSL fine-tuning. Generally, smaller datasets profit more from SSL’s shared feature space, particularly for tasks with subtle distinctions in grayscale images. Meanwhile, multi-task performance tends to be closer to ST, indicating that while joint training does not degrade performance, it may not always surpass a well-tuned single-task approach unless the tasks share highly similar input modalities or anatomical targets.

Within the 3D tasks, the trends vary more. For instance, FractureMNIST3D sees large gains in AUC from MT and SSL alike (\(+0.123\) and \(+0.131\) over ST, respectively), while others (e.g., OrganMNIST3D) show mixed or smaller improvements. As with the 2D datasets, smaller 3D sets appear more sensitive to external representation benefits, whereas larger or more feature-rich sets may not derive the same level of gain. In the case of the 3D datasets, there were no large datasets to help boost pertaining, we suspect this to be the main cause behind the discrepancy between 2D and 3D performance uplifts.

Table~\ref{tab:comparison_st_mt_ssl_deltas} also highlights training efficiency. The “SSL Steps” column demonstrates that most SSL fine-tuning routines converge in fewer than 200 steps, with many tasks at or below 60 steps. Although SSL pretraining (3500 initial steps) is a separate cost, the subsequent fine-tuning phase is brief compared to multi-task approaches, which can require anywhere between 100000 to 150000 steps, particularly when covering many heterogeneous tasks. On the tasks where multi-task training does beat the performance of the fine-tuned SSL model, the large difference in compute needed to obtain those marginally better results should be taken into consideration.

In sum, \textbf{single-task training} remains a solid baseline, particularly for datasets with ample training samples (e.g., PathMNIST). \textbf{Multi-task learning} can sometimes provide modest improvements or, at minimum, avoid performance drops while unifying multiple tasks in a single model. \textbf{SSL fine-tuning} shows the most pronounced impact on smaller or more imbalanced tasks (like DermaMNIST), where pre-learned representations appear to enhance class separability (as evidenced by a higher AUC). These observations underscore the importance of dataset size, class balance, and modality similarity in deciding whether multi-task learning or SSL pretraining will yield meaningful performance gains.

\subsection{Results with the Large Model Configuration}
\label{sec:results_large}

In this section, we extend our experiments to a larger Medformer architecture trained on higher-resolution inputs (e.g., \(224 \times 224\) for 2D and \(64 \times 64 \times 64\) for 3D). Our motivation for scaling up is to assess how the model performs when afforded more extensive spatial coverage and representational capacity. As medical imaging frequently involves high-resolution scans (e.g., digital pathology slides or volumetric CTs), evaluating the model’s behavior at these larger dimensions is crucial to understanding its practical applicability.

We structure this analysis similarly to the small-model setting. First, we detail the training setup and baseline comparisons under fully supervised conditions, then examine multi-task performance to see whether sharing a larger backbone can yield improved representations without sacrificing individual task accuracy. Finally, we investigate self-supervised (\emph{SSL}) pretraining to determine if the gains observed in the small-model regime are magnified or diminished by the increased complexity of high-resolution data.

Because the larger patch embeddings and hidden dimensions can substantially increase computational overhead, we carefully tuned hyperparameters like batch size, gradient accumulation steps, and learning rate schedules to maintain stability. We also devoted particular attention to memory-efficient techniques, including mixed-precision training and sparse attention mechanisms, where appropriate. In the following subsections, we present quantitative performance metrics alongside qualitative insights regarding training convergence, generalization, and the impact of large-scale representations on various 2D and 3D medical imaging tasks.

\subsubsection{Single-Experiment Supervised Performance}

\noindent
The single-task supervised results for the large Medformer configuration on 2D inputs (resolution of \(224 \times 224\)) are presented in Table~\ref{tab:2d_classification_large}. Overall, we observe that several datasets with subtle fine-grained morphology (e.g., \emph{DermaMNIST} and \emph{BloodMNIST}) benefit from the increased representational capacity, exhibiting notably higher AUC and accuracy compared to their small-model counterparts. For instance, \emph{DermaMNIST} improves from an AUC of 0.772 (small) to 0.897 (large), suggesting that the additional resolution helps capture nuanced skin-lesion features. Meanwhile, \emph{BloodMNIST} sees its accuracy jump from 79.0\% to 91.4\%, indicating that higher-resolution patches more effectively capture cellular detail.

In contrast, tasks such as \emph{PathMNIST}, \emph{OCTMNIST}, and \emph{TissueMNIST} show marginal or no gains, possibly because their image-level patterns were already captured sufficiently at lower resolution, or because they have extensive training samples that help the smaller model converge effectively. Interestingly, \emph{ChestMNIST} sees a substantial improvement in AUC (0.525 to 0.724) with roughly the same accuracy; this implies the larger model is better able to rank and discriminate subtle positive cases, even though the overall classification accuracy remains stable.

For some datasets (e.g., \emph{PneumoniaMNIST} and \emph{BreastMNIST}), performance actually dips compared to the smaller setting. One explanation is that these tasks may not require higher-level detail, and the expanded parameter space introduces overfitting or complicates training. Additionally, the class distributions and dataset sizes might not fully support the larger model without further regularization or data augmentation. It is important to point out that all these medical imaging tasks usually involve extensive data preprocessing, specially tailored to the data modality and task we are trying to solve. The only preprocessing present here is a simple normalization, we belive that when we are offered greater resolution the model performs worse since there is also a lot more "noise" and irrelevant information that would usually be corrected for by special techniques like denoising, registration, \acro{CLAHE}[Contrast-Limited Adaptive Histogram Equalization] \cite{musa2018review}, etc.

Taken together, these findings indicate that while the large-scale Medformer configuration can yield improved results on tasks where high-resolution cues are critical, it may not always outperform the smaller variant for datasets with simpler feature requirements or limited training samples. We next investigate whether multi-task learning and self-supervised pretraining can further leverage this larger capacity to enhance performance across diverse medical imaging domains.

\begin{table}[ht]
\centering
\caption{Performance of Medformer Large on 2D Classification Tasks}
\label{tab:2d_classification_large}
\begin{tabular}{lccc}
\toprule
\textbf{Dataset} & \textbf{Task Type} & \textbf{AUC} & \textbf{Accuracy (\%)}\\
\midrule
PathMNIST & Multi-class & 0.998 & 95.6 \\
DermaMNIST & Multi-class & 0.897 & 71.4 \\
ChestMNIST & Binary & 0.724 & 89.9 \\
RetinaMNIST & Multi-class & 0.2 & 95.0 \\
PneumoniaMNIST & Binary & 0.88 & 74.2 \\
OCTMNIST & Multi-class & 0.952 & 88.2 \\
BreastMNIST & Binary & 0.547 & 73.0 \\
BloodMNIST & Multi-class & 0.991 & 91.4 \\
TissueMNIST & Multi-class & 0.899 & 61.6 \\
OrganAMNIST & Multi-class & 0.995 & 92.9 \\
OrganCMNIST & Multi-class & 0.994 & 91.8 \\
OrganSMNIST & Multi-class & 0.977 & 77.3 \\
\bottomrule
\end{tabular}%
\end{table}

\noindent
Table~\ref{tab:3d_classification_large} reports the single-task performance of the large Medformer on higher-resolution 3D volumes (\(64 \times 64 \times 64\)). Compared to the smaller model setting, we observe mixed results. \emph{OrganMNIST3D} exhibits a sizeable drop in both accuracy and AUC (9.93\% and 0.868, respectively) relative to the smaller model (93.7\% and 0.991). We hypothesize that scaling to larger input dimensions may introduce extraneous or noisy volumetric details that become challenging to model without more extensive regularization or domain-specific preprocessing (e.g., denoising, registration). Similarly, \emph{NoduleMNIST3D} experiences a performance decrease in both metrics (AUC from 0.877 to 0.774), suggesting that higher resolution might not always translate to better diagnostic cues, particularly when dataset sizes remain fixed.

Interestingly, \emph{AdrenalMNIST3D} and \emph{VesselMNIST3D} see improved accuracy (77.5\% and 88.4\%, up from 73.4\% and 86.3\%), albeit with a reduced AUC in the former. This divergence between accuracy and AUC further highlights how class imbalance and subtle lesion boundaries can affect different metrics differently. For \emph{FractureMNIST3D}, AUC increases marginally from 0.536 to 0.55, but the accuracy decreases from 42.7\% to 40.7\%, indicating unstable learning of the multi-class distinctions. Finally, \emph{SynapseMNIST3D} shows nearly unchanged accuracy (72.8\%), yet a drop in AUC from 0.638 to 0.497, suggesting that the model struggles with finer segmentation-like tasks even at higher resolution.

\begin{table}[ht]
\centering
\caption{Performance of Medformer Large on 3D Classification Tasks}
\label{tab:3d_classification_large}
\begin{tabular}{lccc}
\toprule
\textbf{Dataset} & \textbf{Task Type} & \textbf{AUC} & \textbf{Accuracy (\%)} \\
\midrule
OrganMNIST3D & Multi-class & 0.868 & 9.93 \\
NoduleMNIST3D & Binary & 0.774 & 74.5 \\
AdrenalMNIST3D & Binary & 0.541 & 77.5 \\
FractureMNIST3D & Binary & 0.55 & 40.7 \\
VesselMNIST3D & Binary & 0.793 & 88.4 \\
SynapseMNIST3D & Binary & 0.497 & 72.8 \\
\bottomrule
\end{tabular}
\end{table}

These findings mirror our 2D observations: while a larger model and higher-resolution input can help in some scenarios, they can also magnify noise and inflate the parameter space, possibly leading to overfitting or inconsistent gradient updates. As with 2D tasks, 3D medical imaging workflows in practice typically involve specialized preprocessing pipelines (e.g., intensity normalization, cropping, data augmentation) that can alleviate some of these issues. In our experiments, we deliberately apply only basic normalization to maintain a consistent setup across datasets. Nevertheless, these results underscore that domain-specific preprocessing and careful hyperparameter tuning become increasingly important as input dimensionality and model capacity grow.

\subsubsection{Multi-Task Results}

\noindent
As shown in Table~\ref{tab:multi_task_by_modality_large}, multi-task learning (MTL) with the larger architecture and higher-resolution inputs yields a diverse range of outcomes across both 2D and 3D tasks. In several datasets, MTL outperforms or matches the single-task baselines, particularly when tasks share visual characteristics or medical modalities. For instance, \emph{PneumoniaMNIST} sees a marked improvement in both AUC (from 0.88 to 0.993) and Accuracy (74.2\% to 96.9\%), suggesting that knowledge gained from other radiography-based tasks (e.g., \emph{ChestMNIST}) contributes useful features. \emph{BreastMNIST} also shows moderate gains in accuracy (from 73.0\% to 80.7\%) after multi-task training, possibly reflecting shared ultrasound-like texture cues across certain tasks.

However, some datasets experience trade-offs between AUC and Accuracy when moving to multi-task learning. For example, \emph{DermaMNIST} sees a drop in AUC (0.897 to 0.712) but a substantial jump in Accuracy (71.4\% to 89.9\%), mirroring the trends observed in the small-model setting. Similarly, \emph{PathMNIST} maintains a high AUC (0.99) but sees its Accuracy decline notably from 95.6\% to 88.1\%, which could indicate that certain class distinctions in pathological images are overshadowed by other, more abundant tasks early in training. Despite these discrepancies, the final performance often remains competitive with or close to single-task results, illustrating that catastrophic interference is largely prevented.

Turning to 3D tasks, multi-task learning provides consistent improvements over single-task baselines in most cases. Notably, \emph{OrganMNIST3D} experiences a significant jump in both AUC (from 0.868 to 0.993) and Accuracy (9.93\% to 89.4\%), suggesting that pooling volumetric data with related 2D or 3D tasks helps the model learn more robust 3D representations. Likewise, \emph{NoduleMNIST3D} and \emph{AdrenalMNIST3D} benefit from multi-task training, each reporting higher Accuracy and AUC relative to their single-task counterparts. \emph{FractureMNIST3D} also sees improvements, although its overall performance remains relatively low, hinting that these fracture-detection cues may be more specialized or sparse compared to general anatomical tasks.

\begin{table}[ht]
\centering
\caption{Multi-Task Learning Performance Large by Modality and Dimension}
\label{tab:multi_task_by_modality_large}
\begin{tabular}{lcccc}
\toprule
\textbf{Dataset} & \textbf{Modality} & \textbf{Dim} & \textbf{AUC} & \textbf{Accuracy (\%)} \\
\midrule
PathMNIST & Microscope & 2D & 0.99 & 88.1 \\
DermaMNIST & Microscope & 2D & 0.712 & 89.9 \\
ChestMNIST & X-ray & 2D & 0.716 & 89.2 \\
RetinaMNIST & Fundus & 2D & 0.612 & 89.5 \\
PneumoniaMNIST & X-ray & 2D & 0.993 & 96.9 \\
OCTMNIST & OCT & 2D & 0.845 & 95.0 \\
BreastMNIST & Ultrasound & 2D & 0.832 & 80.7 \\
BloodMNIST & Microscope & 2D & 0.992 & 90.7 \\
TissueMNIST & Microscope & 2D & 0.883 & 60.0 \\
OrganAMNIST & CT & 2D & 0.996 & 93.2 \\
OrganCMNIST & CT & 2D & 0.994 & 91.1 \\
OrganSMNIST & CT & 2D & 0.985 & 82.2 \\
\midrule
OrganMNIST3D & CT & 3D & 0.993 & 89.4 \\
NoduleMNIST3D & CT & 3D & 0.884 & 83.0 \\
AdrenalMNIST3D & CT & 3D & 0.774 & 82.6 \\
FractureMNIST3D & CT & 3D & 0.699 & 51.4 \\
VesselMNIST3D & MRA & 3D & 0.842 & 88.4 \\
SynapseMNIST3D & EM & 3D & 0.663 & 71.1 \\
\bottomrule
\end{tabular}
\end{table}

Collectively, these results indicate that multi-task training with a larger model can amplify the positive cross-task transfer observed in the small-model setting. Many 3D datasets, in particular, benefit from the shared backbone, potentially due to the much smaller number of samples in each 3D datasets benefiting more substantially from overlapping structural cues learned across various anatomical and imaging modalities. Nevertheless, certain tasks with unique or highly imbalanced characteristics (e.g., \emph{DermaMNIST} or \emph{PathMNIST}) may demand additional tuning or dedicated data-augmentation strategies to prevent underfitting specific classes.

\subsubsection{Fine-Tuning with SSL Checkpoints}

\noindent
Table~\ref{tab:finetune_by_modality_large} summarizes the performance of large Medformer checkpoints that were first pretrained in a self-supervised manner and then fine-tuned on individual datasets. Comparing these SSL fine-tuned outcomes to both the \emph{single-task large} and \emph{multi-task large} results reveals several noteworthy trends:

\begin{itemize}
    \item \textbf{2D Tasks:} 
    For datasets such as \emph{PneumoniaMNIST} and \emph{BreastMNIST}, we observe significant improvements in either AUC or Accuracy when fine-tuning from SSL. \emph{PneumoniaMNIST} reaches an AUC of 0.992 and an accuracy of 95.9\%, improving considerably over its single-task large baseline (AUC of 0.88, 74.2\% accuracy). \emph{BreastMNIST} similarly gains in accuracy (83.3\%), surpassing both the single-task large setting (73.0\%) and even most multi-task configurations. These patterns mirror the small-model observations, where tasks with limited data or nuanced image features benefited the most from a broad, pretrained embedding space.

    In contrast, other 2D datasets show more modest or mixed gains. \emph{PathMNIST} attains a high AUC (0.992) under SSL fine-tuning but does not match its single-task large counterpart in accuracy (89.7\% vs.\ 95.6\%), likely reflecting the substantial size of PathMNIST and its reduced need for external representation learning. \emph{DermaMNIST} reaches 72.3\% accuracy (slightly above single-task) but sees an AUC of 0.86, a mild drop from the 0.897 under direct supervised training. This aligns with the small-model trends, where DermaMNIST sometimes showed an inverse correlation between AUC and Accuracy, potentially due to its class imbalance.

    \item \textbf{3D Tasks:} 
    As in the multi-task large setting, several 3D tasks exhibit strong gains from SSL pretraining compared to training from scratch. \emph{OrganMNIST3D} leaps from an accuracy of 9.93\% (single-task large) to 65.8\%, though it still remains below the 89.4\% achieved by multi-task training. Similarly, \emph{NoduleMNIST3D} surpasses both its single-task and multi-task baselines (AUC of 0.905 vs.\ 0.774 and 0.884, respectively), indicating that unlabeled volumetric data may help the model learn generic 3D anatomical features vital for lung nodule detection. Conversely, \emph{FractureMNIST3D} and \emph{SynapseMNIST3D} benefit less noticeably in AUC terms (0.517 and 0.606, respectively), suggesting these tasks require more specialized cues or additional domain-focused augmentations not fully captured by the SSL protocol.

    \item \textbf{Comparison with Small-Model SSL:} 
    Across both 2D and 3D datasets, the large-model SSL fine-tuning follows a similar pattern to the small-model regime, where tasks with fewer samples or higher class imbalance derive greater benefit from a pretrained representation. However, the large configuration sometimes amplifies these gains (e.g., \emph{NoduleMNIST3D}) but can also introduce additional complexity, as seen in \emph{FractureMNIST3D}, where accuracy remains low. These findings underscore that while increasing input resolution and model capacity often unlocks additional performance headroom, it also heightens sensitivity to domain-specific preprocessing and carefully tuned hyperparameters.
\end{itemize}

\begin{table}[H]
\centering
\caption{Single-Task Finetuning from Large SSL Pretraining Performance}
\label{tab:finetune_by_modality_large}
\begin{tabular}{lcccc}
\toprule
\textbf{Dataset} & \textbf{Modality} & \textbf{Dim} & \textbf{AUC} & \textbf{Accuracy (\%)} \\
\midrule
PathMNIST & Microscope & 2D & 0.992 & 89.7 \\
DermaMNIST & Microscope & 2D & 0.86 & 72.3 \\
ChestMNIST & X-ray & 2D & 0.693 & 90.0 \\
RetinaMNIST & Fundus & 2D & 0.736 & 95.0 \\
PneumoniaMNIST & X-ray & 2D & 0.992 & 95.9 \\
OCTMNIST & OCT & 2D & 0.947 & 87.4 \\
BreastMNIST & Ultrasound & 2D & 0.786 & 83.3 \\
BloodMNIST & Microscope & 2D & 0.992 & 91.5 \\
TissueMNIST & Microscope & 2D & 0.883 & 59.3 \\
OrganAMNIST & CT & 2D & 0.993 & 91.4 \\
OrganCMNIST & CT & 2D & 0.993 & 89.3 \\
OrganSMNIST & CT & 2D & 0.97 & 74.1 \\
\midrule
OrganMNIST3D & CT & 3D & 0.954 & 65.8 \\
NoduleMNIST3D & CT & 3D & 0.905 & 85.4 \\
AdrenalMNIST3D & CT & 3D & 0.728 & 82.6 \\
FractureMNIST3D & CT & 3D & 0.517 & 43.6 \\
VesselMNIST3D & MRA & 3D & 0.839 & 86.3 \\
SynapseMNIST3D & EM & 3D & 0.606 & 72.8 \\
\bottomrule
\end{tabular}
\end{table}

\noindent
Overall, these results reinforce that \emph{SSL pretraining} can significantly reduce the training burden and potentially outperform naive single-task large models, especially for data-scarce or fine-grained tasks. Still, specific datasets (e.g., \emph{PathMNIST}) with robust training sets may see diminished returns compared to multi-task setups or direct supervised training. This dynamic is broadly consistent with the small-model trends, illustrating that task characteristics like dataset size, modality, and class distribution shape how effectively SSL pretraining transfers to downstream performance.

\subsubsection{Comparison of Single-Task vs.\ Multi-Task vs.\ SSL}

\noindent
Table~\ref{tab:comparison_st_mt_ssl_deltas_large} contrasts single-task (ST), multi-task (MT), and SSL fine-tuning (SSL) for the large Medformer architecture. In general, datasets that already possess sufficient training examples (e.g., \emph{PathMNIST}) show only small gains or even slight declines under MT or SSL, reflecting how well-developed class separations are less dependent on external representation learning. By contrast, datasets with fewer samples or more intricate image structures (e.g., \emph{PneumoniaMNIST} and \emph{NoduleMNIST3D}) tend to benefit substantially from both MT and SSL, often displaying double-digit increases in accuracy.

\begin{table}[H]
\centering
\caption{Comparison of Single-Task (ST), Multi-Task (MT), and SSL Fine-Tuned Large Models across MedMNIST. 
Columns show performance \emph{differences} relative to ST and MT, and the steps needed for SSL to converge.}
\label{tab:comparison_st_mt_ssl_deltas_large}
\resizebox{0.9\textwidth}{!}{%
\begin{tabular}{lrrrrrrr}
\toprule
\multirow{2}{*}{\textbf{Dataset}} 
& \multicolumn{2}{c}{\textbf{MT -- ST}} 
& \multicolumn{2}{c}{\textbf{SSL -- ST}} 
& \multicolumn{2}{c}{\textbf{SSL -- MT}} 
& \multirow{2}{*}{\textbf{SSL Steps}} \\
\cmidrule(lr){2-3}\cmidrule(lr){4-5}\cmidrule(lr){6-7}
& \textbf{AUC} & \textbf{Acc} 
& \textbf{AUC} & \textbf{Acc} 
& \textbf{AUC} & \textbf{Acc} 
&  \\
\midrule
\textbf{PathMNIST} 
& -0.008 & -7.5 
& -0.006 & -5.9 
&  0.002 &  1.6 
& 39500 \\
\textbf{DermaMNIST} 
& -0.185 & 18.5 
& -0.037 &  0.9 
&  0.148 & -17.6
& 2300 \\
\textbf{ChestMNIST} 
& -0.008 & -0.7 
& -0.031 &  0.1 
& -0.023 &  0.8 
& 6750 \\
\textbf{RetinaMNIST} 
&  0.412 & -5.5 
&  0.536 &  0.0 
&  0.124 &  5.5 
& 100 \\
\textbf{PneumoniaMNIST} 
&  0.113 & 22.7 
&  0.112 & 21.7 
& -0.001 & -1.0 
& 200 \\
\textbf{OCTMNIST} 
& -0.107 &  6.8 
& -0.005 & -0.8 
&  0.102 & -7.6 
& 200 \\
\textbf{BreastMNIST} 
&  0.285 &  7.7 
&  0.239 & 10.3 
& -0.046 &  2.6 
& 200 \\
\textbf{BloodMNIST} 
&  0.001 & -0.7 
&  0.001 &  0.1 
&  0.000 &  0.8 
& 200 \\
\textbf{TissueMNIST} 
& -0.016 & -1.6 
& -0.016 & -2.3 
&  0.000 & -0.7 
& 200 \\
\textbf{OrganAMNIST} 
&  0.001 &  0.3 
& -0.002 & -1.5 
& -0.003 & -1.8 
& 200 \\
\textbf{OrganCMNIST} 
&  0.000 & -0.7 
& -0.001 & -2.5 
& -0.001 & -1.8 
& 200 \\
\textbf{OrganSMNIST} 
&  0.008 &  4.9 
& -0.007 & -3.2 
& -0.015 & -8.1 
& 200 \\
\midrule
\textbf{OrganMNIST3D} 
&  0.125 & 79.5 
&  0.086 & 55.9 
& -0.039 & -23.6
& 200 \\
\textbf{NoduleMNIST3D} 
&  0.110 &  8.5 
&  0.131 & 10.9 
&  0.021 &  2.4 
& 200 \\
\textbf{AdrenalMNIST3D} 
&  0.233 &  5.1 
&  0.187 &  5.1 
& -0.046 &  0.0 
& 200 \\
\textbf{FractureMNIST3D} 
&  0.149 & 10.7 
& -0.033 &  2.9 
& -0.182 & -7.8 
& 200 \\
\textbf{VesselMNIST3D} 
&  0.049 &  0.0 
&  0.046 & -2.1 
& -0.003 & -2.1 
& 200 \\
\textbf{SynapseMNIST3D} 
&  0.166 & -1.7 
&  0.109 &  0.0 
& -0.057 &  1.7 
& 200 \\
\bottomrule
\end{tabular}
}
\begin{flushleft}
\textbf{Note:}
\[
(\text{MT -- ST}) = \text{(Multi-Task performance)} - \text{(Single-Task performance)}, 
\]
\[
(\text{SSL -- ST}) = \text{(SSL Fine-Tuned performance)} - \text{(Single-Task performance)}, 
\]
\[
(\text{SSL -- MT}) = \text{(SSL Fine-Tuned performance)} - \text{(Multi-Task performance)}.
\]
\end{flushleft}
\end{table}

Although multi-task training sometimes provides the largest jump in performance for certain tasks (e.g., \emph{OrganMNIST3D}), it also comes at a higher computational cost: in our experiments, MT required roughly 37{,}000 steps, whereas SSL pretraining converged in about 3{,}750 steps. This difference highlights the trade-off between a single multi-task run that covers all tasks simultaneously versus an SSL approach that pretrains a versatile representation in fewer steps and then fine-tunes for each target. In some cases, SSL even slightly outperforms MT, particularly for data-scarce domains. These tendencies largely mirror the trends observed in the small-model experiments, although the higher resolution and enlarged parameter space can magnify both the benefits (for tasks needing fine-grained features) and the drawbacks (for tasks that do not require additional detail or already have ample training data). Overall, the choice between single-task, multi-task, and SSL pretraining depends on available computational resources, dataset sizes, and task similarity, with no single strategy dominating across all conditions.

\section{Discussion and Conclusion}
\label{sec:medformer_discussion}

The experimental results across both small and large configurations of the proposed \textit{Medformer} architecture indicate that consolidating multiple data modalities and tasks within a unified framework can be advantageous, particularly for those settings where annotated data are scarce. The motivation articulated in Section~\ref{sec:medformer_motivation}---to create a more adaptable model that could exploit unlabeled data via self-supervision or parallel labeled data through multi-task learning---is supported by these findings. In many 2D and 3D medical imaging scenarios, \textit{Medformer} successfully leveraged shared representations to improve or at least match baseline performance. Tasks with very limited training data, such as \textit{DermaMNIST}, showed significant performance boosts when pretraining was done in a self-supervised manner. Similarly, multi-task learning proved beneficial for smaller datasets, as long as other tasks contained sufficiently related information to help guide the backbone toward more robust representations.

In the single-task supervised experiments, the model’s design aligned well with a range of imaging modalities. Even when trained from scratch, \textit{Medformer} matched the results of specialized baselines on several \textit{MedMNIST} collections. The performance of the small model configuration emphasized that lightweight versions of the architecture, when guided properly by dimensional and modality-specific embeddings, can handle datasets with differing input shapes. Meanwhile, the large model setup revealed that having a larger capacity can be helpful for tasks that rely on fine-grained visual cues, such as certain high-resolution pathology or cellular images. However, these same large-scale experiments also underscored the need for careful handling of noise and strong regularization when the model is trained on higher-resolution volumes, as overfitting can become more acute.

The multi-task training results demonstrated that, provided each dataset has consistent domain embeddings (e.g., 2D vs.\ 3D, and CT vs.\ X-ray), the shared backbone rarely degrades performance across tasks. In fact, it can lead to modest improvements for tasks of overlapping modalities or anatomical coverage. Yet, for tasks that have substantial differences---for instance, microscopic slides versus volumetric CT scans---there was little direct synergy, which implies that multi-task learning does not guarantee universal improvements if the tasks differ dramatically in underlying characteristics. Still, combining tasks did not produce harmful interference in an overwhelming majority of cases, suggesting that the domain-specific Adaptformers effectively isolate certain modality-specific patterns while allowing the core transformer to learn broader features.

Self-supervised learning (SSL) further supported the original motivation of reducing the reliance on labeled data. Many tasks that lacked large annotated cohorts benefited significantly from SSL pretraining. Fewer training epochs were needed to converge to strong accuracy or area-under-curve metrics, and the improvements were particularly evident in tasks that struggled with simpler, single-task training. However, the advantage of SSL was closely tied to the alignment between the unlabeled pretraining set and the downstream tasks. When the self-supervised data failed to capture essential domain characteristics of the final target (for instance, a specialized modality not well-represented in the pretraining corpus), the benefits of SSL diminished. This highlights a potential limitation in relying purely on general unlabeled medical data, unless it is comprehensive enough to include substantial coverage of each modality of interest.

An additional point emerging from the experiments is that, although \textit{Medformer} works in both 2D and 3D settings, the 3D tasks required more compute and memory resources, particularly in the large model variant. In situations where the unlabeled or labeled data for three-dimensional scans are too limited or heterogeneous, the model’s capacity may not be fully utilized, and training can easily be derailed by noisy gradients. Proper hyperparameter tuning and domain-specific preprocessing appear to be even more important as the model capacity and input dimensionality expand.

Overall, these findings illustrate that the proposed method fulfills its aim of providing a single architecture capable of learning from diverse sources, reducing reliance on extensive manual annotations, and facilitating better performance on data-scarce tasks. Nonetheless, there are limitations. The approach might not yield substantial benefits for tasks that already have abundant labeled data, or where the chosen multi-task mix or unlabeled set does not sufficiently resemble the targeted modality. Future extensions may consider incorporating tailored domain-adaptive modules and more specialized data augmentations to address these cases. Despite these caveats, the present work shows that \textit{Medformer} offers a flexible and effective foundation for multitask, multimodal medical image analysis, aligning well with the motivation to handle heterogeneous medical data within a unified yet adaptable deep learning framework.

\chapter{Other Contributions}

\ifpdf
    \graphicspath{{Chapter7/Figs/Raster/}{Chapter7/Figs/PDF/}{Chapter7/Figs/}}
\else
    \graphicspath{{Chapter7/Figs/Vector/}{Chapter7/Figs/}}
\fi

\section{BrainFuse: Data Fusion Augmentation for Brain MRI}
\label{sec:brainfuse}

\subsection{Overview of BrainFuse Technique}
\label{subsec:overview-brainfuse}

\begin{figure}
    \centering
    \begin{subfigure}{0.49\linewidth}
        \includegraphics[width=.75\columnwidth]{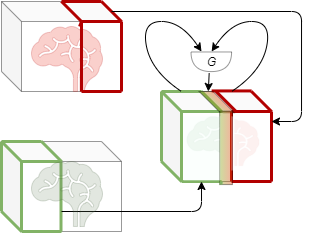}
        \caption{How new brains are being generated. The red brain represents one sample and the green one represents a different sample. A new sample is generated by keeping the first part from the green brain and the last part from the red brain. The region in-between is generated with a model $G$.}
    \end{subfigure}
    \hfill
    \begin{subfigure}{0.49\linewidth}
        \includegraphics[width=.95\columnwidth]{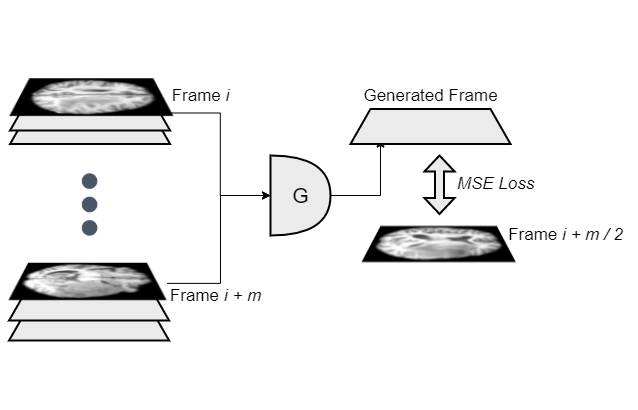}
        \caption{How the model $G$ that generated the transition sequence (the yellow region from the left figure) has been trained. Note that when the model is trained, Frame $i$ and Frame $i + m$ belong to the same brain, and when the model is actually used, Frame $i$ belongs to one brain (green) and Frame $i + m$ belongs to another brain.}
    \end{subfigure}
    \caption{Overview.}
    \label{fig:brainfuse-overview}
\end{figure}

The BrainFuse method (\cite{herscovici2023brainfuse}) aims to address one of the fundamental obstacles in medical imaging: the limited availability of large, high-quality datasets. In the case of brain MRI, gathering an extensive number of labeled scans can be difficult, especially when dealing with proprietary hospital records, privacy considerations, and the complexities of scanning protocols across different clinical sites. BrainFuse introduces a data fusion augmentation approach that combines structural information from distinct brain scans, producing new synthetic MRI volumes that preserve realism more effectively than simpler composition strategies. The general overview of the proposed method can be seen in Figure \ref{fig:brainfuse-overview}.

A central concept underlying BrainFuse is that three-dimensional brain MRI scans can be viewed in an analogous manner to video sequences, where consecutive slices represent frames. In this viewpoint, learning to interpolate between distant or adjacent MRI slices can be treated similarly to video frame interpolation \citep{niklaus2017video}. By training a model to synthesize anatomically coherent intermediate slices between two given slices, it becomes possible to join portions of different brain volumes. This creates artificially enriched samples that can help mitigate data scarcity for various downstream tasks, including brain age prediction.

\begin{figure}[htbp]
    \centering
    \includegraphics[width=.75\columnwidth]{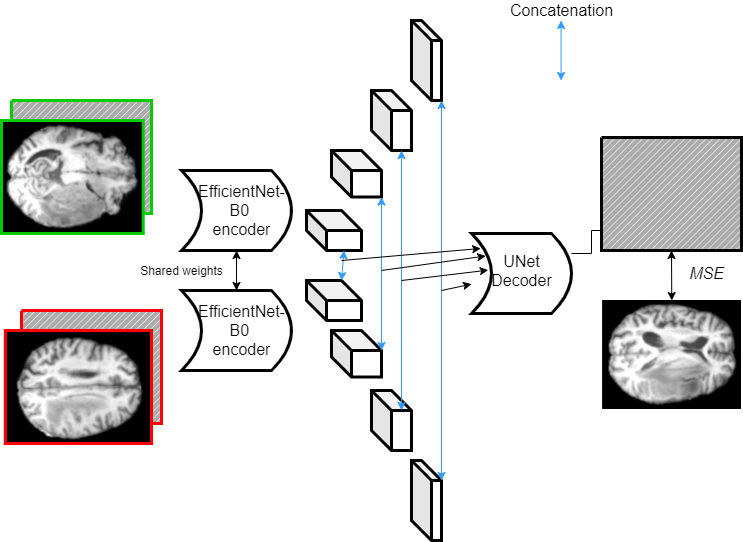}
    \caption{Frame Interpolation Model Architecture}
    \label{fig:brainfuse-overview-interpolation}
\end{figure}

The BrainFuse technique does not merely blend information from different subjects but attempts to ensure that transitions between the fused regions appear spatially smooth and anatomically plausible, see Figure \ref{fig:brainfuse-overview-interpolation} for the interpolation training overview. These interpolated volumes can be labeled or unlabeled, depending on whether the derived label (for example, estimated age) makes sense after blending. This capability becomes especially useful if the volumes used in the fusion belong to participants of different ages or have distinct anatomical features.

\subsection{Methodology of BrainFuse}
\label{subsec:methodology-brainfuse}

The development of the BrainFuse methodology is grounded in two complementary steps. First, a self-supervised learning approach is used to train a frame interpolation model. This model learns to generate a missing slice given two surrounding slices within a single MRI volume. After it becomes proficient at local slice interpolation, it generalizes to more demanding tasks, such as synthesizing transition slices between distinct brains. Second, the method exploits the trained interpolation model to fuse entire 3D volumes along one or multiple axes, combining different portions of two or more brains.

In practical terms, suppose each brain MRI is resampled to a uniform grid of dimensions (for example, $64 \times 64 \times 64$), forming a sequence of slices along the depth axis. A trained model $g(\cdot; \theta)$ takes as input two slices at positions $i$ and $j$ and generates one or more intermediary slices, thereby modeling what should appear between those input slices. Early in training, the distance between slices $i$ and $j$ is kept small so the model can learn local continuity and anatomical flow. As training proceeds, the distance increases, leading the model to handle progressively dissimilar slices. This gradual exposure, often referred to as a curriculum learning strategy \citep{bengio2009curriculum}, helps the model adapt to larger discrepancies, which is essential when eventually fusing slices from different subjects.

When combining two different scans, the method chooses a portion of one volume (for example, the first 20 slices) and a portion of another volume (the remaining slices). The interpolation network then synthesizes a transition region to smoothly connect these portions. If the intention is to preserve the label—for instance, brain age—it is possible to approximate the fused scan's label by linearly interpolating between the original labels of each volume, typically weighting them according to the ratio of the combined portions. In some cases, no single clear label can be assigned if the fusion involves many volumes or exhibits significant anatomical differences. Such unlabeled fusions can still serve a valuable function, including use in self-supervised pretraining methods (e.g., contrastive or consistency-based approaches \citep{henaff2020data, lee2013pseudo}).

\begin{figure}
  \centering
  \begin{subfigure}{0.3\linewidth}
    \centering
    \includegraphics[width=.85\linewidth]{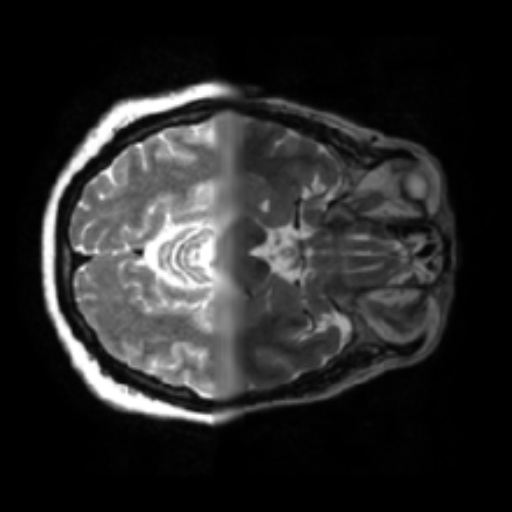}
    \caption{Generated brain by fusing two patients.}
    \label{fig:two_brain}
  \end{subfigure}
  \hfill
  \begin{subfigure}{0.3\linewidth}
  \centering
    \includegraphics[width=.85\linewidth]{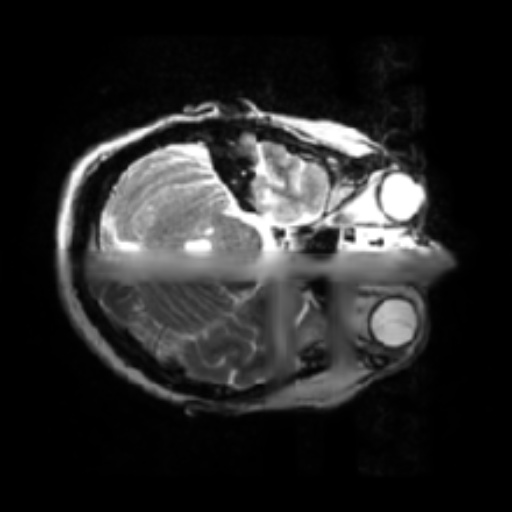}
    \caption{Generated brain by fusing eight patients.}
    \label{fig:eight_brain}
  \end{subfigure}
  \hfill
  \begin{subfigure}{0.3\linewidth}
     \centering
    \includegraphics[width=.85\linewidth]{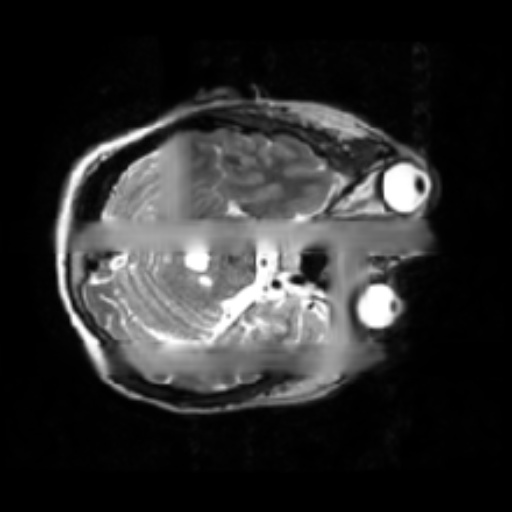}
    \caption{Deformed brain obtained from fusing six patients.}
    \label{fig:desfigured_brain}
  \end{subfigure}
  \caption{Example of slices from BrainFuse generated brains.}
\end{figure}

Although conceptually straightforward, creating fused volumes requires careful bookkeeping to avoid excessive noise accumulation. When iterated many times, each newly generated scan may deviate from valid anatomical structures, especially if the source volumes differ substantially in shape or size. Accordingly, BrainFuse typically sets practical limits on the number of times a newly generated volume can be reused for further fusion, you can see figures \ref{fig:two_brain},\ref{fig:eight_brain},\ref{fig:desfigured_brain} to see examples of fusing two, six and eight patients. It can be observed that while we lose some fine details in the fusing locations, the new samples are mostly consistent, even in the case of the eight sample fusion. An observed failure case is when the discrepancies are too great that simply filling in the missing frames isn't sufficient to correctly align the different parts. In parallel, it can also filter out obviously malformed images to maintain dataset quality.

\subsection{Application in Brain Age Prediction}
\label{subsec:brainfuse-brain-age}

BrainFuse has been tested in the context of brain age prediction on publicly available datasets such as the IXI dataset \citep{peng2021accurate}, which contains T1-, T2-, and PD-weighted MRI scans from healthy individuals. The motivation for focusing on age prediction is its clinical value: misalignments between chronological age and predicted age can help identify accelerated aging or early signs of neurodegenerative conditions.

In practice, each scan is preprocessed to a standard orientation, clipped intensity range, and uniform resolution. Three-channel volumes are formed by stacking different MRI modalities (T1, T2, and PD). A residual neural network, adapted for 3D convolutions \citep{he2016deep}, is then trained to classify age by dividing the continuous variable into bins and often smoothing the bin labels to mitigate harsh boundaries.

During experiments, the fused datasets are used to supplement the original training set. Fused samples that result from combining two brains with known ages can be assigned approximate ages via linear interpolation. Such label-preserving fusions increase the variability in brain structure, while still providing the training algorithm with coherent labels. Another category of merged scans, which fuse multiple brains extensively or where structural differences become too great, may have no direct label and can be leveraged in self-supervised or semi-supervised learning. Methods such as Barlow Twins \citep{zbontar2021barlow} can be trained on these unlabeled volumes to acquire more robust representations. Subsequently, the downstream predictor can be fine-tuned on labeled data, benefiting from representations formed over a more diverse dataset.

Empirical findings suggest that integrating BrainFuse samples frequently raises accuracy metrics. Baseline models that train only on the original dataset tend to plateau sooner, whereas models augmented with fused data appear more robust, particularly if combined with consistency regularization \citep{lee2013pseudo}. This type of regularization enforces similar predictions under different augmentations, strengthening generalization. When tested on a held-out portion of the IXI dataset, BrainFuse-based approaches typically yield improved top-$k$ accuracy for age classification, reflecting a deeper resilience to the limited size and variability of standard medical imaging collections.

\subsection{Limitations and Future Prospects}
\label{subsec:brainfuse-limitations}

Although BrainFuse shows promise for improving performance in settings constrained by data availability, it also has certain drawbacks that merit attention. One challenge relates to computational requirements, since training the interpolation network on numerous slices under a curriculum can be time-consuming. Generating additional datasets of fused scans can also demand significant storage if not performed on the fly. In situations where multiple fusions of the same partial scans are generated, the risk of propagating synthetic artifacts grows unless carefully monitored.

A further consideration is that some fusions may be anatomically incompatible. Pronounced differences in head orientation, brain volume, or pathology can make the transitions learned by the interpolation network incomplete. If the model attempts to join mismatched structures, severe distortions can arise that do not resemble realistic anatomy. While discarding malformed outputs is an option, doing so reduces the efficiency of data generation.

Future developments might include improved selection criteria for deciding which scans to fuse based on similarity metrics or auxiliary anatomical information. Adaptive methods could be designed to detect and discard problematic transitions automatically, while generative adversarial or diffusion-based refinement could correct localized artifacts. Additionally, BrainFuse could be extended to support more specialized tasks, such as lesion segmentation, by fusing scans that contain lesions of different sizes, thereby increasing dataset diversity in a controlled manner.

In summary, BrainFuse offers an approach to synthesizing new, more varied brain MRI scans by leveraging a frame interpolation model initially trained in a self-supervised manner. It appears to be particularly effective when integrated with robust representation learning and semi-supervised strategies. While it does not eliminate the need for real, high-quality data, it can be a valuable augmentation resource in tasks like brain age prediction where data scarcity and variability are leading concerns.

\section{Revert Project}
\label{sec:revert-project}

\subsection{Problem Description}
\label{subsec:revert-problem}

The REVERT initiative addresses the critical need for improved treatment outcome predictions in complex medical conditions such as metastatic colorectal cancer (mCRC). Traditional analytic approaches rely on limited datasets or single algorithms, often hampering personalized therapy decisions. A principal challenge is the heterogeneity of the available medical information, which includes gene-level mutations, imaging data, and assorted clinical features. Consequently, the project proposes a multi-stage pipeline that draws on machine learning models and advanced data processing to stratify patients and guide oncologists toward selecting the most suitable interventions.

Within this context, Work Package~3 (WP3) of REVERT focuses on \emph{ICT methods and bioinformatics for clinical outcome prediction}, specifying tasks aimed at developing robust models and creating a web application recommending most suitable treatment option. The relevant tasks include:
\begin{itemize}
    \item \textbf{Task~3.1 (Predictive Models for Stratifying Populations):} surveying and defining state-of-the-art machine learning (ML) methods, including interpretability techniques for medical applications.
    \item \textbf{Task~3.2 (Implementation of Predictive Models):} experimenting with various ML algorithms using structured patient data, such as that of Metastatic Colorectal Cancer (MSKCC) from the Cancer Cell repository.
    \item \textbf{Task~3.3 (Testing Predictive Models):} thoroughly validating classification performance to select optimal algorithms.
    \item \textbf{Task~3.4 (Predictive Model Web App):} integrating top-performing models in a web system for convenient data collection and on-demand predictions in a client--server architecture.
    \item \textbf{Task~3.5 (Real-World Testing and Optimization):} deploying a multi-model ensemble approach to enhance reliability, including AUC-based filtering and weighted ensembling of CatBoost models.
\end{itemize}

\subsection{Proposed Solution}
\label{subsec:revert-solution}

The solution hinges on two main pillars: (\textit{i})~a core set of ML algorithms rigorously trained on real-world colorectal cancer data, including patient demographics, metastatic sites, and treatment histories, and (\textit{ii})~a lightweight web application that delivers these models’ predictions directly to healthcare professionals. Key details of each workflow phase are summarized below.

\paragraph{Data Gathering and Preprocessing.}
In line with standard ML practices \citep{mitchell1997machine}, the research team obtained data from the “Metastatic Colorectal Cancer (MSKCC, Cancer Cell 2018)” dataset \citep{volovat2022use}, covering 1134 patients aged 20--80. These data encompassed genetic, histologic, and treatment exposures:
\begin{itemize}
    \item \emph{Feature Engineering:} Non-numerical factors (e.g., \textit{first\_site\_of\_metastasis}) were translated into numeric form through dictionary-based encodings or array-of-integers representations.
    \item \emph{Cleanup Steps:} Negative or corrupted entries were discarded, ensuring consistency with baseline assumptions of the underlying ML techniques.
\end{itemize}
The label of interest was the \textit{living\_status} (0 or 1), indicating whether the patient is deceased or alive.

\paragraph{Predictive Models.}
Six classification models were initially prototyped:
\begin{enumerate}
    \item \textbf{\acro{NB}[Naïve Bayes]}: A Bayesian approach with additive smoothing set to 1.0. Although it provides strong interpretability, results showed moderate accuracy \citep{mitchell1997machine, serban2013combining}.
    \item \textbf{Random Forest (RF):} An ensemble of decision trees \citep{breiman2001random}, tested with parameters such as \texttt{maxDepth=5} and \texttt{numTrees=20}, yielding near-perfect performance on this dataset.
    \item \textbf{Decision Tree (DT):} Single-tree classification \citep{quinlan1986induction} surprisingly achieved $100\%$ accuracy on the test portion, suggesting that features are highly separable for this cohort.
    \item \textbf{Gradient Boosted Trees (GBT):} Built on the idea of stacked generalization \citep{wolpert1992stacked} with a logistic loss and default parameters. This approach also achieved $100\%$ accuracy, confirming strong feature significance.
    \item \textbf{Logistic Regression (LR):} A linear model for binary classification \citep{tsiatis1980note}, reaching about $97\%$ accuracy, close to the best results.
    \item \textbf{Support Vector Machine (SVM):} Using default PySpark parameters \citep{cusmuliuc2018identifying}, the SVM also delivered $100\%$ accuracy on test data, matching the top results.
\end{enumerate}
While perfect or near-perfect results might reflect strong data separability, they also raise concerns about overfitting or data peculiarities. Further real-world validation remains essential.

\paragraph{Web Application and Deployment.}
To enable practical usage in clinical scenarios, the best-performing models (chiefly Decision Trees or Gradient Boosted Trees) were integrated into a web-based platform. This client--server system is divided into:
\begin{itemize}
    \item \emph{Web Front-End:} Collects patient data (age, sex, metastases, tumor stage) with from uploaded spreadsheets.
    \item \emph{Server Components:} 
    \begin{enumerate}
    \item A Python-based inference server that routes requests over HTTPS and manages authentication/authorization.
    \item A Python back-end module hosting the ML model, providing responses for predicted ranking the different treatment options.
    \end{enumerate}
\end{itemize}
Data confidentiality is safeguarded by encrypting database records and employing SSL for data transmission.

\paragraph{Ensemble Optimization and Real-World Testing.}
Subsequent efforts under Task~3.5 introduced CatBoost ensembles \citep{volovat2022use} to accommodate the complexities of mCRC. This included:
\begin{enumerate}
    \item \emph{Grid Search Over 20,000+ Parameter Sets:} Various learning rates, tree depths, sampling methods, and regularization values.
    \item \emph{AUC-Based Filtering:} Only models above a certain AUC threshold were kept. Approximately 235 well-performing models remained.
    \item \emph{Weighted Ensemble:} Combining top models proportionally to their AUC yields robust, consistent predictions. 
\end{enumerate}
Preliminary real-world tests emphasize the need for continuous updates: as additional patient data accumulates, the model can be retrained, thus refining classification power for even more personalized therapy recommendations.

\subsection{Results}
\label{subsec:revert-results}

The REVERT pipeline was tested within a pilot clinical study enrolling 111 metastatic colorectal cancer patients across five Medical Oncology Units. Gene expression profiling, actionable mutation screening, and circulating biomarker analyses were integrated with each patient’s clinical history, and the combined information was used by an AI-based platform to suggest optimal first-line treatment regimens. In most instances, the oncologists’ final treatment decisions aligned with the system’s recommendations, with non-concordant cases frequently related to clinical concerns regarding reduced dihydropyrimidine dehydrogenase (DPD) activity, which predisposes patients to toxicity from fluoropyrimidine-based therapy. Preliminary observations at the six-month follow-up point revealed a progression-free survival (PFS) of approximately 13.5 months for patients treated according to the AI-recommended regimens, compared to a historical PFS of around 10.2 months for a similar patient population. These findings suggest a potentially favorable impact of the REVERT approach on patient outcomes, though larger-scale prospective studies are needed to validate its generalizability. One participating center was not able to secure formal authorization to administer treatments based on the REVERT framework but did test the proposed drug combinations on cell line-derived organoids, further exploring the platform’s applicability in a preclinical setting.

From an implementation standpoint, the system integrates classical and ensemble machine learning models, including Decision Trees, Random Forests, and Gradient Boosted methods such as CatBoost, which exhibited high predictive performance following extensive hyperparameter tuning. An Android-based application ensures rapid and interpretable inference, allowing oncologists to query patient-specific survival probabilities or recommended therapeutic choices at the point of care. In addition, data management is handled through secure storage, retrieval, and encryption mechanisms that comply with legal requirements, thereby enabling privacy-conscious create-read-update-delete (CRUD) operations on patient records. As the project progresses, efforts are focused on expanding patient enrollment in ongoing clinical trials, increasing the model’s interpretability through methods such as SHAP or LIME, and extending the REVERT framework to accommodate a broader range of clinical scenarios. This multi-pronged strategy seeks to mitigate potential overfitting and establish a robust foundation for future clinical integration.

\section{Backforward Propagation}
\label{chap:backforward}
This work \cite{stoica2023backforward} experimented with a different approach to solving the problem of Internal Covariate Shift.

\subsection{Methodology}
\label{sec:methodology-backforward}

\subsubsection{Internal Covariate Shift}
\label{subsec:ics}
Neural networks frequently encounter what is termed \emph{Internal Covariate Shift} (ICS), a phenomenon wherein the distribution of activations in deeper layers changes after updates to earlier layers within a single training iteration. The term ``covariate shift'' was originally discussed in \citet{shimodaira2000improving}, and the adaptation of this idea to internal layers of neural networks is attributed to \citet{ioffe2015batch}. In essence, ICS can be viewed as a moving target: by the time the backpropagated gradients reach later layers, the relevant inputs and activations used to compute those gradients may already have changed due to updates in previous layers.

In traditional backpropagation, a forward pass is conducted once per mini-batch, followed by a backward pass that calculates gradients for every layer and updates them simultaneously. This process implies that the updates for the deeper layers are based on \emph{stale} information from prior layers, leading to a shift in the effective input distribution seen by those deeper layers. As a result, training typically resorts to smaller learning rates and additional normalization techniques to maintain stability.

\subsubsection{Backforward Propagation}
\label{subsec:backforward}

To eliminate ICS entirely, the technique known as \textbf{Backforward Propagation} proceeds by recalculating gradients layer-by-layer, ensuring that each layer's update fully accounts for the most recently updated parameters of the preceding layer(s). The key steps are as follows:

\begin{enumerate}
    \item \textbf{Layer-by-Layer Update:} Begin with the first layer. Perform a forward pass through that layer to compute its output, then carry out the backward pass specific to it. Once its weights are updated, move on to the second layer, re-doing the forward pass (now using the updated first-layer weights) before calculating and applying the second layer's gradients.  
    \item \textbf{Repeated Forward-Backward Computations:} Continue in this sequential manner until all layers have been updated. By the time deeper layers receive gradients, they reflect the latest distribution of activations from the earlier layers.  
    \item \textbf{Computational Overhead:} The core drawback is the increased computational expense. Rather than a single forward and backward pass per batch, multiple passes are required, one for each layer. Although partial recalculations or parallelization may mitigate this cost, such optimizations often demand custom code or specialized frameworks.
\end{enumerate}

Intuitively, this approach removes ICS because each layer sees a ``frozen'' version of its previous layer only for the duration of its own gradient computation. Hence, the input distribution for that layer does not shift unexpectedly when its own parameters are updated.

\subsection{Experimentation and Results}
\label{sec:experiments}

We evaluate Backforward Propagation on the MNIST dataset \citep{lecun1998gradient}, which includes images of handwritten digits from 0 to 9. A simple multi-layer perceptron (MLP) is used to highlight ICS effects in a controlled environment. The network structure is:

\begin{itemize}
    \item One hidden layer with 256 neurons (ReLU activations).
    \item An output layer with 10 units (softmax activation).
\end{itemize}

All experiments use a basic stochastic gradient descent (SGD) optimizer without momentum or additional regularization. No data augmentation is performed to isolate ICS as the primary variable of interest. We compare:

\begin{itemize}
    \item \textbf{Baseline (Standard Backpropagation):} A single forward pass followed by a single backpropagation pass for every mini-batch.
    \item \textbf{Backforward Propagation (Proposed Method):} Multiple forward-backward cycles per batch, updating each layer in turn.
\end{itemize}

Different learning rates are tested, including $0.1$, $0.05$, and $0.01$, to observe how ICS removal interacts with varying step sizes.

\subsection{Key Observations and Analysis}
\label{subsec:analysis}

\begin{enumerate}
    \item \textbf{Rapid Fitting with High Capacity:}  
    In principle, if the hidden layer is sufficiently large and the learning rate is high, one might expect to fit a small batch perfectly in a single update. However, ICS complicates this outcome in standard backpropagation since deeper layers assume the pre-update parameters of earlier layers. With Backforward Propagation, such fitting is more consistent because the gradient calculations are fully up-to-date.

    \item \textbf{Regularizing Effect of ICS:}  
    Interestingly, experiments show that when ICS is eliminated, weight norms can grow more quickly, suggesting that ICS acts as a mild regularizer. By preventing large changes in deeper layers, ICS naturally damps weight growth. Removing ICS thus requires careful management of learning rates to avoid instability or overfitting.

    \item \textbf{Convergence Behavior:}  
    At moderate learning rates (e.g., 0.01--0.05), Backforward Propagation can converge faster and sometimes achieve higher validation accuracy. However, for higher learning rates (e.g., 0.1), this method can be more prone to local plateaus. One hypothesis is that standard backpropagation occasionally benefits from slight misalignment of the gradients, which can help escape shallow minima.
\end{enumerate}

\begin{figure}[htbp]
    \centering
    \includegraphics[width=0.6\textwidth]{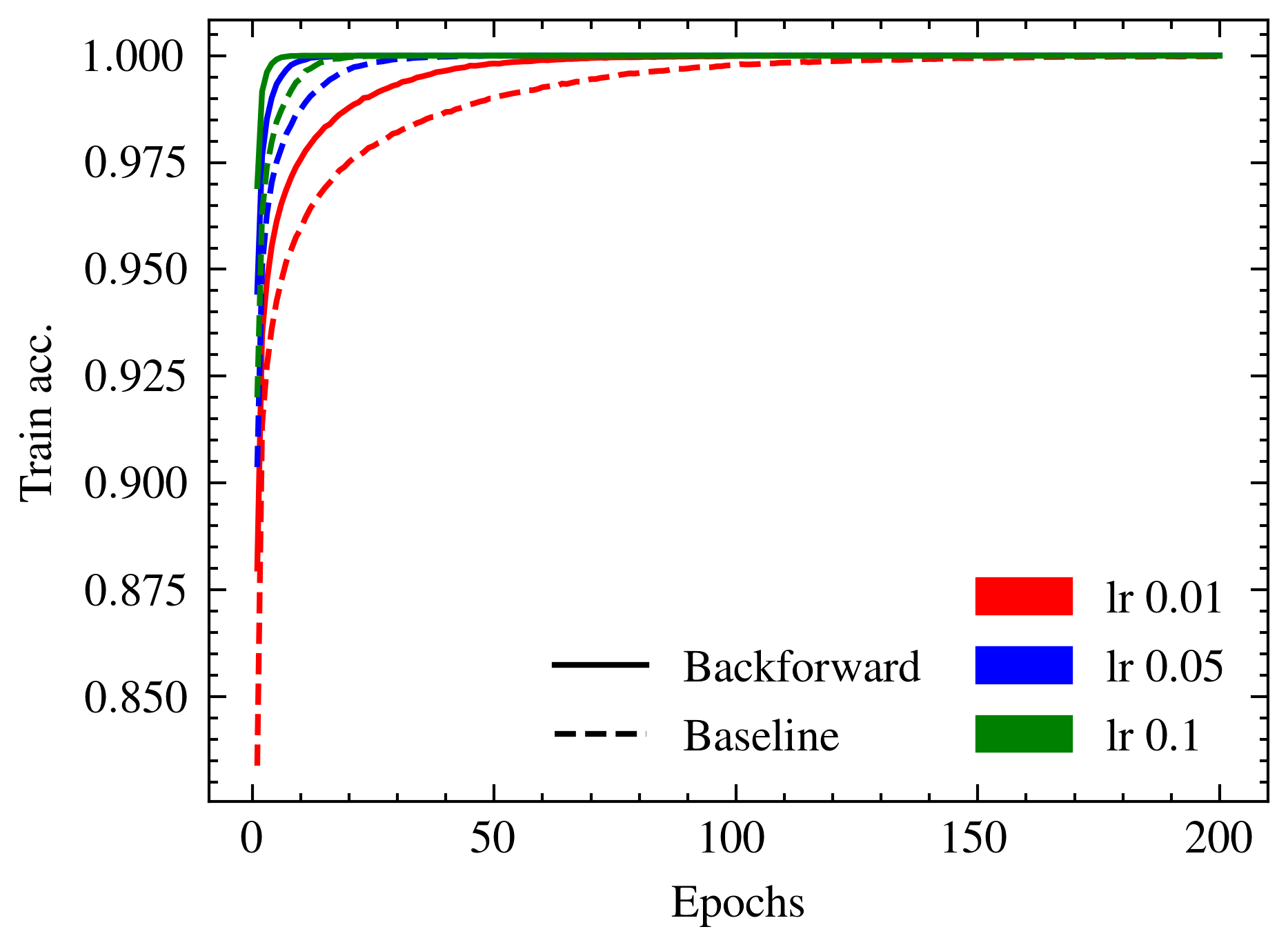}
    \includegraphics[width=0.6\textwidth]{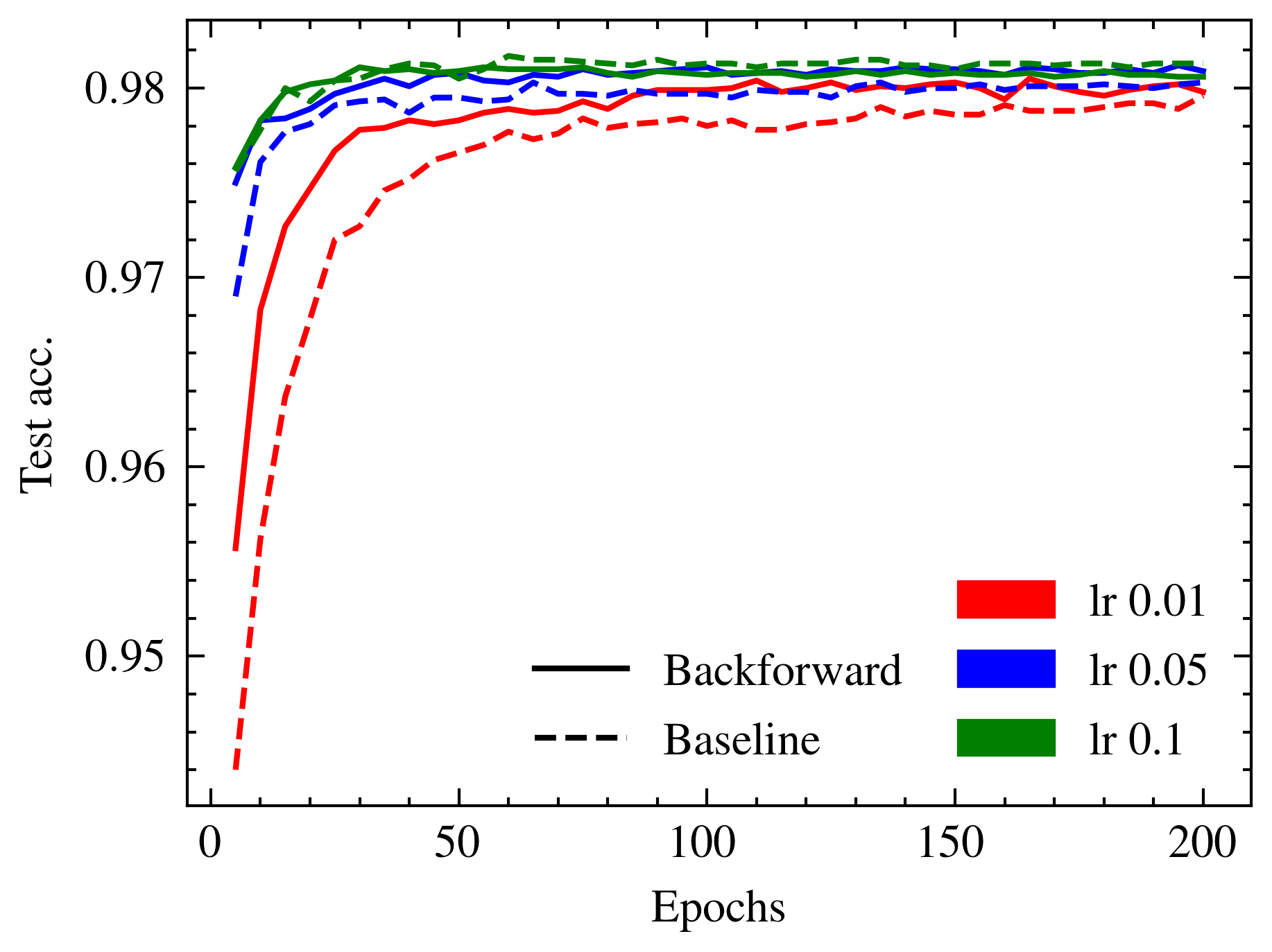}
    \caption{Training vs.\ Validation Accuracy for Backforward Propagation vs.\ Standard Backpropagation on MNIST, across different learning rates.}
    \label{fig:bf-curve}
\end{figure}

For a visual illustration, Figure~\ref{fig:bf-curve} plots the training and validation accuracy over epochs. One sees a generally faster convergence rate for Backforward Propagation at moderate learning rates, while the baseline can sometimes exhibit beneficial noise at higher rates.

\subsection{Conclusion and Future Work}
\label{sec:conclusion}

Backforward Propagation directly eliminates Internal Covariate Shift by refreshing the forward pass and gradient calculation after each layer update. Through experimentation with a simple MLP on MNIST, it is evident that removing ICS can accelerate convergence and boost performance under specific conditions. It also underscores ICS’s unintended role as a form of implicit regularization that tempers overly large updates.

Nonetheless, the computational cost of multiple forward-backward cycles per batch can be prohibitive. Future research could explore:

\begin{itemize}
    \item \textbf{Deeper Network Architectures}: Investigate whether ICS removal yields more pronounced gains in very deep or complex networks.
    \item \textbf{Selective Application}: Apply the approach intermittently, such as during network warmup or final fine-tuning, to reduce overall training cost.
    \item \textbf{Integration with Normalization Layers}: Compare or combine Backforward Propagation with normalization techniques (e.g., BatchNorm \citep{ioffe2015batch}, NormProp \citep{arpit2016normalization}) to examine potential synergy or redundancy.
    \item \textbf{Optimized Implementations}: Develop partial-recomputation strategies or specialized frameworks that reduce the repeated computation burden.
\end{itemize}

By precisely addressing the root cause of ICS, Backforward Propagation highlights new possibilities for gradient-based optimization techniques. Although real-world applicability may hinge on further refinement and computational optimization, the findings discussed here provide a solid foundation for both theoretical and practical exploration.

\subsection{Lung Analysis for Tuberculosis Classification}
\label{subsec:tb-classification}

Accurate automated analysis of lung CT scans for tuberculosis (TB) classification stands as a vital step in controlling the spread and impact of this disease. Despite notable improvements in tuberculosis therapies, drug resistance and delayed diagnosis continue to pose formidable obstacles. This subsection describes a framework that processes volumetric lung CT scans—both via direct 3D convolutional networks and projection-based 2D approaches—to classify pulmonary TB into distinct subtypes (Infiltrative, Focal, Tuberculoma, Miliary, Fibro-cavernous). It specifically builds on the work introduced as part of the ImageCLEFmedical 2021 challenge, in which participants were tasked with differentiating TB types through supervised learning approaches, our proposed approach \cite{hanganu2021uaic} is described below.

\paragraph{Context and Motivation.}
Tuberculosis remains one of the world’s leading causes of death from infectious diseases, claiming an estimated 1.6 million lives per year, according to the World Health Organization. Computer-aided detection and classification methods can reduce diagnostic time and improve screening reliability. Recent competitions, such as the ImageCLEFmedical TB tasks \citep{hanganu2021uaic}, have focused on identifying lesion types directly from CT volumes. Although chest X-rays are simpler to obtain, CT scans can offer a higher resolution view of lung structures and TB lesions, thus presenting an opportunity for advanced image processing and deep learning to distinguish between various tuberculosis manifestations.

\paragraph{Related Work.}
Earlier editions of the TB classification task have yielded multiple strategies:
\begin{itemize}
    \item \textbf{Volumetric CNNs:} Some teams attempt direct 3D classification models \citep{liauchuk2018deep, kozlovski2020tuberculosis}, albeit high memory consumption and smaller batch sizes hamper stable training.
    \item \textbf{Projection-based 2D CNNs:} Several groups convert 3D volumes into 2D images via maximum/average intensity projection along different axes \citep{liauchuk2019cnn, miron2020senticlab}. This dimensionality reduction can significantly mitigate GPU memory constraints.
    \item \textbf{Hybrid Approaches:} Combining 2D slices with volumetric cues or ensemble strategies have also shown promise \citep{coca2020cnn}.
\end{itemize}
Nevertheless, handling large CT volumes in limited hardware conditions remains a consistent theme. Data augmentations, careful normalization, and pre-trained model adaptations are frequently employed to manage overfitting and training instability.

\paragraph{Dataset Characteristics and Preprocessing.}
For the ImageCLEFmedical 2021 TB challenge, the provided dataset included chest CT scans with five categories: \emph{Infiltrative}, \emph{Focal}, \emph{Tuberculoma}, \emph{Miliary}, and \emph{Fibro-cavernous} \citep{hanganu2021uaic}. Masks delineating lung regions were also supplied, although each had limitations. To address segmentation inaccuracies:
\begin{itemize}
    \item \textbf{Mask Combination}: Two different automated lung masks were merged to cover potential missed areas in severe TB lesions or erroneous lung boundaries.
    \item \textbf{Outlier Removals and Corrections}: Faulty scans with corrupt headers or malformed structures were excluded or manually labeled if reparable. Some mask files incorrectly included the trachea, necessitating manual fixes.
\end{itemize}
Before training, images were normalized (thresholded to \([-1200,600]\) HU), then re-scaled to a consistent voxel size or reduced to 2D through projection to limit hardware usage.

\paragraph{3D Volumetric Classification.}
To preserve spatial correlations, the team experimented with a MedicalNet pre-trained 3D-ResNet variant \citep{chen2019med3d}. This network was originally trained on multiple medical datasets. However, due to memory constraints (a single GPU with limited VRAM), images had to be downsampled to \(256\times256\times32\). Gradient accumulation simulated larger effective batch sizes, but the models often encountered overfitting or exploding gradients beyond 40 epochs of training. The best 3D approach (MedicalNet10) was fine-tuned for 44 epochs using AdamP \citep{heo2021adamp} and cross-entropy loss, producing moderate performance but indicating the need for more stable regularization or data augmentation strategies.

\begin{figure}[ht]
    \centering
    \includegraphics[width=0.3\textwidth]{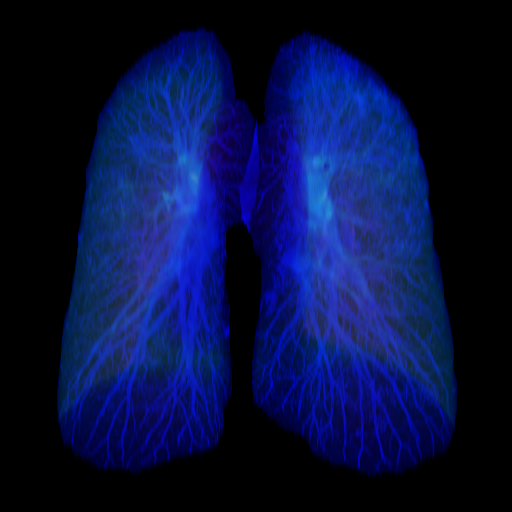}
    \includegraphics[width=0.3\textwidth]{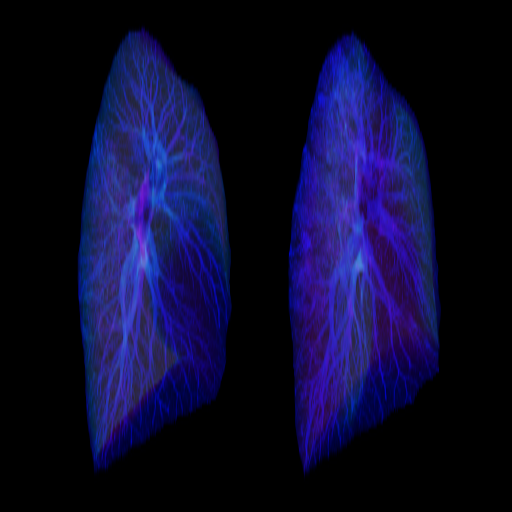}
    \includegraphics[width=0.3\textwidth]{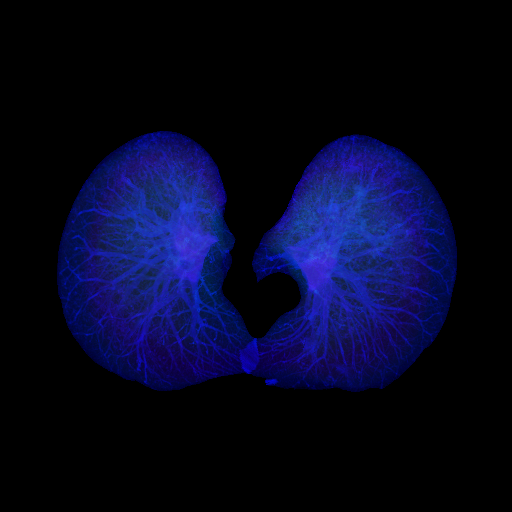}
    \caption{X, Y and Z Axis Projections Approximations.} \label{fig:tuberculosis2021_1}
\end{figure}

\begin{figure}[ht]
    \centering
    \includegraphics[width=0.45\textwidth]{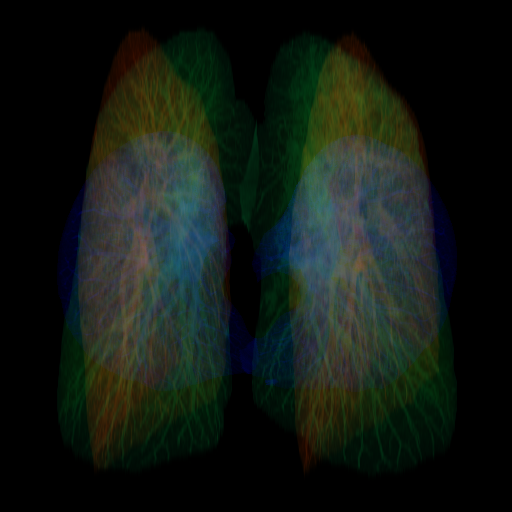}
    \caption{Stacked Projection Approximation} \label{fig:tuberculosis2021_2}
\end{figure}

\paragraph{2D Projection Classification.}
Inspired by prior successes, the group also explored a \emph{projection-based} pipeline:
\begin{enumerate}
    \item \textbf{Projection Generation}: Slices were aggregated along the X, Y, and Z axes using maximum, standard deviation, and a modified \textit{k-means}-based approach. This yields multi-channel 2D images (e.g., 7 channels per axis, see Figure \ref{fig:tuberculosis2021_1}).
    \item \textbf{Model Architecture}: PreResNet56 \citep{he2016deep} was trained from scratch on these projections. The combined channels for all three axes (X, Y, Z) can stack to 21 channels, effectively capturing volumetric traits in a single 2D representation, see Figure \ref{fig:tuberculosis2021_2}.
    \item \textbf{Training Setup}: RAdam \citep{liu2019variance} and Lookahead \citep{zhang2019lookahead} optimizers were employed. Weak augmentations (flips, rotations) avoided instability but led to quick overfitting. Stronger augmentation methods (Randaugment, Random Erasing) sometimes triggered exploding gradients. 
\end{enumerate}

Ultimately, two PreResNet56 models achieved the highest local validation scores, though final official results indicated room for improvement in stability and generalization.

\paragraph{Results and Observations.}
Submissions to the TB challenge yielded variable \emph{kappa} scores. Although the volumetric approach pre-trained on multiple medical tasks initially showed potential, final performance was hampered by resource limitations and overfitting. The 2D projection approach also encountered similar pitfalls, indicating that more controlled regularization strategies or advanced augmentations (e.g., Mixup \citep{zhang2017mixup}, CutMix \citep{yun2019cutmix}) require further fine-tuning to avoid training collapse.

\paragraph{Future Directions.}
Several areas could enhance TB classification:
\begin{itemize}
    \item \textbf{Stable Augmentation Strategies}: Developing specialized data augmentations for medical CT volumes—ones that do not trigger catastrophic gradient behaviors—remains a priority.
    \item \textbf{Additional Normalization and Pooling Layers}: Revisiting normalization steps or using deeper pooling before the classification head may mitigate exploding gradients, especially in volumetric models.
    \item \textbf{Hardware Upgrades}: Larger VRAM could allow training with bigger batch sizes or more sophisticated 3D architectures without resorting to heavy downsampling.
\end{itemize}
Addressing these challenges can significantly improve automated TB type detection, ultimately accelerating and standardizing the clinical workflow in diagnosing complex forms of tuberculosis.


\section{Domain-Specific Contributions}
\label{sec:domain_specific_contributions}

Although the principal emphasis of this dissertation lies in advancing deep learning methods within medical imaging, the breadth of related research undertaken during the doctoral period extends into other application domains as well. These additional endeavors, while not central to the primary thesis theme, illustrate both the versatility of contemporary artificial intelligence techniques and the practical benefits that can arise from adapting them to specialized problems. Each of the following subsections addresses a distinct domain-specific challenge, describing how artificial intelligence and deep learning frameworks were leveraged to make meaningful contributions to disparate fields.

Collectively, these projects demonstrate a capacity to transfer insights and methodologies developed for healthcare applications into broader contexts, including real-time stroke detection on mobile devices, enhanced travel-time forecasting in sparse data conditions, and predictive analysis for urban development, bird song classification, and social media post impact assessment. In many instances, the challenges encountered—such as limited data availability, the need for domain-specific feature engineering, or the demand for efficient deployments—parallel those in medical imaging, thereby reinforcing the overarching lessons of model adaptation and careful data handling. By detailing these additional research efforts, this section provides an expanded perspective on how the underlying techniques of deep learning, computer vision, and natural language processing can cross traditional boundaries to address a wide spectrum of contemporary issues.

\subsection{Prehospital Cerebrovascular Accident Detection Using Artificial Intelligence-Powered Mobile Devices}
\label{subsec:prehospital-cva-detection}

Cerebrovascular accidents (CVAs), commonly referred to as strokes, represent a significant global health concern. They rank among the leading causes of mortality and long-term disability worldwide. These realities highlight the importance of prompt identification and treatment. The core idea is that the timely recognition of stroke symptoms and the immediate initiation of clinical care can substantially improve patient outcomes, especially given that certain therapeutic interventions are only effective if administered in the initial hours following symptom onset \cite{CVATreatment}.

This section discusses the development of a mobile application designed to assist the general public in recognizing potential stroke events (\cite{simionescu2020prehospital}). The primary motivation behind this application is to bridge the gap between stroke onset and the decision to seek urgent medical care. While many regions provide protocols for ambulances and hospitals to streamline stroke treatment, these processes invariably depend on people’s ability to identify warning signs quickly and accurately. Surveys indicate that only a relatively small proportion of the general public is aware of all stroke warning indicators \cite{CDCStroke}, which often leads to delayed interventions.

\begin{figure}[t]
\centerline{\includegraphics[width=0.15\textwidth]{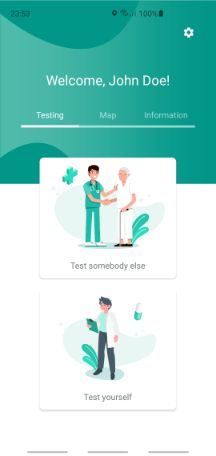}
\hspace*{2mm}\includegraphics[width=0.15\textwidth]{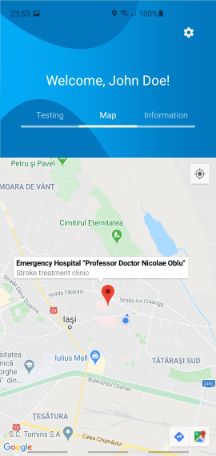} \hspace*{2mm}\includegraphics[width=0.15\textwidth]{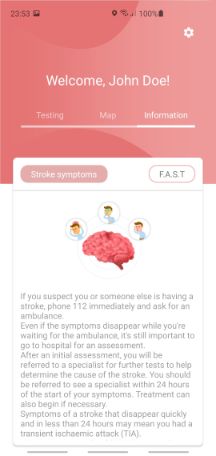}
\hspace*{2mm}\includegraphics[width=0.15\textwidth]{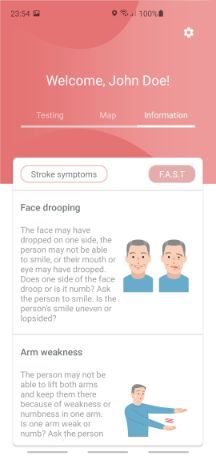}
\hspace*{2mm}\includegraphics[width=0.15\textwidth]{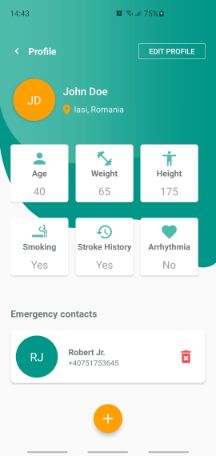}}
\caption{Testing Page; Nearby Clinics Page; Stroke Symptoms Page; F.A.S.T. Description Page; User Profile;}
\label{fig:application_design}
\end{figure}

\paragraph{Application Rationale and Design.}
The presented mobile application, named \emph{Stroke Help}, aims to guide users through a structured stroke evaluation by leveraging standard smartphone technologies, including cameras, microphones, motion sensors, and location services. The familiar F.A.S.T.\ framework (Face, Arms, Speech, Time) was adapted into the app to standardize the detection of key stroke symptoms. Specifically, the system incorporates automated checks of facial symmetry, arm strength or dexterity, and speech clarity. These measurements then feed into a wider computation of stroke risk based on the Japan Urgent Stroke Triage (JUST) score \cite{JUST}, a scoring method that differentiates the likelihood and potential subtype of stroke. See Figure \ref{fig:application_design} for an overview of the application design.

Users are prompted to complete a short calibration at installation. This calibration records baseline facial features, speech patterns, and physical dexterity. Such personalized reference data is essential in distinguishing acute neurological changes from a user’s regular facial asymmetry or speech nuances. The application also collects basic health history details that the JUST score requires, including age, any history of prior cerebrovascular incidents, and lifestyle factors such as smoking. Following these steps, the application is ready to assist in two main testing scenarios: (1) a \emph{self-test} for individuals who suspect they may be having a stroke and (2) a \emph{guided F.A.S.T.\ test} for evaluating someone else. After either test, the application computes a cumulative score and then indicates low, medium, or high stroke risk, suggesting the next immediate steps.

\paragraph{Facial Analysis for Symmetry.}
Facial drooping is often an observable sign in stroke cases, particularly when intracranial hemorrhage is implicated \cite{JUST}. The application’s face-analysis module uses a lightweight machine-learning approach to detect landmarks—like eye contours and mouth corners—within a live camera feed. After extracting these landmarks, the system measures asymmetry by comparing angles and distances. To maintain robust performance on mobile hardware, the algorithm applies a two-step mechanism: it first detects the approximate face region at a lower resolution, and then processes that region at a higher resolution to identify fine landmarks. The difference between the user’s baseline calibration and the new measurements is used to account for any personal facial characteristics, thus reducing false positives.

\paragraph{Arms and Dexterity Assessment.}
The arms test investigates whether a user exhibits unusual weakness or coordination issues, reflecting a potential stroke sign \cite{JUST}. This procedure is segmented into two checks. The first examines the user’s ability to hold the smartphone steadily at various angles, interpreting any difficulty in extending the arms as a possible weakness on one side. The second examines fine motor skills by having the individual trace a simple pattern on the touchscreen. Each side is tested separately, and results are compared to baseline measurements from the calibration stage.

\paragraph{Speech Evaluation.}
Speech disruption, known as aphasia, is another common stroke symptom \cite{JUST}. The mobile application prompts the user to repeat a short phrase. Leveraging a built-in speech recognition interface, it notes how closely the spoken words match an expected pattern, as well as the overall confidence of the recognition. A sharp drop in confidence from baseline may suggest that the user is struggling to articulate or pronounce words, possibly indicating a speech-related stroke deficit. This test is designed to handle short utterances, as extended voice inputs could be more difficult for distressed users.

\paragraph{Additional Symptom Checks and JUST Scoring.}
Beyond F.A.S.T.\ indicators, other warning signs can be relevant. The app allows users to optionally report supplementary symptoms, such as headaches or dizziness, through a simple checklist. All acquired information then feeds into the JUST score \cite{Jap, JUST}. This score provides a practical way to stratify stroke risk and offers a preliminary indication of stroke subtype. A higher final score prompts immediate guidance to call emergency services.

\paragraph{Integration with Notification and Mapping Services.}
If the resulting evaluation suggests a moderate or high probability of stroke, the application triggers automated notifications. These notifications can include contact details for the closest equipped stroke treatment centers, using the phone’s GPS data to locate suitable clinics. Additionally, the user can configure emergency contacts, who will receive text messages with time-stamped risk assessments and a link to the user’s location on Google Maps. These measures intend to ensure that essential information is available to the patient, bystanders, and medical personnel, ideally shortening the interval between onset and intervention.

\begin{figure}[b]
    \centerline{\includegraphics[width=0.16\textwidth]{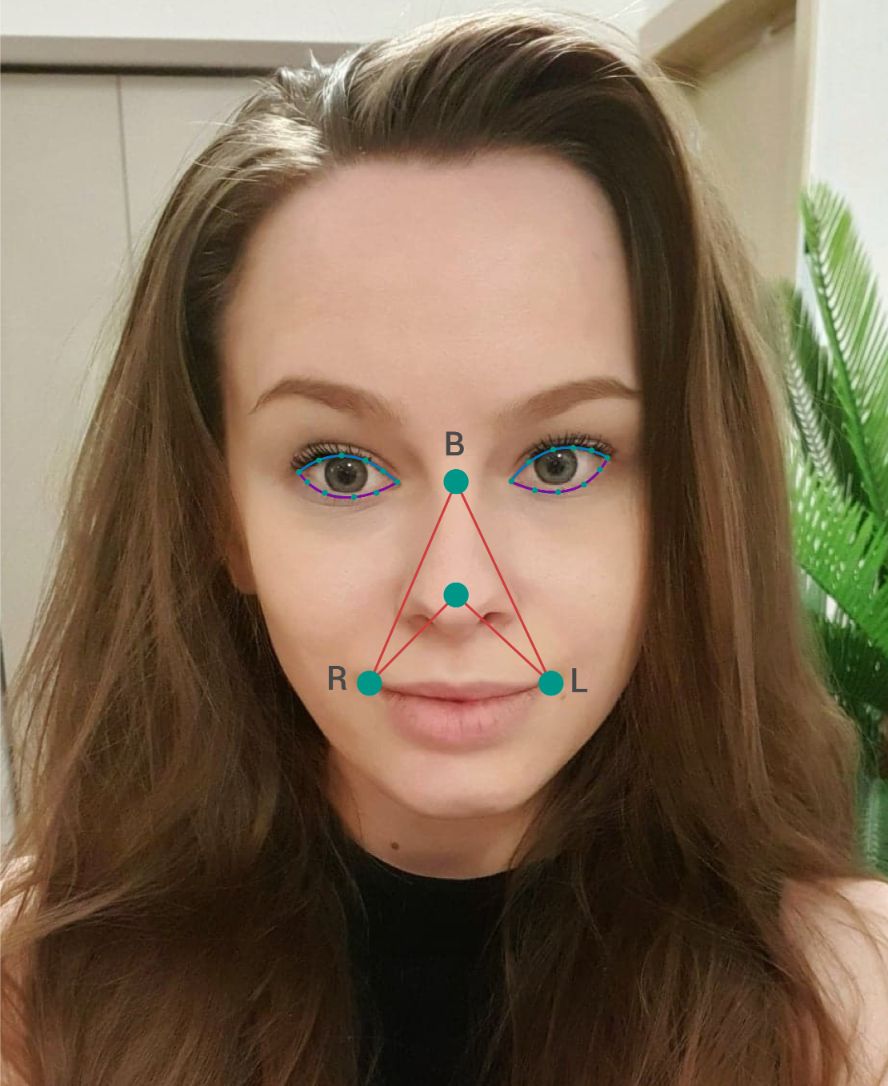}
    \hspace*{5mm}\includegraphics[width=0.16\textwidth]{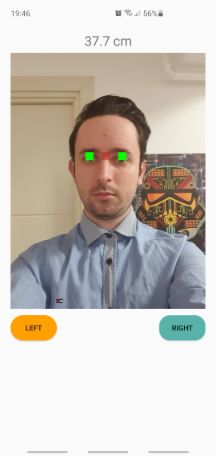}
    \hspace*{5mm}\includegraphics[width=0.16\textwidth]{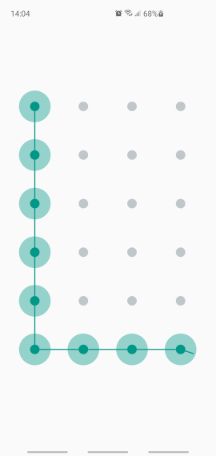}
    \hspace*{5mm}\includegraphics[width=0.16\textwidth]{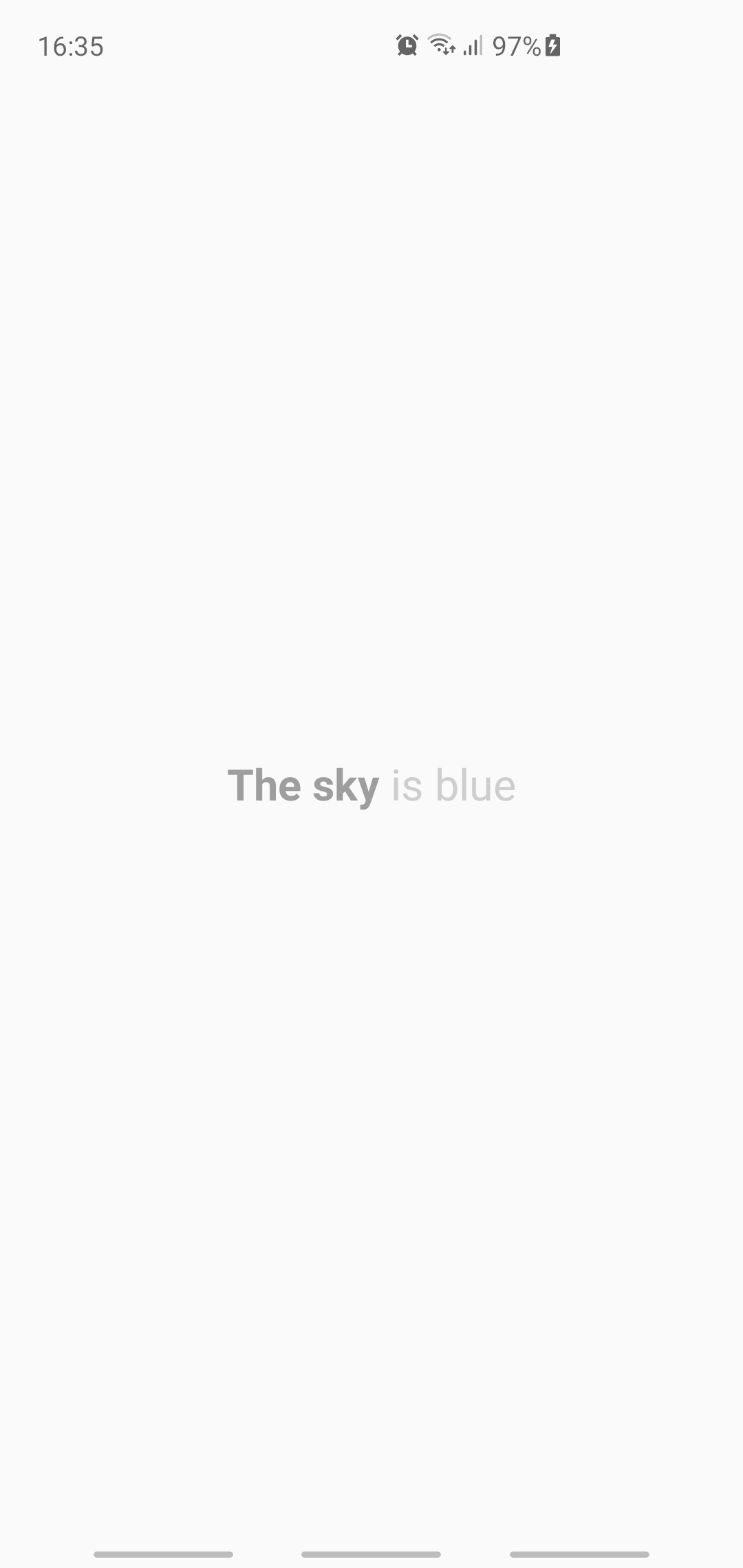}}
    \caption{Face Test; Strength Test, Dexterity Test; Speech Test;}
    \label{fig:stroke_tests}
\end{figure}

\paragraph{Evaluation, Usability, and Preliminary Findings.}
Initial efforts focused on usability, considering that stroke symptoms often compromise a user’s capacity to navigate complex digital interfaces. Adhering to recognized guidelines for mobile interface design and integrating concise steps has led to a system that requires minimal taps to complete the evaluation. A small-scale test using sample data (e.g., images or videos indicative of facial droop) helped refine the decision thresholds for asymmetry, see Figure \ref{fig:stroke_tests} for screenshots of the application during test taking. Potential inaccuracies do persist, particularly when lighting conditions are poor or background noise interferes with speech recognition. Nonetheless, early feedback suggests that the application’s layered testing approach, combined with calibrated baselines, provides a practical tool for laypersons to identify possible stroke events.

Future research is directed toward clinical partnerships. There is an intention to gather genuine patient data under proper consent and confidentiality measures, enabling a systematic validation of the approach. Currently, thresholds for classifying a failed test are somewhat heuristic, derived from limited image collections and sample analyses. Access to real-world instances would allow for more robust thresholds, improved accuracy, and a deeper understanding of how such an application might integrate into existing emergency medical systems.

\paragraph{Concluding Remarks on Prehospital CVA Detection.}
Prehospital cerebrovascular accident detection using artificial intelligence-powered mobile devices offers a novel perspective on first-response medical assessments. While mass public awareness campaigns remain important in reducing treatment delays, an accessible and guided tool can help individuals execute standardized tests more accurately. By adapting the F.A.S.T.\ method, adding machine-learning components for facial, arm, and speech analysis, and connecting users with medical resources, the system aims to improve stroke outcomes. Although broader clinical trials and refinements remain necessary, the initial design underscores the feasibility of smartphone-based stroke detection and the potential contributions this approach can bring to public health.

\subsection{Zonal Estimation Method for Offline Travel Time Prediction in Sparse Data Environments}
\label{subsec:zonal-estimation}

This section presents a structured method for estimating offline travel times under conditions where data are sparse and historical records are limited. While state-of-the-art systems typically rely on extensive continuous data streams and long-term historical profiles, many practical use cases involve newly instrumented urban areas or short-term deployments in which collecting large datasets may be difficult. The proposed approach, referred to here as a \textit{zonal estimation method}, is explicitly tailored to these low-data contexts. It combines a Gaussian Mixture Model (GMM) for urban zoning with hexagonal segmentation and modest temporal aggregation to stabilize speed estimates. The method aims to deliver useful travel time predictions when only brief data snapshots are available or when live sensor feeds are unavailable.

\paragraph{Context and Motivation.}
Accurate travel time predictions are integral to modern transportation systems, underpinning navigation tools, logistics optimization, and municipal planning. In large metropolitan areas with high sensor coverage, data-driven models can update continuously with near-real-time inputs. However, many mid-sized or smaller cities operate with limited or sporadic telemetry. Even in well-instrumented cities, newly constructed roads or irregular events can temporarily reduce data coverage. These gaps undermine the effectiveness of conventional techniques, which typically assume a wealth of training data spanning months or years. By focusing on minimal data requirements, the zonal estimation method addresses practical constraints faced during early-stage deployments or in resource-constrained environments.

\paragraph{Related Work.}
Past research in travel time estimation (TTE) has explored an array of data-driven solutions. Some rely on consensus-based approaches or combine multiple data sources, including floating-car data, GPS trajectories, and Bluetooth detections \citep{chiabaut2020trafficcongestion, gemma2024, zheng2023}. While such methods demonstrate strong performance in high-coverage scenarios, they often degrade significantly when confronted with limited historical records. Approaches incorporating Gaussian Mixture Models for traffic distributions have shown promise in capturing multimodal patterns \citep{chen2023}, but most examples involve substantially more data than available in the present context. Efforts to address sparse data explicitly \citep{luong2022, zygouras2019} often focus on selected major roads or require partial real-time feeds to augment otherwise incomplete data.

In the method discussed here, Gaussian Mixture Models are used in a different capacity, namely to organize an urban region into concentric zones derived from the distribution of local \acro{POI}[Points of Interest]. This zoning supplies a coarse spatial partition, subsequently refined by a hexagonal segmentation that assigns localized speed profiles. Such layering ensures that even a short data collection window can offer meaningful information by aggregating it in structured spatial and temporal bins.

\paragraph{Challenges in Sparse Data Environments.}
Travel time prediction with limited data entails specific difficulties. First, data scarcity amplifies the risk of overfitting if the model is too granular, but oversimplified approaches can average away crucial local patterns. Second, the spatial heterogeneity of urban traffic means that different neighborhoods exhibit distinct speed behaviors, which must be captured efficiently in the absence of rich data. Third, typical daily and weekly fluctuations in traffic remain present, yet subdividing the data into many temporal slices can yield low-volume clusters and unstable estimates. Finally, offline methods cannot rely on sensor updates for near-instant corrections, requiring that any learned pattern be robust to short-term variations. The zonal estimation framework proposed here deals with these challenges through careful zoning, controlled segmentation, and modest temporal aggregation.

\begin{figure}[ht]
\centering
\begin{subfigure}[b]{0.45\textwidth}
    \includegraphics[width=\textwidth]{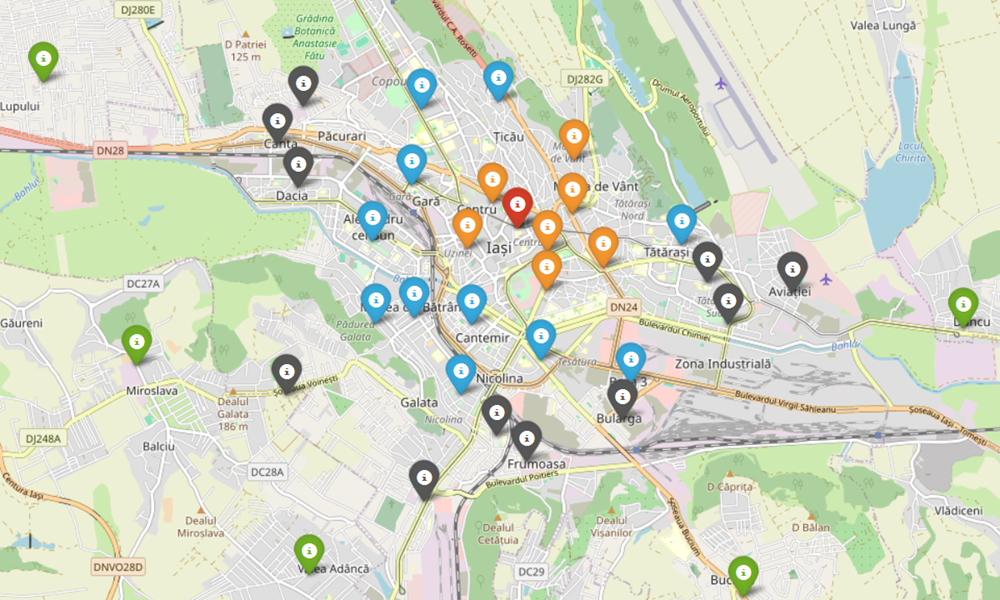}
    \caption{Map of Iași with color-coded POIs representing their GMM-assigned zones.}
\end{subfigure}
\hfill
\begin{subfigure}[b]{0.53\textwidth}
    \includegraphics[width=\textwidth]{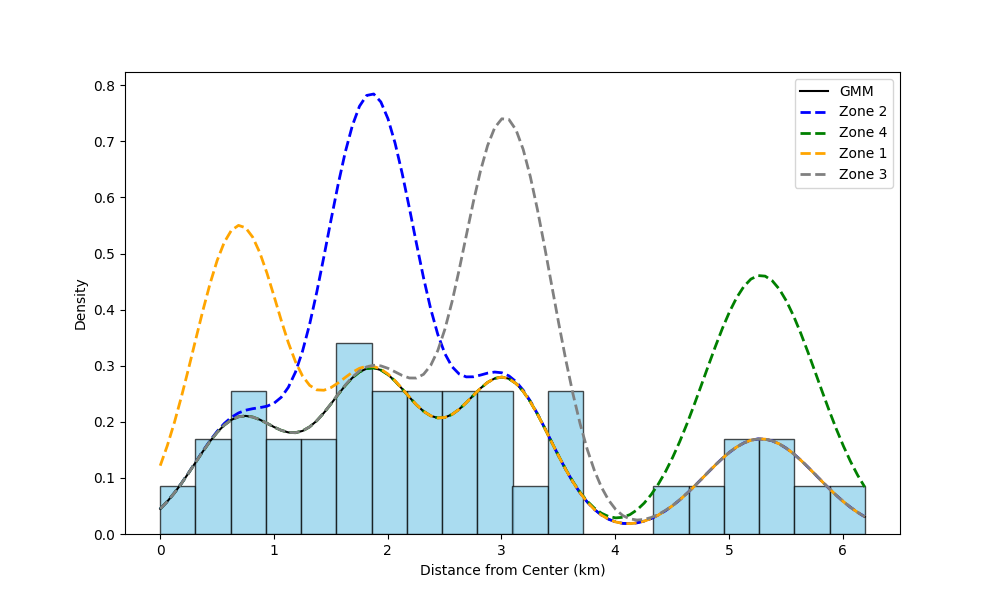}
    \caption{Density curves for each zone and the overall GMM distribution.}
\end{subfigure}
\caption{Zoning using a Gaussian Mixture Model on POI distances. The partitioning into four components aligns with concentric urban development patterns.}
\label{fig:poi_zones}
\end{figure}

\begin{figure}[ht]
\centering
\includegraphics[width=0.45\textwidth]{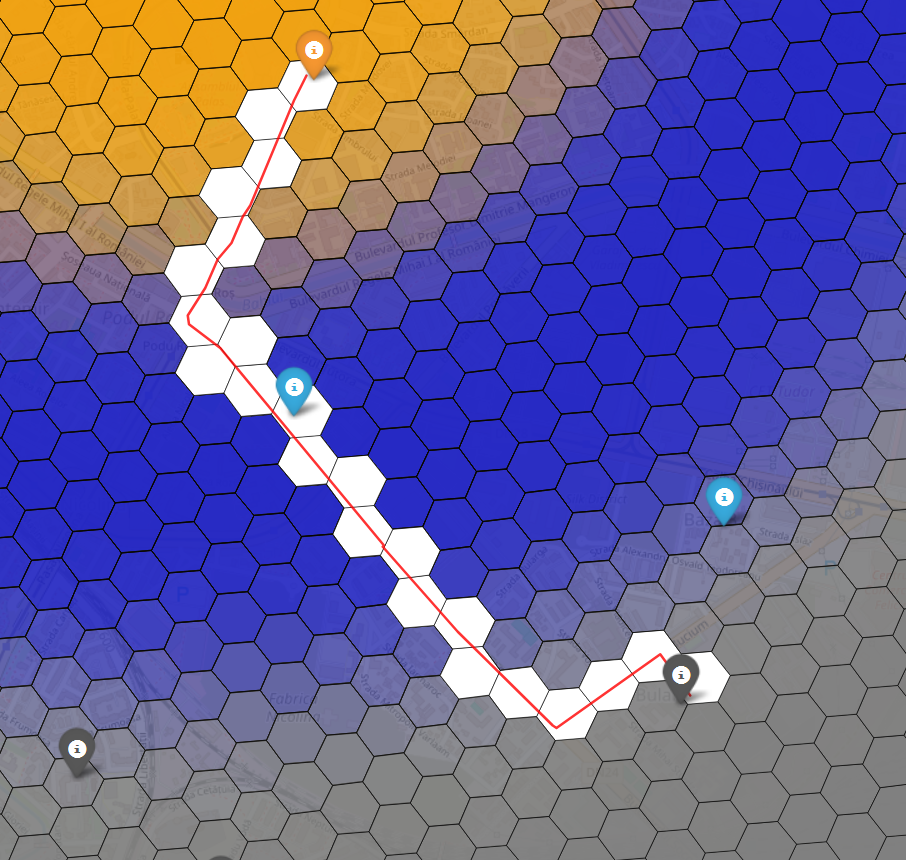}
\caption{A route overlaid on hexagonal cells, each cell contributing a localized speed estimate. The final travel time is derived by combining the time spent in each relevant cell.}
\label{fig:hex_route}
\end{figure}

\paragraph{Methodology.}
Conceptually, the method follows a sequence of steps. First, a \acro{GMM}[Gaussian Mixture Model] is fitted to the radial distribution of urban POIs relative to the city center, generating concentric zones; see figure \ref{fig:poi_zones} for a visualization of the POI in Iasi, Romania. Each zone represents a ring-like region whose boundaries are defined probabilistically, which allows for soft transitions rather than sharp borders. Second, the method applies a hexagonal grid overlay within each zone, thereby dividing the city into cells that capture localized speed dynamics. The choice of hexagonal cells is motivated by their uniform adjacency properties and their suitability for modeling radial movement patterns. Third, speed estimates in each cell are aggregated over short time intervals, typically distinguishing weekdays from weekends. This strategy captures the fundamental weekly traffic cycle without excessively fragmenting an already limited dataset. Fourth, a requested trip’s route is projected onto the hex grid (see figure \ref{fig:hex_route}, with relevant speed data retrieved for each cell and combined to form an overall travel time prediction.

\paragraph{Experimental Setting and Results.}
Implementation was tested with approximately 23,000 route traces collected over a 10-day period in a mid-sized European city. This volume of data reflects a scenario where traffic information is available but remains substantially sparser than the months or years of data on which many commercial APIs rely. In practice, the approach attained a mean absolute percentage error (MAPE) of around 12.9\% against ground-truth travel times, surpassing baseline route estimation APIs under the same constraints. Analysis of zone-level speeds confirms the intuitive observation that central areas exhibit significantly lower average speeds than outer suburban zones, especially during weekday peak hours. The method also highlights how even modest distinctions—such as weekday versus weekend time slices—can substantially refine offline predictions in data-scarce environments.

\paragraph{Limitations and Future Directions.}
Despite its merits, the zonal estimation method has inherent trade-offs. Relying on a single 10-day dataset leaves open questions about how performance scales with larger or smaller data windows. Additionally, the chosen four-zone GMM decomposition might not optimally capture all urban layouts, especially in cities where traffic patterns deviate from a concentric layout. Future work may investigate alternative zoning strategies or adaptive hex resolution to handle more nuanced spatial variations. The integration of partial or delayed real-time data could further enhance estimates, particularly during events that significantly disrupt typical patterns. Smoothing techniques for anomaly detection and correction represent another avenue for strengthening reliability. Overall, the method’s modular design—combining GMM zoning, hex-based segmentation, and temporal aggregation—makes it adaptable and amenable to incremental refinement.

\paragraph{Conclusion.}
The Zonal Estimation Method for Offline Travel Time Prediction offers a practical solution for contexts where data availability is limited. By leveraging a Gaussian Mixture Model to delineate concentric zones, complemented by a hexagonal segmentation and minimal temporal grouping, the approach yields stable speed estimates for route-level predictions. The final system remains sufficiently light to deploy early in the lifecycle of traffic monitoring systems or in regions lacking comprehensive sensor infrastructure. While there is scope for enhancing granularity, fine-tuning temporal slices, and incorporating partial real-time inputs, the current implementation demonstrates that a methodical spatial and temporal structuring can recover a significant portion of predictive accuracy—even with sparse historical data.

\subsection{Cascading Sum Augmentation}
\label{subsec:cascading-sum-augmentation}

Data augmentation plays a critical role in improving the generalization of deep learning systems. Conventional approaches often manipulate individual samples using geometric or photometric transformations, yielding images interpretable by humans. However, recent work suggests that combining multiple samples—even in ways that produce visually unintelligible outcomes—can help neural networks discover beneficial, less obvious patterns \citep{zhang2017mixup, inoue2018japaneseguy}. Building on this idea, \acro{CSA}[Cascading Sum Augmentation] (\cite{10.1007/978-981-97-7419-7_2}) extends the linear combination paradigm to larger groups of examples and deploys a tiered (or “cascading”) training schedule to enhance accuracy on image classification tasks.

\paragraph{Sum Augmentation.}
The simplest form of combining two training images $x_i$ and $x_j$ in a linear interpolation is often written as:
\[
  x_{new} \;=\; \lambda \,x_i \;+\; (1 - \lambda)\, x_j,
\]
with $0 \le \lambda \le 1$ and a corresponding interpolation of labels (e.g., in one-hot format). In \emph{Sum Augmentation}, this idea generalizes to averaging $K$ images at a time:
\[
  x_{\text{new}} \;=\; \frac{1}{K}\sum_{k=1}^{K} x_k,
\]
where each $x_k$ is drawn from the training set. After adjusting the labels accordingly (e.g., dividing label vectors by $K$ or counting how many times a class appears in the group), one obtains multiple aggregated samples for training. Although such averages become less comprehensible visually as $K$ increases, prior experiments show that the neural network still extracts meaningful patterns, presumably because it sees a wide variety of hybrid examples in feature space \citep{zhang2017mixup, inoue2018japaneseguy}.

\paragraph{Cascading Approach.}
A key contribution of CSA is its multi-step training process. Instead of training directly with a fixed $K$, one begins with a large $K$ (e.g., $K=8$), thereby producing heavily averaged training images. After training converges, the best-performing checkpoint is retained, and $K$ is reduced by half. Training resumes from the saved checkpoint, typically converging faster because the model is already partially adapted to linear combinations of images. Eventually, $K$ drops to 1, so the network fine-tunes on the original dataset for a final stage that consolidates its performance. Algorithmically, this might look like:

\begin{center}
\texttt{Train(K=8) -> Save Best -> Train(K=4) -> Save Best -> Train(K=2) -> ... -> Train(K=1)}
\end{center}

This cascading scheme seeks to gradually guide the model from learning features under heavy mixture conditions to less mixed conditions. Empirically, it often yields improved test accuracy relative to using a single value of $K$.

\paragraph{Performance Observations.}
Experiments on CIFAR-10 and CIFAR-100 indicate that CSA delivers notable reductions in error rates over baselines without linear mixing. Gains are especially pronounced when the amount of training data is small or when the network has relatively high capacity. Furthermore, increased $K$ values can impart stronger regularization but also reduce the clarity of individual samples. The cascading approach allows the model to exploit these strongly mixed samples early on, then refine its representations as $K$ gradually moves toward 1. In some experiments, final error rates significantly outperform standard Mixup \citep{zhang2017mixup} and SamplePairing \citep{inoue2018japaneseguy} baselines. See table \ref{tab:combined-cascading-sum-augmentation} where we evaluate CSA in different sample scarcity domains (100 samples/class and 500 samples/class) and model sizes over CIFAR-10 and CIFAR-100.

\begin{table}[ht]
    \caption{CIFAR-10 and CIFAR-100 Test Error using Cascading Sum Augmentation. This table compares the test error rates for different Starting Sum Groups across two datasets (CIFAR-10 and CIFAR-100) and configurations of WideResNet (WRN).}
    \label{tab:combined-cascading-sum-augmentation}
    \begin{center}
    \setlength\tabcolsep{4pt} 
    \renewcommand{\arraystretch}{1.2} 
    \begin{tabular}{l c c c c | c c c c}
        \cmidrule[2pt]{1-9}
        \multirow{2}{*}{\textbf{SSG}} & \multicolumn{4}{c|}{CIFAR-10} & \multicolumn{4}{c}{CIFAR-100} \\
        \cmidrule{2-9} 
        & \multicolumn{2}{c|}{WRN(40,4)} & \multicolumn{2}{c|}{WRN(28,10)} & \multicolumn{2}{c|}{WRN(40,4)} & \multicolumn{2}{c}{WRN(28,10)} \\ 
        \cmidrule{2-9}
        & 100 & 500 & 100 & 500 & 100 & 500 & 100 & 500 \\
        \cmidrule{1-9}
        8 & 31.19\% & 13.51\% & \textbf{29.34\%} & 13.21\% & 35.3\% & 20.34\% & 35.01\% & 18.63\% \\
        4 & \textbf{30.65\%} & \textbf{13.29\%} & 30.36\% & \textbf{12.81\%} & \textbf{34.7\%} & \textbf{19.85\%} & \textbf{33.01\%} & \textbf{18.09\%} \\
        2 & 32.44\% & 14.88\% & 30.99\% & 13.1\% & 35.56\% & 20.17\% & 34.44\% & 18.17\% \\
        Baseline & 47.36\% & 20.15\% & 43.47\% & 19.96\% & 43.2\% & 23.87\% & 41.72\% & 21.94\% \\
        \cmidrule[2pt]{1-9}
    \end{tabular}
    \end{center}
\end{table}

\paragraph{Test-Time Sum Augmentation.}
Although Sum Augmentation is primarily geared toward training, it can also be extended to inference. One method, called \emph{Test-Time Sum Augmentation}, combines each query image with one or more random images from the test set to create augmented inputs. Predictions from multiple such hybrids are then averaged (or “ensembled”) to produce a final classification. This approach can further boost accuracy (see table \ref{tab:combined-test-error}) and has shown some promise in mitigating adversarial vulnerabilities, since creating adversarial perturbations that consistently fool the model on all hybrid combinations appears more challenging \citep{su2019onepixel, ilyas2019adversarial}.

\begin{table}[ht]
    \caption{CIFAR-10 and CIFAR-100 Test Error using Test Time Sum Augmentation. NT: Normal Test, TTSA: Test Time Sum Augmentation. The table presents error rates for different $K$ values under both NT and TTSA for WideResNet(40,4) (WRN) and WideResNet(28,10) configurations, comparing performance on CIFAR-10 and CIFAR-100 datasets.}
    
    \label{tab:combined-test-error}
    \begin{center}
    \begin{tabular}{l c c c c | c c c c}
        \cmidrule[2pt]{1-9}
        & \multicolumn{4}{c|}{CIFAR-10} & \multicolumn{4}{c}{CIFAR-100} \\
        \cmidrule{2-9}
        \multirow{2}{*}{\textbf{$K$ values}} & \multicolumn{2}{c}{WRN(40,4)} & \multicolumn{2}{c|}{WRN(28,10)} & \multicolumn{2}{c}{WRN(40,4)} & \multicolumn{2}{c}{WRN(28,10)} \\
        \cmidrule{2-9}
        & NT & TTSA & NT & TTSA & NT & TTSA & NT & TTSA \\
        \cmidrule{1-9}
        8-4-2 & 6.53\% & 6.32\% & 5.58\% & 5.11\% & 27.18\% & 24.06\% & 25.11\% & \textbf{20.25\%} \\
        4-2 & 6.91\% & 6.32\% & 5.35\% & \textbf{5.09\%} & 27.22\% & 23.9\% & 24.42\% & 20.46\% \\
        2 & 6.19\% & \textbf{6.03\%} & 5.74\% & 5.39\% & 27.19\% & \textbf{23.58\%} & 24.32\% & 20.29\% \\
        \cmidrule[2pt]{1-9}
    \end{tabular}
    \end{center}
\end{table}

\paragraph{Future Directions.}
Areas that remain to be explored include altering the order or spacing in which $K$ is reduced, investigating more adaptive mixing ratios (rather than the fixed 0.5 average), and examining diverse loss functions. Another line of research involves systematically studying CSA’s robustness under adversarial attack scenarios, as preliminary evidence suggests that mixing strategies might enhance defenses by diluting attack perturbations. Finally, verifying its scalability to high-resolution datasets and tasks such as object detection or segmentation could clarify how general this concept is beyond classification on small images.

\subsection{Urban Development Prediction Using Satellite Imagery}
\label{subsec:urban-development-prediction}

Forecasting how urban areas will evolve over time is pivotal for effective city planning and resource management. As metropolitan regions undergo rapid changes driven by demographic growth and economic pressures, advanced modeling can highlight hotspots of potential infrastructure development and guide policymakers toward sustainable expansion. The work presented here proposes a framework that uses multi-temporal satellite imagery to predict urban growth patterns over monthly or annual intervals. Central to this approach is an autoencoder—a deep neural network architecture composed of an EfficientNet-based encoder and a convolutional decoder—which captures spatial structures and temporal signals from high-resolution satellite data.

\paragraph{Context and Motivation.}
Urban environments are continuously reshaped by factors such as population shifts, economic stimuli, and infrastructure initiatives \citep{mantelas2010fuzzy, mantelas2007fuzzyca}. Precisely modeling and anticipating these transformations can inform stakeholders—planners, investors, and environmental groups—who strive to balance urbanization demands with sustainable practices. The surge of open-access remote sensing data and improvements in geospatial technology have made possible fine-scale observation of city development at resolutions and frequencies once considered impractical. Of particular note is the SpaceNet~7 initiative, which provides monthly, high-resolution satellite mosaics spanning diverse global locations, thereby enabling machine learning systems to track and predict subtle land-use changes.

\paragraph{Related Work.}
Deep learning has emerged as a powerful tool for analyzing time-series remote sensing data, excelling in both semantic segmentation and change detection tasks \citep{corbane2021cnn, elmendili2020multitemporal}. Early studies often focused on broad land-cover classification or built-up area delineation using CNNs, but fewer attempts have been made to capture monthly or sub-annual urban development trends. Nonetheless, prior efforts on sentinel-based datasets highlight that advanced encoder-decoder architectures (e.g., autoencoders) can robustly learn local patterns when sufficient labeled data are accessible. This underscores the applicability of specialized networks for multi-temporal prediction in urban landscapes.

\paragraph{Dataset Choice and Preparation.}
We adopt the \textbf{SpaceNet~7} dataset \citep{vanetten2021challenge}, a comprehensive collection of monthly satellite imagery from 2017 to 2020, covering more than a hundred distinct urban or peri-urban sites worldwide. Each site typically has 24 images captured at roughly one-month intervals, providing consistent temporal snapshots suitable for modeling. The images exhibit relatively uniform cloud-free conditions and consistent spatial alignment, making them well-suited for detecting new infrastructure or incremental expansions. A PyTorch data loader facilitated shuffling, batching, and applying standard augmentation procedures such as flips, rotations, and random crops.

\paragraph{Model Architecture: Autoencoder.}
An \emph{autoencoder} underlies our approach, split into three main components:
\begin{itemize}
  \item \textbf{Encoder (EfficientNetV2-S):} Drawing on the model family described in \citep{tan2021efficientnetv2}, this encoder strikes a balance between speed and capacity. We remove its original classification head, preserving high-level representations while retaining spatially rich feature maps.
  \item \textbf{Temporal Embedding Layer:} This module translates a “time skip” indicator (e.g., \(-12\) to look one year into the past, or \(+12\) to predict one year forward) into a 256-dimensional embedding. The embedding is added to the encoder’s latent feature space, guiding the subsequent decoder on whether to reconstruct a historical or future state.
  \item \textbf{Decoder (CNN):} A sequence of upsampling (transpose convolution) layers recovers the original image resolution. If the network targets RGB output, the final activation is Tanh (ensuring outputs lie within \([-1,1]\)); for single-channel outputs (e.g., heatmaps), a ReLU activation is retained.
\end{itemize}

\paragraph{Loss Functions and Training.}
The model’s training objective combines two primary losses:
\begin{enumerate}
  \item \textbf{Mean Squared Error (MSE):} Encourages pixelwise fidelity by minimizing squared differences between prediction and ground-truth images.
  \item \textbf{Multi-Scale Structural Similarity (MS-SSIM):} Evaluates perceptual alignment by measuring the luminance, contrast, and structural similarity across multiple scales \citep{pondaven2022cnp}.
\end{enumerate}
A weighted sum of MSE and MS-SSIM fosters sharper reconstructions and better local-global coherence. The AdamW optimizer (weight decay of \(1\!\times\!10^{-5}\)) is used, starting with a learning rate of \(1\!\times\!10^{-3}\). Training on Kaggle’s GPU resources can extend from 40 to 90 hours per run, with additional fine-tuning at \(1\!\times\!10^{-4}\) learning rate for stability in later epochs.

\paragraph{Evaluation and Experiments.}
\begin{itemize}
  \item \textbf{Local Romanian Dataset:} We tested the model on cities like Iași, Bucharest, and Constanța using custom satellite images spaced one year apart. Predictions accurately identified expansions into open areas while refraining from marking major roads as growth zones.
  \item \textbf{Predicting the Past (SpaceNet~7):} By specifying negative time skips, the model reconstructs scenes from up to a year earlier. It showed a proclivity to expand structures in less-populated areas, reflecting typical sprawl patterns.
  \item \textbf{Long-Horizon Forecasting (24 months):} A two-stage inference (each with +12-month increments) yielded 24-month predictions. Results indicated incremental outward expansion from established city cores, aligning with standard urban growth trends.
  \item \textbf{Desert Environment Case}: Minimal building changes in arid zones led to near-constant predictions, matching known slow-growth conditions in harsh climates.
\end{itemize}

\paragraph{Findings and Limitations.}
The model consistently projects developments into semi-rural or vegetated fringes, mirroring observed urbanization trends. Visualizing pixelwise differences between inputs and predictions highlights high-confidence development zones in red, potentially aiding planning. Nevertheless, hyperparameter tuning is extensive, and training time is long, primarily due to the high dimensionality of monthly imagery. Overfitting risks and subtle errors in stable regions (e.g., deserts) also underscore the need for careful validation.

\paragraph{Conclusions and Future Directions.}
By embedding temporal cues in an autoencoder framework, this approach shows strong promise for anticipating city expansions. MSE and MS-SSIM jointly optimize spatial accuracy and perceptual detail. Going forward, researchers could:
\begin{enumerate}
  \item \textbf{Incorporate Ancillary Data:} Augment pixel intensities with socioeconomic or climatic factors for richer modeling.
  \item \textbf{Adopt Attention Mechanisms:} Enhance performance by allowing the network to selectively focus on the most rapidly changing areas.
  \item \textbf{Implement Real-time Updates:} Periodically retrain with newly available satellite scenes to track accelerating or changing growth patterns.
\end{enumerate}
Such refinements may provide even greater clarity and foresight to urban stakeholders, helping align infrastructural expansions with environmental and social considerations.

\subsection{Social Media Post Impact Prediction using Computer Vision and Natural Language Processing}
\label{subsec:social-media-impact}

An important question for businesses, influencers, and marketing specialists is how a particular social media post will perform once published. On platforms such as Twitter, image- and text-based features can jointly affect the level of user engagement (e.g., likes, retweets). This subsection details a system (\cite{minuct2022social}) that predicts the anticipated popularity of a tweet by integrating image and text processing through deep learning methods.

\paragraph{Overall Architecture and Motivation.}
The application employs a microservices-based architecture that divides responsibilities among multiple modules: (1)~user account management, (2)~a prediction service, and (3)~a management layer for maintaining a history of previous predictions. Each module is deployed on cloud infrastructure and communicates through well-defined interfaces. The core idea is that users can log in with their Twitter credentials, submit a proposed tweet containing both text and an attached image, and receive an estimate of how many likes or reactions that tweet is likely to obtain. By combining insights from both visual and textual cues, the system aims to identify how the style of an image and the semantics of a tweet’s text jointly contribute to overall engagement.  

\paragraph{Dataset Construction.}
A proprietary dataset of approximately 120,000 tweets was created by daily querying the Twitter API for specific categories such as travel, animals, sports, and technology. Each record in this dataset includes the tweet’s text, an associated image, and metadata reflecting various engagement metrics (e.g., number of likes, user followers). The images were resized (to dimensions such as 366\(\times\)512\(\times\)3) and padded if necessary, ensuring a uniform shape for training. Preliminary analysis of the dataset indicated that over 30\% of tweets have zero likes, and the distribution of likes is highly skewed, with some tweets garnering tens of thousands of likes. To address this, the continuous variable (number of likes) was binned into intervals, creating 10 discrete classes (bins) based on quantiles of observed engagement.

\paragraph{Modeling Approach.}
A hybrid architecture combines:
\begin{itemize}
    \item \textbf{Computer Vision Sub-model:} A ResNet-18 convolutional neural network \citep{b7} was fine-tuned on the resized images. ResNet-18’s output layer is adapted to produce logits corresponding to the 10 output bins.
    \item \textbf{Language Sub-model:} A BERT-based model \citep{b6} handles the tweet’s textual content. Textual preprocessing includes removing URLs, trimming extraneous whitespace, and tokenizing. As with ResNet-18, the final layer is adjusted to produce 10-dimensional logits.
\end{itemize}
After both sub-models are trained, their outputs (logits) are averaged or concatenated and passed through a final decoder layer, thereby generating a single prediction. At inference, the system applies an \(\mathrm{argmax}\) over the 10 bins to predict a discrete range of likely engagement.

\paragraph{Implementation and Deployment.}
The hybrid model is stored in a cloud-based environment, and upon startup, the prediction service downloads the model weights and initializes them for inference. Users can submit requests through a web interface where they provide an image and optional text. Once the model has produced an engagement bin for the tweet, the service transforms that bin into an approximate center value for display. The entire system interacts with Twitter’s OAuth mechanism for user authentication, ensuring that posted or predicted content remains tied to legitimate user accounts. In addition, historical predictions are saved, allowing users to revisit and compare previously forecasted engagements for different tweets.

\paragraph{Practical Use Cases and Limitations.}
In typical usage, the user provides multiple images or textual variations for a potential tweet, obtains predicted engagement scores for each combination, and then selects the posting most likely to succeed. This can be advantageous for content creators who wish to test multiple designs or phrasings before officially tweeting. However, when the model encounters niche or unusual topics far removed from the categories included in the training data, predictions can be unreliable. Additionally, care must be taken when comparing older predictions to new ones, as shifts in user trends and platform dynamics mean that the underlying distribution of tweet engagements can evolve over time.

\paragraph{Initial Results and Observations.}
Using a holdout test set of 20,000 tweets, the system’s accuracy in predicting the correct impact bin was around 10\%, indicating that further refinement is needed. The skewed nature of like counts poses a challenge. In particular, many tweets concentrate in lower-impact bins, making moderate or high-impact bins more difficult to predict. Despite these obstacles, the approach demonstrates potential: the hybrid model outperforms single-modality baselines (i.e., just text or just image), suggesting that modeling both visual and textual aspects improves performance.

\paragraph{Comparison to Related Systems.}
Similar commercial or research applications, such as LikelyAI, focus on predicting Instagram photo engagement. Although both solutions harness image and text features, the differences in platform user behavior (Instagram vs. Twitter) and the underlying model architectures make direct comparisons difficult. Nonetheless, both types of systems similarly integrate with their respective social platforms for seamless posting and data retrieval, reinforcing that such integration strategies are beneficial for building user-friendly, real-time prediction services.

\paragraph{Conclusions and Future Directions.}
The developed system illustrates how combining computer vision and language modeling can yield valuable insight into a tweet’s prospective popularity. Continued improvements involve:  
\begin{enumerate}
    \item \textbf{Addressing Skewed Labels:} Techniques such as focal loss or more fine-grained binning might lessen the imbalance that arises when many tweets have zero or few likes.
    \item \textbf{Adaptive or Online Learning:} Incorporating real-time feedback as Twitter trends change could keep the model updated and mitigate concept drift.
    \item \textbf{Exploring Additional Features:} Incorporating user-specific data (e.g., user history, social network graph features) may improve predictions.
\end{enumerate}
Such enhancements could strengthen the reliability and applicability of social media engagement forecasting in both marketing and personal use scenarios.

\chapter{Conclusion and Future Work}

\ifpdf
    \graphicspath{{Chapter8/Figs/Raster/}{Chapter8/Figs/PDF/}{Chapter8/Figs/}}
\else
    \graphicspath{{Chapter8/Figs/Vector/}{Chapter8/Figs/}}
\fi

This final chapter synthesizes the core insights and overarching findings derived from the research presented throughout this dissertation. In addressing the complexity of medical imaging analysis, we have drawn upon methodologies spanning self-supervised learning, multimodal modeling, and large-scale deep architectures. By conducting experiments across multiple imaging modalities—ranging from MRI scans to microscopic slides—and examining tasks such as classification, segmentation, and risk prediction, this work sought to demonstrate how a unified yet adaptable framework can be both effective and computationally feasible in a diverse set of clinical contexts.

Beyond the principal theme of building robust, transferable models for medical images, several auxiliary contributions were made in related domains. These projects, though not the thesis's central focus, illustrate broader applications of the proposed techniques in areas such as prehospital stroke detection using mobile devices, travel-time forecasting in data-scarce urban settings, and novel data fusion approaches for brain MRI augmentation. Taken together, they underscore the versatility and potential impact of carefully designed deep learning strategies when adapted to the unique characteristics and constraints of different fields.

In the following sections, we revisit the specific contributions made by this dissertation, discuss their significance for medical image analysis, and outline promising directions for future investigations.

\section{Summary of Contributions}

This dissertation advances deep learning for medical imaging by introducing frameworks and strategies that address diverse tasks, including classification, segmentation, and risk prediction, under both supervised and self-supervised regimes. The work emphasizes unifying designs that can accommodate a variety of modalities (e.g., MRI, X-ray, microscopic slides) in order to foster reusable and adaptable models suitable for real-world clinical scenarios. A summary of the major contributions follows:

\begin{itemize}
    \item \textbf{Medformer Architecture:} A novel multitask and multimodal model that integrates advanced transformer components with flexible input and output ``Adaptformers.'' This design unifies different dimensionalities (2D and 3D) and imaging modalities into a single backbone, enabling joint learning and knowledge transfer. Empirical evaluations on the MedMNIST suite illustrate its capability to handle classification tasks across highly varied datasets and confirm the merits of both multi-task training and self-supervised pretraining.

    \item \textbf{Robust Self-Supervised Learning Approaches:} The thesis underscores the importance of self-supervision, particularly in data-scarce medical contexts. Through tasks such as contrastive learning and pretext-based consistency training, the models benefit from unlabeled scans and improve performance on downstream classification and segmentation tasks.

    \item \textbf{Supplementary Techniques in Brain MRI and Clinical Data:} Beyond Medformer, the BrainFuse method offers a data-fusion augmentation approach for brain MRI scans, increasing dataset diversity via slice interpolation and smooth volumetric transitions. In parallel, work on stroke risk prediction and cancer treatment recommendation (REVERT project) and backforward propagation for training stability highlights further methodological progress relevant to complex clinical datasets.

    \item \textbf{Domain-Adapted Extensions and Multidisciplinary Case Studies:} Several complementary projects demonstrate how techniques developed for medical imaging can be adapted to other domains. These include specialized mobile applications for prehospital stroke detection, offline travel-time estimation with sparse data, augmentation techniques like Cascading Sum, and predictive analytics in social media and urban development. Although these are not strictly medical tasks, they reinforce that careful model design and domain adaptation can yield tangible benefits in heterogeneous data environments.
\end{itemize}

Collectively, these contributions strengthen the case for flexible, multimodal deep learning systems in medical imaging. They also show that careful integration of domain knowledge, self-supervised methods, and consistent architectural choices can extend into various related fields, thereby enlarging the overall reach and applicability of the research.

\section{Implications for Medical Image Analysis}

The findings and methodologies presented throughout this dissertation underscore the growing potential of deep learning in transforming medical imaging workflows, particularly in diverse and data-scarce scenarios. By introducing architectures that can accommodate multiple imaging modalities under a unified framework, this research offers a tangible pathway toward reducing the proliferation of single-task models typically required in clinical practice. For instance, radiology departments often rely on separate systems for distinct modalities (MRI, CT, or X-ray), and even finer task-specific models are used within each modality. In contrast, the proposed multitask, multimodal designs encourage a shared representation space that can enhance transferability and reduce the maintenance burden. This, in turn, can streamline operational pipelines and potentially lower the costs associated with training and deploying multiple specialized networks in hospital settings.

An important consideration in the clinical domain is the frequent scarcity of labeled data, arising from the complexity of annotations and the privacy constraints on patient information. The self-supervised approaches developed here offer a practical strategy to exploit unlabeled data, which are far more abundant in most medical institutions. By learning generalizable features from large-scale, unlabeled scans and then fine-tuning on smaller labeled sets, these methods present viable solutions for both large hospitals with varied archives and smaller clinics seeking to adopt data-driven diagnostics. The capacity of these models to adapt efficiently across diverse tasks and image types indicates that they can be particularly valuable for conditions in which annotated samples are difficult to gather or for which pathology incidence rates are relatively low.

Additionally, the modular nature of the proposed architectures aligns well with the need for explainability and adaptiveness in clinical environments. Tasks such as segmentation of tumors or classification of lesions often demand interpretability and trust, given the high stakes associated with diagnostic recommendations. The thesis demonstrates that it is possible to incorporate flexible domain embeddings (as in the Adaptformer modules) to represent modality-specific nuances without sacrificing the ability to provide coherent and well-structured visual explanations of the learned features. Moreover, as new modalities or tasks emerge—be they a novel diagnostic imaging protocol or an updated labeling scheme—the extensible backbone can be reconfigured by adding corresponding domain or task embeddings, minimizing the need for complete architectural rewrites. In this sense, the work points toward a more adaptive and resilient approach to deep learning in medical imaging, one that readily accommodates technological changes and evolving clinical standards.

\section{Future Research Direction}

Although the methods and architectures introduced in this dissertation have shown promise in handling a range of modalities, tasks, and data regimes, there remain several avenues for continued investigation. Further development of these ideas can strengthen their clinical viability, broaden their applicability, and refine theoretical understanding. In particular, the following directions suggest ways to expand upon and deepen the contributions made thus far:

\paragraph{Scaling to Larger and More Complex Datasets.}
While the models evaluated here operate effectively on datasets such as MedMNIST or specific institutional repositories, real-world medical imaging often involves considerably larger and more heterogeneous collections. Future work might explore scaling multitask architectures to millions of unlabeled or partially labeled scans, potentially requiring distributed training strategies, more sophisticated data loaders, and refined memory management. Research into robust large-scale self-supervision for three-dimensional imaging holds particular promise, given the growing number of institutions that store extensive volumetric data.

\paragraph{Enhanced Domain Adaptation and Personalization.}
The use of domain-specific latent embeddings in the proposed frameworks serves as a basis for adaptation to varied imaging contexts. Devising even more flexible domain-adaptation mechanisms—particularly ones that can incrementally learn new body regions or disease types—could further streamline the integration of emerging modalities. Additionally, there is scope for methods that adapt models to specific patient populations or individual characteristics, which could help address anatomical variability and yield more personalized diagnostic tools.

\paragraph{Improving Interpretability and Trust.}
Although the transformer-based and self-supervised models presented here provide promising results, many clinicians still regard interpretability as a key factor in adopting AI. Investigating improved explainability tools—ranging from saliency maps to more structured hierarchical explanations—might increase clinical acceptance. Frameworks that combine attention-based mechanisms with domain-specific interpretability cues (e.g., highlighting clinically meaningful regions of medical scans) could make model decisions more transparent and trustworthy.

\paragraph{Robustness, Fairness, and Privacy.}
As deep learning models move closer to widespread clinical deployment, questions concerning reliability and equity become increasingly critical. Future efforts might focus on systematically evaluating how robust the proposed architectures are to input perturbations, scanner variations, or domain shifts caused by demographic factors. Researchers could also examine differential privacy approaches or federated learning protocols, ensuring compliance with legal and ethical standards while allowing diverse healthcare institutions to collaborate on shared models.

\paragraph{Extending the Range of Downstream Tasks.}
The current thesis has emphasized classification, but many applications in medical imaging—such as lesion characterization, disease progression prediction, or multi-modal text-to-image retrieval—remain less explored. Adaptations of the proposed transformer-based frameworks or the BrainFuse augmentation scheme could be tested in these contexts, possibly involving multimodal data types such as EHR text, genomic information, or wearable-sensor streams. Investigating new downstream tasks would also confirm the models’ versatility and potential for cross-domain knowledge transfer.

Taken together, these directions suggest a wealth of opportunities to refine, extend, and solidify the contributions outlined in this dissertation. The evolving landscape of medical imaging, coupled with the rapid pace of methodological innovation in deep learning, invites ongoing research. In pursuing these lines of inquiry, future studies can continue to build toward practical, adaptive, and trustworthy systems that can further enhance the effectiveness of diagnostic tools and ultimately improve patient care.


\begin{spacing}{0.9}


\bibliographystyle{apalike}
\cleardoublepage
\bibliography{References/references} 



\end{spacing}





\printthesisindex 

\end{document}